\pdfoutput=1
\documentclass{article} 
\usepackage{veriedit_preprint,times}

\usepackage{amsmath,amsfonts,bm}

\def\eqref#1{equation~\ref{#1}}

\def\1{\bm{1}}

\DeclareMathAlphabet{\mathsfit}{\encodingdefault}{\sfdefault}{m}{sl}
\SetMathAlphabet{\mathsfit}{bold}{\encodingdefault}{\sfdefault}{bx}{n}

\usepackage{hyperref}
\usepackage{url}
\usepackage{xcolor}
\usepackage{graphicx}
\usepackage{amssymb} 
\usepackage{longtable}
\usepackage{float} 
\newcommand{\draft}[1]{#1}
\newcommand{\humanauditfix}[1]{#1}
\newcommand{\revisionblue}[1]{#1}
\newcommand{\revisionred}[1]{#1}
\newcommand{\exampleswap}[1]{#1}
\newcommand{\aSixFix}[1]{#1}
\newcommand{\abbrreview}[1]{#1}
\newcommand{\submissionreview}[1]{#1}
\newcommand{\humanauditreview}[1]{#1}

\title{Beyond Plausibility: Verifiable Fine-Grained\\Image Editing on Structured Assets}

\author{%
  Muyao Wang\textsuperscript{1,2}\thanks{This work was completed at SB Intuitions.}\quad
  Chen Zhu\textsuperscript{1}\quad
  Shiqi Yang\textsuperscript{1}\quad
  DongHyun Hwan\textsuperscript{1}\quad
  Hideki Nakayama\textsuperscript{2}\\[3pt]
  \small\textsuperscript{1}SB Intuitions\qquad
  \textsuperscript{2}University of Tokyo}

\newcommand{\datasetareview}[1]{{#1}}
\begin{document}

\maketitle


\begin{abstract}

Fine-grained image editing requires more than producing a visually plausible 
result: an editor must execute the requested attribute change precisely while 
leaving everything else intact. However, existing benchmarks leave a critical 
gap between realism and verifiability: benchmarks built on realistic images 
typically rely on human or \abbrreview{vision--language model} judgments, while deterministic evaluation has 
largely focused on synthetic shape canvases, with application-oriented 
extensions primarily limited to charts. 
This makes it difficult to determine precisely how much of a requested edit 
was executed, where unintended changes occurred, and whether small differences 
between models reflect genuine editing capability or evaluator uncertainty.
To bridge this gap, we present \textbf{VeriEdit-Bench}, a benchmark for 
fine-grained, instruction-faithful image editing across realistic structured 
assets with deterministic, four-axis evaluation. Its 1,740 cases are compiled 
from the source code of 153 \abbrreview{Scalable Vector Graphics (SVG)} graphics, charts, web interfaces, and 
presentation slides. Controlled source-code edits preserve the original 
visual context while yielding exact target images, pixel-level edit masks, 
and explicit edit specifications, enabling reproducible scoring along four 
axes: edit fidelity, preservation, localization, and magnitude. 
Evaluating eleven editors, we find that even the strongest model remains far from 
full credit; rankings for the same recoloring operation reverse between charts and 
SVG graphics; and outputs with similar pixel-accuracy profiles can still differ 
substantially in localization and change magnitude.
This decomposition yields graded, verifiable feedback and exposes model-specific
capability and failure profiles that holistic scores or evaluator-dependent
judgments may obscure.
The benchmark and evaluation code are available at
\url{https://github.com/flare200020/VeriEdit-Bench}.
\end{abstract}

\begin{figure}[H]
  \centering
  \includegraphics[width=\linewidth]{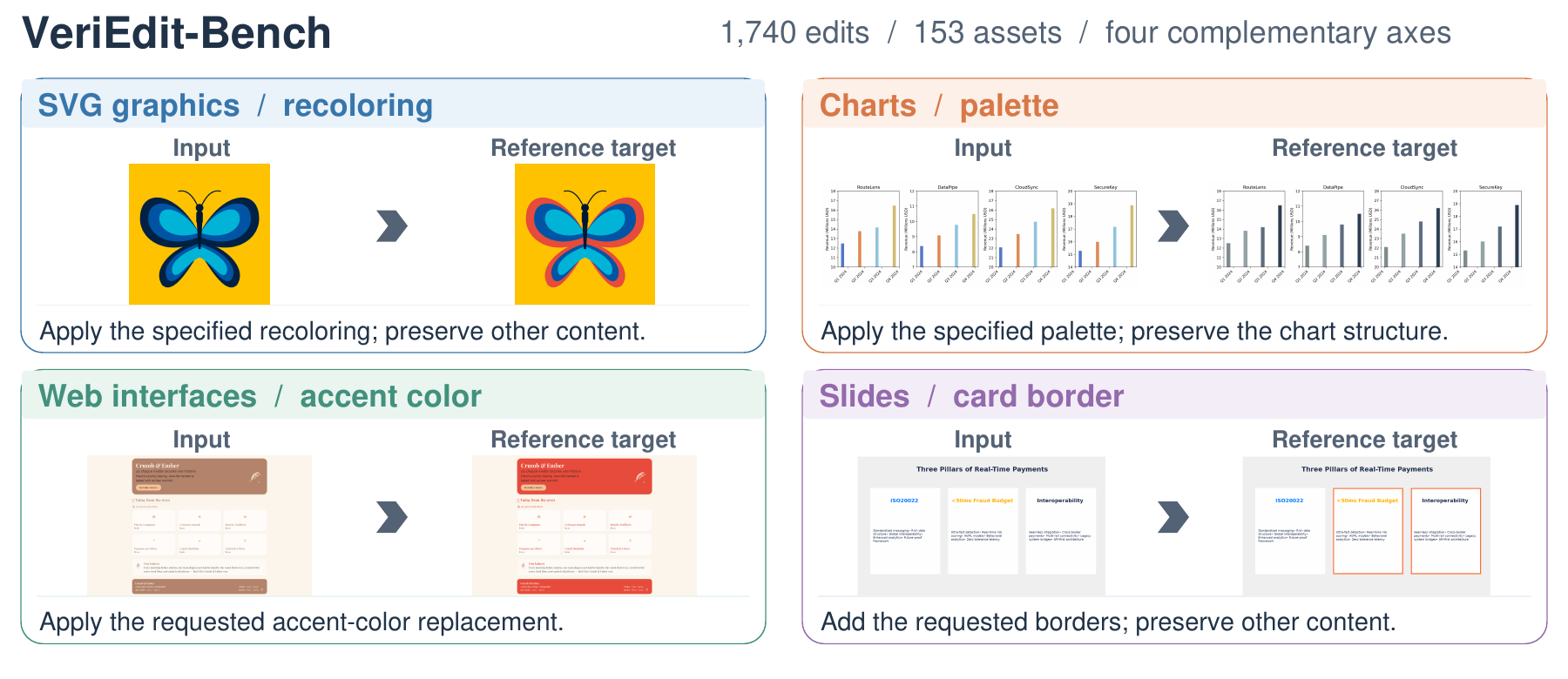}%
  \small
  \caption{Overview of VeriEdit-Bench. Examples from SVG graphics, charts, web interfaces, and slides; each task specifies a controlled edit that preserves the surrounding context.}
  \label{fig:dataset-overview}
\end{figure}

\newpage
\section{Introduction}

Image editing has rapidly evolved from coarse semantic manipulation toward
instruction-following systems capable of modifying increasingly specific
visual attributes while preserving the surrounding content. This progress has
motivated a parallel line of work on image editing benchmarks. Early and
widely used benchmarks evaluate editing models on real images through human
judgment, perceptual similarity, or vision--language models (VLMs), covering
object addition, removal, replacement, and attribute modification
\citep{DBLP:conf/nips/ZhangMCSS23,DBLP:journals/corr/abs-2310-02426,DBLP:conf/nips/MaJYLWZZSJ24}.
More recent benchmarks broaden this scope toward complex, multi-turn, and
reasoning-intensive editing~\citep{DBLP:conf/nips/YeHLLYYHY25,DBLP:journals/corr/abs-2504-13143,han2025unireditbench}.
Meanwhile, increasingly sophisticated VLM-based evaluators decompose editing
quality into fine-grained factors such as instruction fidelity, preservation,
localization, and perceptual quality~\citep{DBLP:journals/corr/abs-2602-13028}.
Together, these benchmarks have substantially improved the coverage and
interpretability of image editing evaluation.

Despite this progress, a central challenge remains in combining
\emph{realism} with \emph{verifiability}.
Existing benchmarks already provide reference images, spatial grounding,
and fine-grained assessments of edit correctness and content
preservation~\citep{DBLP:conf/nips/ZhangMCSS23,DBLP:journals/corr/abs-2505-11493}.
For open-ended edits of realistic images, however, multiple outputs may
satisfy the same instruction, and evaluation commonly relies on human
judgments, learned perceptual measures, or VLM judges.
Such scores do not necessarily provide direct, deterministic verification
against an exact reference, so it remains difficult to determine precisely
how much of a requested change was executed, where unintended changes
occurred, or whether small differences between models reflect capability
or evaluator uncertainty~\citep{kumar2026vlm}.

Deterministic benchmarks offer an important step toward direct verification.
PaintBench constructs precisely specified operations and evaluates their
outputs against pixel-level ground truth, primarily on synthetic shape
canvases, with application-oriented extensions to
charts~\citep{DBLP:journals/corr/abs-2606-00188}.
StructBench extends evaluation to structured visual generation
and editing, assessing factual correctness through VLM-based
questions~\citep{DBLP:journals/corr/abs-2510-05091}.
\submissionreview{PaintBench establishes deterministic verification of specified
edits, including a chart extension; StructBench uses rendered targets to
assess structured facts primarily through VLM questions. What remains
underexplored is how target execution and unintended changes vary across
realistic structured assets under a common deterministic protocol.}

We introduce \textbf{VeriEdit-Bench} to bridge this gap.
Our key insight is that structured visual assets provide editable source
representations that explicitly define their visual attributes---a property
typically unavailable in natural-image editing.
Rather than synthesizing isolated geometric scenes, we construct 1,740
editing cases \abbrreview{by editing 153 constructed source assets:
Scalable Vector Graphics (SVG) graphics, charts, web interfaces, and slides.}
Each edit modifies a specified attribute while preserving the remaining
source context. Rendering the original and edited sources yields exact
input--target pairs and pixel-level edit masks, accompanied by explicit
edit specifications. The source is used 
only to construct and score tasks; the evaluated editor receives the rendered 
image and the instruction alone, so the benchmark measures raster editing rather 
than code manipulation.
These annotations enable deterministic evaluation along four complementary
axes: \emph{edit fidelity}, whether the requested result is achieved;
\emph{preservation}, whether unrelated content remains intact;
\emph{localization}, whether modifications are confined to the intended
region; and \emph{magnitude}, whether their extent matches the requested
change.
This formulation turns fine-grained editing into directly verifiable
measurements, providing graded feedback even when an edit is only
partially successful.

Finally, evaluating eleven contemporary image editors shows that precise editing remains challenging: the best open-weight model reaches a composite 
of 22 out of 100. The same recoloring operation ranks two models in opposite order 
on charts and SVG graphics; outputs with nearly identical PaintBench accuracy curves 
differ sharply in localization and magnitude; and strong target-color attainment 
coexists with damage to surrounding content. Controlled pixel- and source-level 
perturbations confirm that each axis responds to the error it is designed to detect. 
VeriEdit-Bench thus provides a diagnostic account of both the capabilities that
current models already possess and the failures that still prevent faithful execution.  

Our contributions are summarized as follows:
\begin{itemize}
    \item We introduce \textbf{VeriEdit-Bench}, a benchmark for fine-grained,
    instruction-faithful image editing containing 1,740 cases constructed from
    153 SVG graphics, charts, web interfaces, and presentation slides. By
    applying controlled edits to source representations, each case provides an
    exact target image, pixel-level edit mask, and explicit edit specification.

    \item We propose a deterministic four-axis evaluation framework that
decomposes image editing performance into \emph{edit fidelity},
\emph{preservation}, \emph{localization}, and \emph{magnitude}, enabling
reproducible and graded measurement of partially correct edits without
relying on human or VLM judgments for the core evaluation, and validate
its response with controlled pixel-level and source-level perturbations.

    \item We conduct a systematic evaluation of eleven contemporary image editing
    models and reveal distinct capability and failure profiles that can be
    obscured by aggregate or evaluator-dependent scores, providing diagnostic
    evidence for where current models succeed and where fine-grained editing
    remains challenging.
\end{itemize}


\section{Related Work}

\subsection{Benchmarks and Evaluation for Image Editing}
Image editing benchmarks have expanded both task coverage and evaluation
granularity. MagicBrush~\citep{DBLP:conf/nips/ZhangMCSS23} provides
human-annotated editing triplets, while
I$^2$EBench~\citep{DBLP:conf/nips/MaJYLWZZSJ24} introduces multidimensional
evaluation validated against human perception.
ImgEdit~\citep{DBLP:conf/nips/YeHLLYYHY25} and
Complex-Edit~\citep{DBLP:journals/corr/abs-2504-13143} extend evaluation to
multi-turn editing and compositional instruction complexity, respectively.
GIE-Bench~\citep{DBLP:journals/corr/abs-2505-11493} separates functional
correctness from preservation through \abbrreview{visual question answering (VQA)} and object-aware measurements.
LMM4Edit~\citep{DBLP:conf/mm/XuDLMWYGW0MZL25} and factorized VLM
judges~\citep{DBLP:journals/corr/abs-2602-13028} further provide fine-grained
assessment of instruction fidelity, edit quality, and preservation.
What these benchmarks do not provide is deterministic verification of
what changed, where, and by how much against an exact reference.
PaintBench~\citep{DBLP:journals/corr/abs-2606-00188} offers a complementary
approach by evaluating precisely specified operations through deterministic
pixel-level comparison, primarily on procedurally generated shape canvases,
with extensions to charts.
VeriEdit-Bench extends this measurement principle to fine-grained attribute
changes embedded in realistic structured assets.

\subsection{Generation and Editing of Structured Visuals}

Research on structured visuals examines factual correctness, reasoning,
and visual fidelity. StructBench~\citep{DBLP:journals/corr/abs-2510-05091}
evaluates factual correctness through decomposed VLM questions, while
DisciplineGen~\citep{DBLP:journals/corr/abs-2607-02290} studies data
construction and model training for academic and scientific visuals.
PlanViz~\citep{DBLP:journals/corr/abs-2602-06663} targets spatial and
procedural reasoning; VDE Bench~\citep{DBLP:journals/corr/abs-2602-00122}
and OCRGenBench~\citep{zhang2025ocrgenbench} emphasize text modification
and document or structural fidelity. ChartSync~\citep{DBLP:journals/corr/abs-2607-10301}
uses programmatically rendered targets to study coupled changes in chart
values and geometry, combining objective measurements with VLM-based judgments.
FigEdit~\citep{DBLP:journals/corr/abs-2512-00752} shares our specification-to-target rendering
foundation, covering single, composite, and interactive chart edits. It
combines image similarity, text recognition, and model-based evaluation;
our focus is a common deterministic capability profile across asset types.
Table~\ref{tab:app-benchmark-comparison} summarizes
asset coverage, editing operations, and verification in related datasets.

Structured representations also provide an interface for generation and
editing. SVGEditBench V2~\citep{DBLP:journals/corr/abs-2502-19453},
VectorEdits~\citep{DBLP:conf/lrec/KucharKSS26}, and
Vector-Bench~\citep{DBLP:journals/corr/abs-2607-19056} evaluate SVG editing;
Vector-Bench uses deterministic specifications to assess target attributes
and protected structure. DiagramGenBenchmark and
DiagramAgent~\citep{DBLP:conf/cvpr/Wei0CW0GSYG25}, alongside
VCG-Bench~\citep{DBLP:journals/corr/abs-2605-15677}, formulate
diagram generation and refinement through code. 
For presentations, object-model-based editing~\citep{DBLP:conf/acl/JungCYYJC26}
directly modifies structured slide representations, while
I2E~\citep{DBLP:conf/acl/YuXZLLDWJLBZM26} reconstructs an editable environment
from image pixels. These approaches therefore edit through a structured
representation, either given or recovered from pixels. VeriEdit-Bench instead
evaluates editors that operate directly on the rendered image; the source is
used only to construct and score tasks.

\begin{figure}[!ht]
  \centering
  \includegraphics[width=0.95\linewidth]{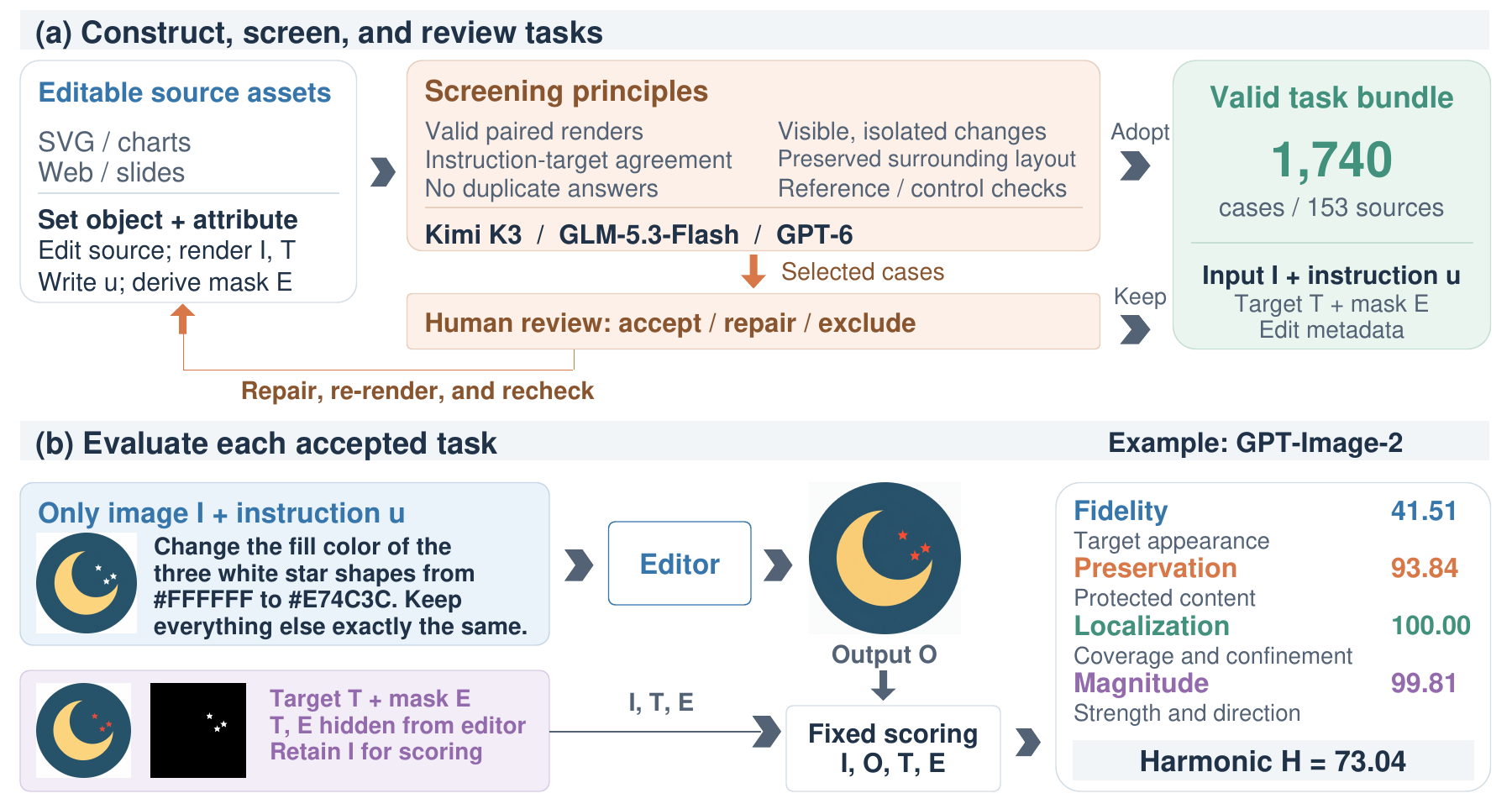}%
  \caption{From editable sources to verifiable evaluation. (a) Source edits,
  \abbrreview{large language model (LLM) screening, and author review curate tasks};
  repairs are re-rendered and rechecked. (b) The editor receives only the input image
  and instruction. \abbrreview{Input, target, edit region, and output yield four
  scores and their harmonic mean, shown for a GPT-Image-2 output.}}
  \label{fig:benchmark-pipeline}
\end{figure}

\section{VeriEdit-Bench}
\label{sec:benchmark}

VeriEdit-Bench measures four complementary aspects of precise editing:
target appearance, preservation of unrelated content, spatial extent, and
visual change magnitude. Each case supplies an input image $I$, instruction
$u$, rendered target $T$, allowed edit region $E$, and source-edit parameters.
The editor receives only $(I,u)$ and returns $O$; the rest serves only to
verify the requested transformation.

\subsection{Dataset Construction}
\label{sec:construction}

\textbf{From editable sources to verifiable tasks.}
We construct tasks from SVG graphics, chart plotting programs, and HTML/CSS
representations of web interfaces and slides. Adapted generation workflows
and custom templates provide these editable assets; their sources and
adaptations are documented in Appendix~\ref{sec:app-asset-sources}.
A controlled source edit fixes the requested attribute values and object
scope. Rendering before and after the edit under identical settings gives
$I$ and $T$, while the instruction describes the visible objects, requested
changes, and preservation requirements. Requested values include
hexadecimal colors, spatial offsets, and multiplicative changes such as
$1.5\times$ font size. The controlled source edit and fixed rendering
settings jointly determine the visual reference. Edits may affect
multiple objects, as in palette replacement. Tasks cover attribute edits of
existing visible content, including appearance, visibility, and position;
open-ended generation and content insertion are outside the benchmark scope.

\textbf{Rendered change defines the allowed region.}
With $d_{76}$ the CIELAB Euclidean distance, we define
\begin{equation}
  E(x)=\mathbf{1}\!\left[d_{76}(I,T)(x)>2\right].
  \label{eq:edit-region}
\end{equation}
The region includes vacated and newly occupied pixels when an object moves;
unchanged pixels remain protected. Its small threshold suppresses negligible
rendering differences.
 
\textbf{Curation for valid measurement.}
Each candidate must satisfy four checks: matching canvas dimensions,
a visible requested change, preserved surrounding layout, and agreement
between the instruction and the edited objects and parameters.
LLM-assisted screening flags potential violations; all retained tasks are
manually reviewed. Identified
issues are resolved through source-level checks and verification of the
rendered outcomes (Appendix~\ref{sec:app-human-verification}).

\label{sec:dataset-statistics}
\begin{figure}[ht]
  \centering
  \includegraphics[width=0.95\linewidth]{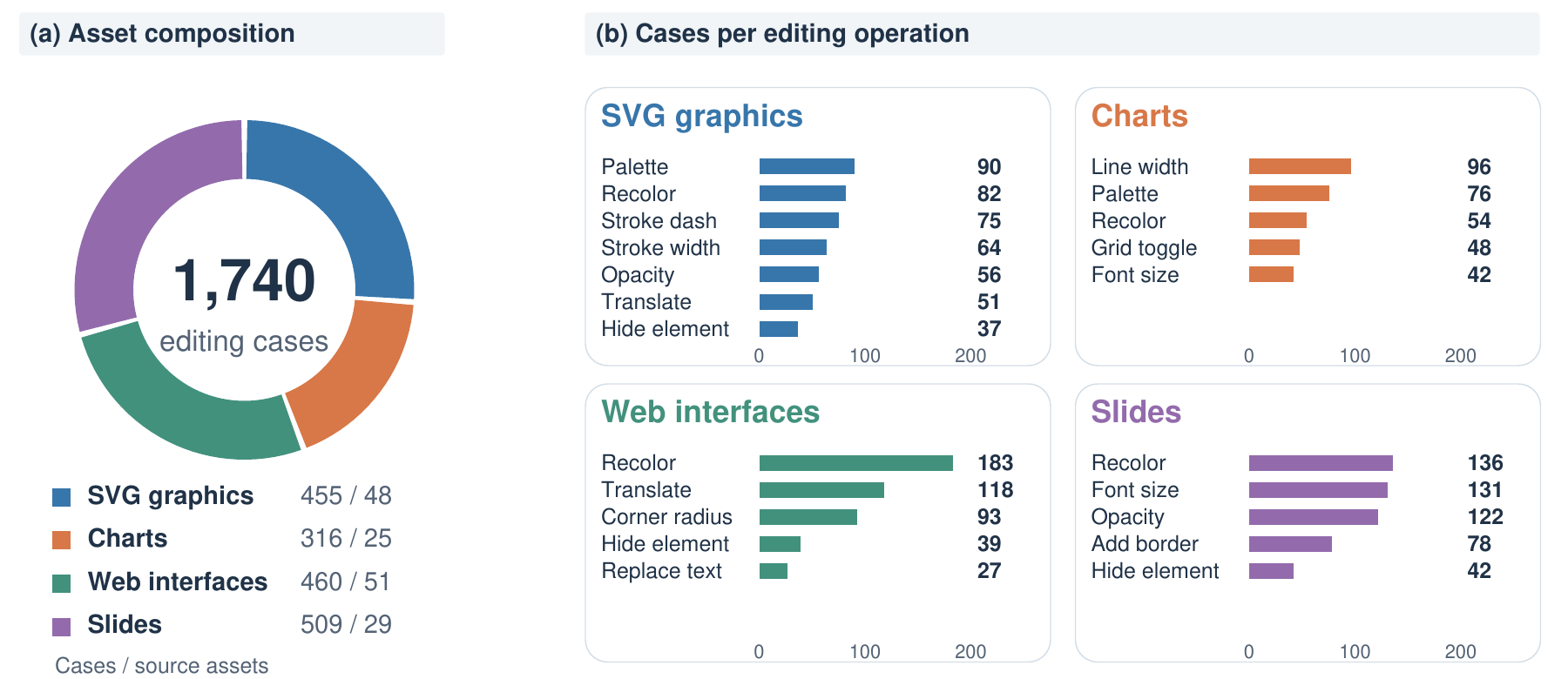}%
  \caption{Dataset composition. Left: case shares and case/source counts. Right: operation counts, sorted within each asset type on a shared scale; instruction variants are grouped by operation.}
  \label{fig:dataset-statistics}
\end{figure}

\subsection{Benchmark Composition}
The benchmark contains 1,740 cases from 153 sources: 48 SVG graphics,
25 charts, 51 web interfaces, and 29 slides. Each case pairs one source
asset with one controlled edit specification, including the operation,
affected object scope, and target parameter values; variants of the same
operation differ in the requested edit values. Figure~\ref{fig:dataset-statistics}
shows their distribution across 22 asset--operation combinations.
Asset-specific operations cover stroke styles, chart grids, web corner
radii, and slide borders; shared operations such as recoloring support
comparisons across asset types with the operation held fixed. Cases share
source assets, with about eleven cases per source on average; accordingly,
Sec.~4 resamples sources rather than individual cases.

\subsection{Evaluation Protocol}
\label{sec:evaluation}

Figure~\ref{fig:benchmark-pipeline}(b) shows the four-axis evaluation.
Images are compared on the target canvas without content-based registration.
All scores use fixed pixel-based rules and lie in $[0,100]$.

For region $R$ and color tolerances $\mathcal K$, define
\begin{equation}
 Q_{\mathcal{K}}(A,B;R)=
 \frac{1}{|\mathcal{K}|}\sum_{t\in\mathcal{K}}
 \frac{1}{|R|}\sum_{x\in R}\mathbf{1}[d_{76}(A,B)(x)\le t].
 \label{eq:color-agreement}
\end{equation}
This averages color-agreement rates to provide graded credit.
A shared binary check $g$ screens for apparent edits explained by negligible
change, noise, blur, or global color transformations. Failure sets fidelity,
localization, and magnitude to zero while leaving preservation independently
measured; passing does not establish agreement with the requested target.
Preprocessing, checks, and boundary cases are described in
Appendix~\ref{sec:app-scoring-details}.

\textbf{Fidelity.}
\label{sec:metric-fidelity}
Fidelity measures improvement toward the target appearance in the edited
region. After mild Gaussian smoothing and one-pixel mask dilation, let
$b=Q_{\mathcal K_F}(\widetilde I,\widetilde T;E^+)$ be the unchanged-input
baseline, with $\mathcal K_F=\{5,10,20\}$. We compute
\begin{equation}
 F=100g\,\operatorname{clip}_{[0,1]}
 \left(\frac{Q_{\mathcal K_F}(\widetilde O,\widetilde T;E^+)-b}{1-b}\right).
 \label{eq:fidelity}
\end{equation}
Baseline subtraction prevents unchanged pixels from earning edit credit.

\textbf{Preservation.}
\label{sec:metric-preservation}
Preservation measures agreement with the input outside the allowed region:
$P=100Q_{\{2,5,10,20\}}(O,I;\overline E)$, using unblurred images and the
original mask. Thus, unrelated content can be preserved well even when the
requested edit is executed poorly.

\textbf{Localization.}
\label{sec:metric-localization}
Localization measures coverage of the intended change and confinement to
the allowed region. Let $G$ and $D$ be thresholded reference and output
change maps, with small background components removed from $D$. Spatial
recall and precision give
\begin{equation}
 r_L=\frac{|D\cap G|}{|G|},\qquad
 p_L=\frac{|D\cap E|}{|D\cap E|+\min(|D\setminus E|,4|E|)},\qquad
 L=100g\,\frac{2r_Lp_L}{r_L+p_L}.
 \label{eq:localization}
\end{equation}
\submissionreview{Capping outside-region area retains sensitivity to target
coverage when the intended edit is small; preservation measures agreement
over the entire protected region.}

\textbf{Magnitude.}
\label{sec:metric-magnitude}
Magnitude compares the strength and direction of visual change on the
target's effective-change core $C\subseteq E$. Let $r_M$ be the ratio of
output to target Lab change amplitudes from $I$, and $c$ the cosine between their per-pixel Lab change fields. An amplitude agreement $a(r_M)$ gives full credit
within 15\% of the target and decays linearly outside this band
(Appendix~\ref{sec:app-scoring-details}):
\begin{equation}
 M=100g\left[\tfrac12 a(r_M)+\tfrac12\frac{c+1}{2}\right].
 \label{eq:magnitude}
\end{equation}
It measures visual change relative to the reference rather than a source
parameter such as font size.

\textbf{Capability profiles and composite score.}
The capability profile $(F,P,L,M)$ records performance on each dimension.
Their harmonic mean
$H=4/(F^{-1}+P^{-1}+L^{-1}+M^{-1})$
summarizes joint performance, with $H=0$ if any axis is zero.
$H$ is the default summary of joint performance, while the individual axes
retain partial credit and diagnose specific aspects of the output. For example,
an unchanged input scores $(0,100,0,0)$.

\section{Experiments}
\label{sec:experiments}


\begin{table}[t]
\centering
\setlength{\belowcaptionskip}{4pt}
\caption{Performance on 1,740 cases (0--100, higher is better). \abbrreview{Models are columns; the four axes are case-weighted means.} Cell-mean $H$ averages per-case harmonic composites within each of the 22 asset--operation cells and weights cells equally. \abbrreview{CI denotes the 95\% bootstrap confidence interval for cell-mean $H$.} \submissionreview{\textbf{Bold}: best overall; \underline{underlined}: best open-weight, per metric.}}
\label{tab:exp-main}
\begingroup\scriptsize
\setlength{\tabcolsep}{2.0pt}\renewcommand{\arraystretch}{1.1}
\begin{tabular}{@{}l*{11}{r}@{}}
\noalign{\hrule height 0.9pt}
Metric & \shortstack{GPT-\\Image-2} & \shortstack{FireRed-\\Edit} & \draft{\shortstack{Qwen-Image-\\Edit-2511}} & \draft{\shortstack{Qwen-Image-\\Edit}} & \shortstack{FLUX.2-\\dev} & BAGEL & \draft{\shortstack{Step1X-\\Edit}} & \shortstack{FLUX.1-\\Kontext} & \draft{\shortstack{FLUX.2-\\klein}} & OmniGen2 & \shortstack{Instruct\\Pix2Pix} \\
\noalign{\vskip 2pt\hrule height 0.45pt\vskip 2pt}
Fidelity $F$ & \datasetareview{\textbf{37.10}} & \datasetareview{\underline{13.42}} & \datasetareview{12.22} & \datasetareview{9.38} & \datasetareview{8.16} & \datasetareview{6.66} & \datasetareview{5.24} & \datasetareview{3.78} & \datasetareview{4.85} & \datasetareview{5.16} & \datasetareview{2.36} \\
Preservation $P$ & \datasetareview{84.52} & \datasetareview{85.09} & \datasetareview{87.16} & \datasetareview{82.06} & \datasetareview{82.61} & \datasetareview{68.75} & \datasetareview{82.78} & \datasetareview{\textbf{\underline{91.64}}} & \datasetareview{78.81} & \datasetareview{60.52} & \datasetareview{39.19} \\
Localization $L$ & \datasetareview{\textbf{56.92}} & \datasetareview{44.29} & \datasetareview{\underline{46.64}} & \datasetareview{44.00} & \datasetareview{43.30} & \datasetareview{42.38} & \datasetareview{40.09} & \datasetareview{28.16} & \datasetareview{38.81} & \datasetareview{35.52} & \datasetareview{42.30} \\
Magnitude $M$ & \datasetareview{\textbf{84.09}} & \datasetareview{\underline{68.76}} & \datasetareview{66.92} & \datasetareview{60.53} & \datasetareview{61.65} & \datasetareview{61.38} & \datasetareview{59.47} & \datasetareview{46.69} & \datasetareview{60.34} & \datasetareview{56.80} & \datasetareview{64.16} \\
\noalign{\vskip 2pt\hrule height 0.45pt\vskip 2pt}
Cell-mean $H$ & \datasetareview{\textbf{47.13}} & \datasetareview{\underline{21.85}} & \datasetareview{20.47} & \datasetareview{16.26} & \datasetareview{13.11} & \datasetareview{10.81} & \datasetareview{9.78} & \datasetareview{8.78} & \datasetareview{8.42} & \datasetareview{7.32} & \datasetareview{3.83} \\
CI lower & \datasetareview{44.95} & \datasetareview{20.28} & \datasetareview{18.97} & \datasetareview{14.85} & \datasetareview{11.68} & \datasetareview{9.49} & \datasetareview{8.44} & \datasetareview{7.63} & \datasetareview{7.21} & \datasetareview{6.11} & \datasetareview{3.09} \\
CI upper & \datasetareview{49.05} & \datasetareview{23.29} & \datasetareview{21.85} & \datasetareview{17.64} & \datasetareview{14.44} & \datasetareview{12.28} & \datasetareview{11.20} & \datasetareview{10.00} & \datasetareview{9.69} & \datasetareview{8.56} & \datasetareview{4.69} \\
\noalign{\vskip 2pt\hrule height 0.9pt}
\end{tabular}\endgroup
\end{table}

\subsection{Experimental Setup}
\label{sec:exp-setup}

We evaluate eleven editors. Ten open-weight models:
FireRed-Edit~\citep{DBLP:journals/corr/abs-2602-13344},
FLUX.2-dev and FLUX.2-klein~\citep{blackforestlabs2025flux2},
FLUX.1-Kontext~\citep{DBLP:journals/corr/abs-2506-15742},
Qwen-Image-Edit and Qwen-Image-Edit-2511~\citep{DBLP:journals/corr/abs-2508-02324}\footnote{\raggedright Qwen-Image-Edit-2511 model and configuration: \url{https://huggingface.co/Qwen/Qwen-Image-Edit-2511}.},
BAGEL~\citep{DBLP:journals/corr/abs-2505-14683},
Step1X-Edit~\citep{DBLP:journals/corr/abs-2504-17761},
OmniGen2~\citep{DBLP:journals/corr/abs-2506-18871},
and InstructPix2Pix~\citep{DBLP:conf/cvpr/BrooksHE23};
GPT-Image-2~\citep{openai2026chatgptimages2} provides a closed-source comparison.

We evaluate one output per model and case using model-specific inference
settings. Images are centered on a white square canvas with their aspect
ratio preserved; outputs are mapped back to the reference canvas for
scoring. Appendix~\ref{sec:app-inference} specifies the checkpoints,
processing resolutions, and coordinate mapping. All analyses use the same
model outputs.
The primary score, cell-mean $H$, averages per-case composites within each
asset--operation cell and weights the 22 cells equally; individual-axis
entries in Table~\ref{tab:exp-main} average cases equally. Confidence
intervals for $H$ come from 10,000 source-level bootstrap resamples within
each asset type, keeping all operations of a source together and sharing
draws across models, reported as 95\% bias-corrected and accelerated (BCa)
intervals (Appendix~\ref{sec:app-source-statistics}). Seed and
instruction-wording controls are reported in Appendix~\ref{sec:app-controls}.

\textbf{Scorer validation.}
We check that each axis responds to the error it is designed to detect.
On all 1,740 cases, the rendered target scores $(100,100,100,100)$ and the
unchanged input $(0,100,0,0)$. \submissionreview{Controlled perturbation results appear in
Appendix~\ref{sec:app-controlled}.}
Constructed source-level editing errors give complementary mean
responses: omitting one
color group lowers $F$ to 49 and $L$ to 70 while preservation stays at 100;
editing an extra group keeps $F=96$ but lowers $P$ to 72; and swapping
target colors keeps \mbox{$L\approx100$} but drops $F$ to 3. Each failure mode is
thus visible on the axis that measures it.

\subsection{Main Results}
\label{sec:exp-main}

\textbf{Precise editing remains challenging.}
Table~\ref{tab:exp-main} ranks GPT-Image-2 first with a cell-mean $H$ of
\datasetareview{47.1}, more than twice the best open-weight model, FireRed-Edit (\datasetareview{21.9}),
followed by Qwen-Image-Edit-2511 and Qwen-Image-Edit.
Paired bootstrap intervals separate every adjacent pair among the six
leading models and OmniGen2 from InstructPix2Pix, whereas the four adjacent
comparisons between BAGEL and OmniGen2 remain unresolved (Appendix
Table~\ref{tab:app-paired-gaps}); a case-weighted comparison appears in
Appendix~\ref{sec:app-case-weighting}. Composites are limited mainly by
fidelity: open-weight $F$ ranges from \datasetareview{2.4 to 13.4} while $P$ exceeds 78 for
all but three models, and between \datasetareview{53\%} and 85\% of open-weight outputs
receive $H=0$. Across all models, the zero-composite breakdown shows that
87\% arise from outputs that pass the eligibility check and change the
region measurably ($L,M>0$) yet do not raise agreement with the target
above the unchanged-input baseline; the check itself accounts for 11\% (Appendix
Table~\ref{tab:app-zero-sources}). The benchmark
thus separates substantial gaps from uncertain adjacent rankings, while
even the strongest editor scores below half of the attainable composite.

\textbf{Capability depends on the requested operation.}
The operation breakdown (Appendix Figure~\ref{fig:exp-operations}) shows
that \draft{FireRed-Edit} edits web visibility well but struggles with corner radii;
\draft{GPT-Image-2} likewise combines strong SVG recoloring with weak web corner-radius
control. Both perform less well on chart font-size edits than on their
strongest operations. Appendix~\ref{sec:app-axis} gives the corresponding
four-axis profiles. Comparing the two Qwen-Image-Edit checkpoints likewise
reveals larger gains for chart line width and web visibility than for
slide typography or web corner radii
(Appendix~\ref{sec:app-qwen-versions}). 
Overall scores alone obscure operation-specific strengths and gains
across model versions.

\subsection{Same Operation, Different Assets}
\label{sec:exp-assets}
\textbf{Setup.}
To test whether performance on the same editing operation generalizes
across asset types, we evaluate recoloring on 455 cases spanning SVG
graphics, charts, web interfaces, and slides. We report case-mean $H$ for
each asset type with 95\% confidence intervals obtained by resampling
source assets (Appendix~\ref{sec:app-shared-operation}).

\begin{figure}[!htb]
  \centering
  \includegraphics[width=0.95\linewidth]{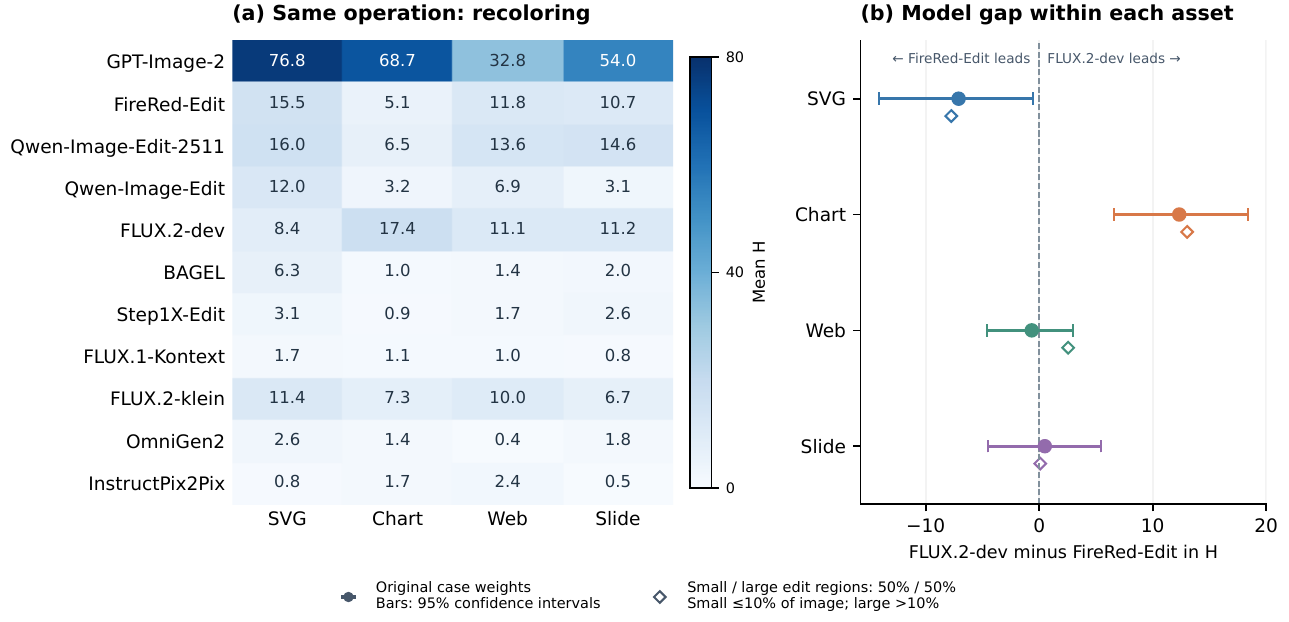}\par\nointerlineskip
  \caption{\datasetareview{Fixed-operation comparison on 455 recoloring cases and 5,005
  outputs from Table~\ref{tab:exp-main}. Left: all eleven models' mean $H$ by asset type. The color bar shows the shared score scale. Right:
  the FLUX.2-dev--FireRed-Edit gap with source-group bootstrap intervals;
  hollow diamonds give the gap after equally weighting edits covering
  at most 10\% and more than 10\% of the image.}}
  \label{fig:exp-assets}
\end{figure}

\textbf{Results and analysis.}
Figure~\ref{fig:exp-assets} shows a ranking reversal: FLUX.2-dev outperforms
\draft{FireRed-Edit} on charts but trails it on SVG. Both gap intervals exclude zero,
and equal weighting of small and large edit regions preserves the reversal.
\draft{GPT-Image-2} also performs better on charts than on web interfaces. Although chart
rankings broadly correlate with other asset rankings, they miss these
pairwise and absolute differences. Appendix~\ref{sec:app-shared-operation}
gives further cross-asset comparisons; Appendix~\ref{sec:app-area} examines
spatial precision across edit-region sizes.
A chart result cannot substitute for measuring the same attribute in other
structured assets. These differences characterize the observed collections;
they do not isolate layout complexity as a cause.

\begin{figure}[!htb]
  \centering
  \includegraphics[width=\linewidth]{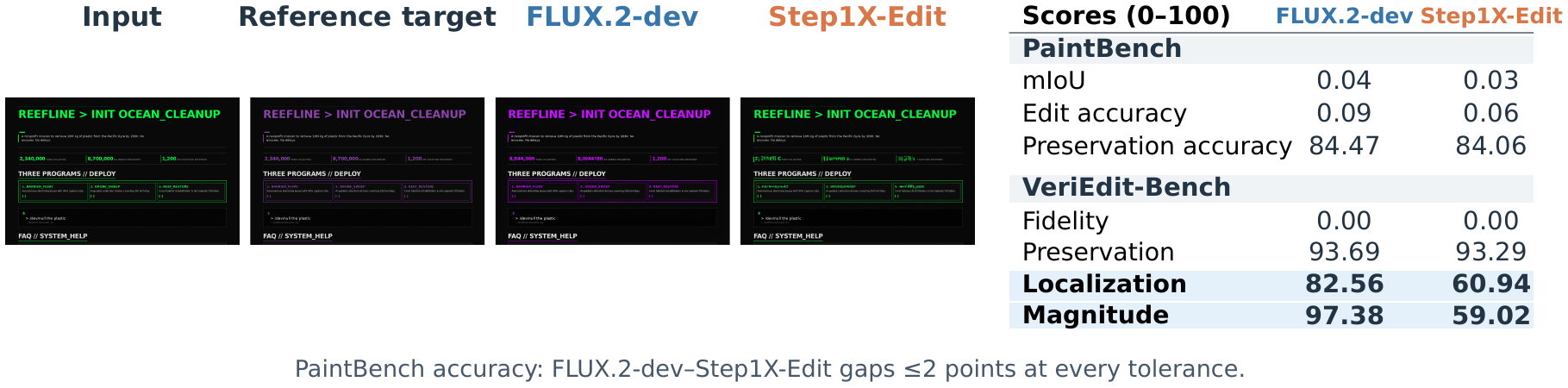}\par\nointerlineskip
  \caption{\datasetareview{Complete images and per-output scores. PaintBench rows average CIE76 tolerances 0--10; shading highlights localization and magnitude differences. Larger views and full accuracy curves appear in Appendix Figures~\ref{fig:app-paintbench-example} and~\ref{fig:app-paintbench-curves}.}}
  \label{fig:exp-paintbench-matched}
\end{figure}

\subsection{Distinct Failures behind Matched Pixel-Accuracy Profiles}
\label{sec:exp-diagnosis}
\textbf{Setup.}
PaintBench \abbrreview{deterministically measures target-pixel correctness
inside the edit region and preservation outside it against a known reference,
using mean intersection over union (mIoU) and accuracy curves across color tolerances.} Our profile additionally quantifies where and how much the
output changes relative to the input, separately from target appearance.
To test whether these measurements expose failures that pixel accuracy
alone does not, we score the same outputs of all eleven models with both
scorers on a 607-case panel (one case per source--operation group, all 153
sources) and select same-case model pairs whose PaintBench mIoU differs by
at most one point and whose two accuracy curves differ by at most two
points at every tolerance
(Appendix~\ref{sec:app-paintbench-matched}).

\textbf{Results and analysis.}
Of 33,385 same-case pairs, \datasetareview{416} satisfy this matching rule, and \datasetareview{223} of them
differ by at least 20 points in $L$ or $M$, spanning \datasetareview{68 cases and 56 sources}
(Appendix Table~\ref{tab:app-pb-607}).
Figure~\ref{fig:exp-paintbench-matched} illustrates the distinction with a
matched pair from the 101-case diagnostic panel
(Appendix~\ref{sec:app-paintbench-matched}):
\exampleswap{FLUX.2-dev recolors the requested page elements to a
magenta-shifted purple rather than the requested \#8e44ad, while Step1X-Edit
leaves the target pixels green and repaints unrelated page regions.}
\datasetareview{Both receive zero fidelity.}
\exampleswap{FLUX.2-dev better matches the reference change
magnitude and localizes its changes more precisely.}
\datasetareview{The axis profiles distinguish these errors behind similar target-pixel
accuracy.} Further examples across asset
types appear in Appendix Figures~\ref{fig:app-example-1}--\ref{fig:app-example-4}.
Separately measuring target agreement and edit execution reveals which
capability needs improvement, even when pixel-accuracy profiles nearly
coincide.

\subsection{Independent Analysis of Color Attainment}
\label{sec:exp-color}

\begin{figure}[!htbp]
  \centering
  \includegraphics[width=0.95\linewidth]{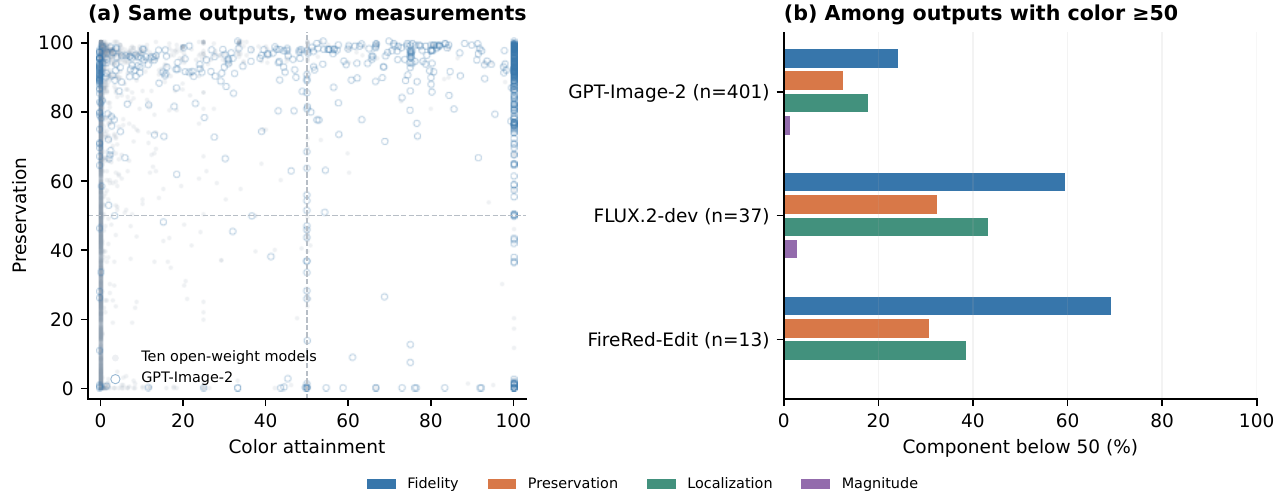}\par\nointerlineskip
  \caption{\datasetareview{Attribute execution and precise editing on identical outputs.
  Left: all 6,831 eligible outputs, showing color attainment against
  preservation. Transparency reduces overlap; hollow points mark GPT-Image-2. Right: component scores below 50 among the color-score
  $\ge50$ outputs of GPT-Image-2, FLUX.2-dev, and FireRed-Edit; the
  denominators are printed beside their names. Components can overlap. Appendix~\ref{sec:app-color-diagnosis}
  reports every model and alternative color cutoffs.}}
  \label{fig:exp-color-diagnosis}
\end{figure}

\textbf{Setup.}
The preceding analysis separates different failures behind similar
pixel-accuracy profiles. We now ask whether strong attainment of a requested
attribute can still coexist with other editing failures. We use color because
color-editing tasks span all four asset types and target-color attainment is
directly measurable. We check whether the same outputs also preserve unrelated
content and confine changes to the intended region.
For 621 eligible cases per model, CIEDE2000 measures target-color attainment
within stable object regions (Appendix~\ref{sec:app-color-measurement}). This
diagnostic is separate from the four axes and does not enter $H$.
We examine axis scores below 50 when color attainment is at least 50.
These thresholds define descriptive slices of continuous scores, not a new
success criterion.

\begin{table}[t]
\centering
\setlength{\belowcaptionskip}{4pt}
\caption{Independent color-attainment experiment on 621 color-editing cases. Values average the covered asset--operation cells (0--100). Color attainment is not a component of the four-axis composite. Model order matches Table~\ref{tab:exp-main}; \textbf{bold}: best model; \underline{underlined}: best open-weight model.}
\label{tab:exp-color}
\begingroup\scriptsize
\setlength{\tabcolsep}{2.0pt}\renewcommand{\arraystretch}{1.1}
\begin{tabular}{@{}l*{11}{r}@{}}
\noalign{\hrule height 0.9pt}
Metric & \shortstack{GPT-\\Image-2} & \shortstack{FireRed-\\Edit} & \draft{\shortstack{Qwen-Image-\\Edit-2511}} & \draft{\shortstack{Qwen-Image-\\Edit}} & \shortstack{FLUX.2-\\dev} & BAGEL & \draft{\shortstack{Step1X-\\Edit}} & \datasetareview{\shortstack{FLUX.1-\\Kontext}} & \datasetareview{\shortstack{FLUX.2-\\klein}} & OmniGen2 & \shortstack{Instruct\\Pix2Pix} \\
\noalign{\vskip 2pt\hrule height 0.45pt\vskip 2pt}
Color attainment & \datasetareview{\textbf{64.92}} & \datasetareview{2.98} & \datasetareview{4.65} & \datasetareview{2.25} & \datasetareview{\underline{6.41}} & \datasetareview{1.00} & \datasetareview{0.85} & \datasetareview{0.26} & \datasetareview{3.40} & \datasetareview{0.60} & \datasetareview{0.45} \\
\noalign{\vskip 2pt\hrule height 0.9pt}
\end{tabular}\endgroup
\end{table}

\textbf{Results and analysis.}
GPT-Image-2 leads color attainment (Table~\ref{tab:exp-color}), whereas every
open-weight model scores below 7. The criterion is strict: a core pixel
counts only within a CIEDE2000 distance of at most 5 of the requested color
(Appendix~\ref{sec:app-color-measurement}). Open-weight editors do alter the
target regions, with nonzero localization and magnitude on
\datasetareview{73--98\%} of these outputs, yet 63--85\% of them receive zero color
attainment (Appendix Table~\ref{tab:app-color-execution}): the requested
color is rarely reached. Among GPT-Image-2's outputs meeting the color
threshold, \datasetareview{12.5\%} have low preservation and \datasetareview{17.7\%} have low localization
(Figure~\ref{fig:exp-color-diagnosis}). The pattern persists at a color
cutoff of 80 and also occurs for FLUX.2-dev; complete counts and uncertainty
estimates appear in Appendix~\ref{sec:app-color-diagnosis}.
Strong color attainment therefore does not by itself establish control over
the full edit. Measuring it alongside preservation and localization reveals
whether progress toward the requested attribute damages the existing
design.

\subsection{Aggregation and Scoring Sensitivity}
\label{sec:exp-ablation}

\begin{figure}[!htb]
  \centering
  \includegraphics[width=0.95\linewidth]{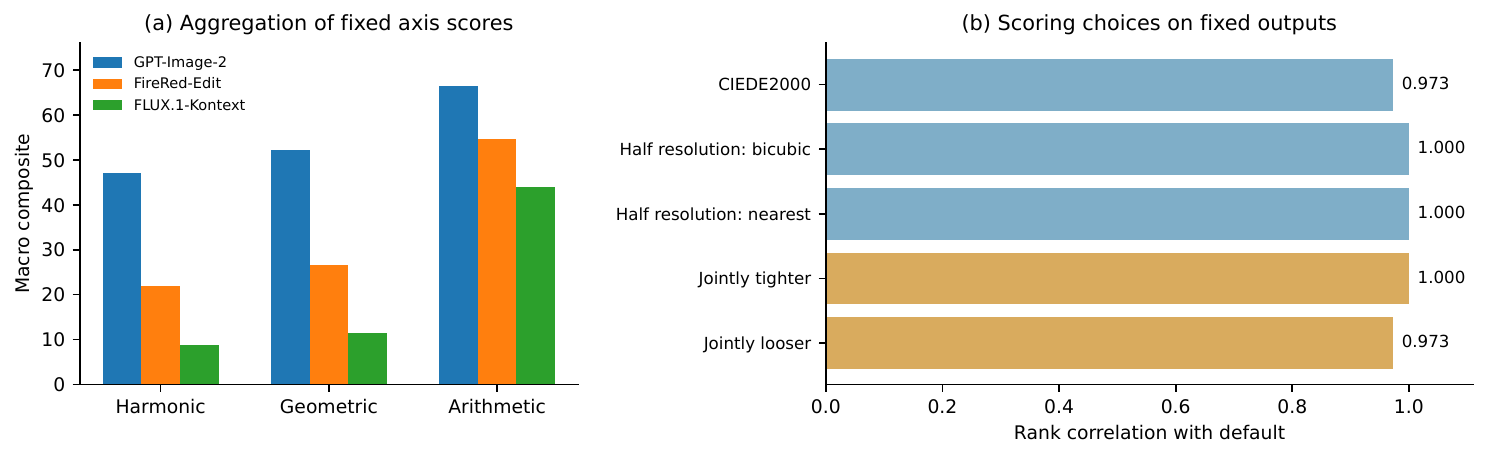}\par\nointerlineskip
  \caption{\datasetareview{Fixed-output ablations. Left: changing the aggregation of identical
  axis scores changes the reward for unbalanced edits. Right: rank correlation
  across all eleven models after changing the scoring configuration. Complete
  model scores and single-parameter sensitivity are in Appendix~\ref{sec:app-ablation}.}}
  \label{fig:exp-ablation}
\end{figure}

\textbf{Aggregation changes what the summary rewards.}
We compare harmonic, geometric, and arithmetic means of each output's
four axes, followed by the same cell weighting
(Figure~\ref{fig:exp-ablation}). Arithmetic aggregation substantially raises
scores for unbalanced edits, particularly for \draft{FLUX.1-Kontext}, because strong
components offset weak ones. The harmonic composite emphasizes joint
performance, while the axis profile retains partial strengths.

\textbf{The ranking is robust to scoring choices.}
Across the 44 single-parameter settings, the full eleven-model ranking
remains unchanged in \datasetareview{40} settings. The remaining \datasetareview{4} settings change
only lower-ranked positions; the leading five models never move. These
variations concern close default scores whose adjacent comparisons are
statistically unresolved (Appendix Table~\ref{tab:app-paired-gaps}). Appendix~\ref{sec:app-ablation} reports all settings and
model scores.

\section{Conclusion}

We introduced \textbf{VeriEdit-Bench} to bridge the gap between realism
and verifiability in fine-grained image editing.
Controlled source-code edits across SVG graphics, charts, web interfaces,
and slides yield exact targets, edit masks, and specifications for
deterministic evaluation of fidelity, preservation, localization, and
visual change magnitude.
Our evaluation of eleven editors shows that precise editing remains
challenging. Similar pixel-accuracy profiles can conceal distinct failures,
strong target-color attainment can coexist with poor preservation or
localization, and model advantages can reverse across asset types for the
same recoloring operation.
VeriEdit-Bench supports reproducible diagnosis of capabilities and the
contexts in which gains hold, beyond aggregate scores.
\pagebreak[4]



\bibliography{iclr2027_conference}

@article{zhang2025ocrgenbench,
  title={OCRGenBench: A Comprehensive Benchmark for Evaluating OCR Generative Capabilities},
  author={Zhang, Peirong and Xu, Haowei and Zhang, Jiaxin and Zheng, Xuhan and Xu, Guitao and Zhang, Yuyi and Liu, Junle and Yang, Zhenhua and Zhou, Wei and Jin, Lianwen},
  journal={arXiv preprint arXiv:2507.15085},
  year={2025}
}

@article{han2025unireditbench,
  title={Unireditbench: A unified reasoning-based image editing benchmark},
  author={Han, Feng and Wang, Yibin and Li, Chenglin and Liang, Zheming and Wang, Dianyi and Jiao, Yang and Wei, Zhipeng and Gong, Chao and Jin, Cheng and Wang, Jiaqi},
  journal={arXiv preprint arXiv:2511.01295},
  year={2025}
}

@article{kimi2026k3,
  title={{Kimi k3}: Open frontier intelligence},
  author={{Kimi Team} and Bai, Tongtong and Bai, Yifan and Bao, Yiping and Cai, Jianfeng and Cai, Xinyuan and Cao, Peizhou and Cao, Yuxuan and Chai, Ziwei and Charles, Y and others},
  journal={arXiv preprint arXiv:2607.24653},
  year={2026}
}

@misc{zai2026glm53flash,
  title  = {{GLM-5.3-Flash}: Frontier Intelligence, Flash Cost},
  author = {{Z.ai}},
  year   = {2026},
  month  = aug,
  url    = {https://z.ai/blog/glm-5.3-flash}
}

@misc{openai2026gpt6astra,
  title  = {{GPT-6 Astra} System Card},
  author = {{OpenAI}},
  year   = {2026},
  month  = sep,
  url    = {https://deploymentsafety.openai.com/gpt-6-astra}
}

@article{kumar2026vlm,
  title={VLM Judges Can Rank but Cannot Score: Task-Dependent Uncertainty in Multimodal Evaluation},
  author={Kumar, Divake and Tayebati, Sina and Naik, Devashri and Krishnan, Ranganath and Trivedi, Amit Ranjan},
  journal={arXiv preprint arXiv:2604.25235},
  year={2026}
}

@inproceedings{DBLP:conf/acl/YuXZLLDWJLBZM26,
  author       = {Jinghan Yu and
                  Junhao Xiao and
                  Chenyu Zhu and
                  Jiaming Li and
                  Jia Li and
                  HanMing Deng and
                  Xirui Wang and
                  Guoli Jia and
                  Jianjun Li and
                  Xiang Bai and
                  Bowen Zhou and
                  Zhiyuan Ma},
  editor       = {Maria Liakata and
                  Viviane P. Moreira and
                  Jiajun Zhang and
                  David Jurgens},
  title        = {{I2E:} From Image Pixels to Actionable Interactive Environments for
                  Text-Guided Image Editing},
  booktitle    = {Proceedings of the 64th Annual Meeting of the Association for Computational
                  Linguistics (Volume 1: Long Papers), {ACL} 2026, San Diego, California,
                  United States, July 2-7, 2026},
  pages        = {44827--44850},
  publisher    = {Association for Computational Linguistics},
  year         = {2026},
  url          = {https://doi.org/10.18653/v1/2026.acl-long.2076},
  doi          = {10.18653/V1/2026.ACL-LONG.2076},
  bibsource    = {dblp computer science bibliography, https://dblp.org}
}

@article{DBLP:journals/corr/abs-2512-00752,
  author       = {Shawn Li and
                  Ryan A. Rossi and
                  Sungchul Kim and
                  Sunav Choudhary and
                  Franck Dernoncourt and
                  Puneet Mathur and
                  Zhengzhong Tu and
                  Yue Zhao},
  title        = {Charts Are Not Images: On the Challenges of Scientific Chart Editing},
  journal      = {CoRR},
  volume       = {abs/2512.00752},
  year         = {2025},
  url          = {https://doi.org/10.48550/arXiv.2512.00752},
  doi          = {10.48550/ARXIV.2512.00752},
  eprinttype   = {arXiv},
  eprint       = {2512.00752},
  bibsource    = {dblp computer science bibliography, https://dblp.org}
}

@inproceedings{DBLP:conf/cvpr/WuS025,
  author       = {Ronghuan Wu and
                  Wanchao Su and
                  Jing Liao},
  title        = {Chat2SVG: Vector Graphics Generation with Large Language Models and
                  Image Diffusion Models},
  booktitle    = {{IEEE/CVF} Conference on Computer Vision and Pattern Recognition,
                  {CVPR} 2025, Nashville, TN, USA, June 11-15, 2025},
  pages        = {23690--23700},
  publisher    = {Computer Vision Foundation / {IEEE}},
  year         = {2025},
  url          = {https://openaccess.thecvf.com/content/CVPR2025/html/Wu\_Chat2SVG\_Vector\_Graphics\_Generation\_with\_Large\_Language\_Models\_and\_Image\_CVPR\_2025\_paper.html},
  doi          = {10.1109/CVPR52734.2025.02206},
  bibsource    = {dblp computer science bibliography, https://dblp.org}
}

@inproceedings{DBLP:conf/iclr/0002SLS0JXZLZLN25,
  author       = {Cheng Yang and
                  Chufan Shi and
                  Yaxin Liu and
                  Bo Shui and
                  Junjie Wang and
                  Mohan Jing and
                  Linran Xu and
                  Xinyu Zhu and
                  Siheng Li and
                  Yuxiang Zhang and
                  Gongye Liu and
                  Xiaomei Nie and
                  Deng Cai and
                  Yujiu Yang},
  title        = {ChartMimic: Evaluating LMM's Cross-Modal Reasoning Capability via
                  Chart-to-Code Generation},
  booktitle    = {The Thirteenth International Conference on Learning Representations,
                  {ICLR} 2025, Singapore, April 24-28, 2025},
  publisher    = {OpenReview.net},
  year         = {2025},
  url          = {https://openreview.net/forum?id=sGpCzsfd1K},
  bibsource    = {dblp computer science bibliography, https://dblp.org}
}

@article{DBLP:journals/corr/abs-2311-16483,
  author       = {Yucheng Han and
                  Chi Zhang and
                  Xin Chen and
                  Xu Yang and
                  Zhibin Wang and
                  Gang Yu and
                  Bin Fu and
                  Hanwang Zhang},
  title        = {ChartLlama: {A} Multimodal {LLM} for Chart Understanding and Generation},
  journal      = {CoRR},
  volume       = {abs/2311.16483},
  year         = {2023},
  url          = {https://doi.org/10.48550/arXiv.2311.16483},
  doi          = {10.48550/ARXIV.2311.16483},
  eprinttype   = {arXiv},
  eprint       = {2311.16483},
  bibsource    = {dblp computer science bibliography, https://dblp.org}
}

@misc{openai2026chatgptimages2,
  author       = {{OpenAI}},
  title        = {ChatGPT Images 2.0 System Card},
  year         = {2026},
  howpublished = {\url{https://deploymentsafety.openai.com/chatgpt-images-2-0}}
}

@inproceedings{DBLP:conf/cvpr/BrooksHE23,
  author       = {Tim Brooks and
                  Aleksander Holynski and
                  Alexei A. Efros},
  title        = {InstructPix2Pix: Learning to Follow Image Editing Instructions},
  booktitle    = {{IEEE/CVF} Conference on Computer Vision and Pattern Recognition,
                  {CVPR} 2023, Vancouver, BC, Canada, June 17-24, 2023},
  pages        = {18392--18402},
  publisher    = {{IEEE}},
  year         = {2023},
  url          = {https://doi.org/10.1109/CVPR52729.2023.01764},
  doi          = {10.1109/CVPR52729.2023.01764},
  bibsource    = {dblp computer science bibliography, https://dblp.org}
}

@article{DBLP:journals/corr/abs-2506-18871,
  author       = {Chenyuan Wu and
                  Pengfei Zheng and
                  Ruiran Yan and
                  Shitao Xiao and
                  Xin Luo and
                  Yueze Wang and
                  Wanli Li and
                  Xiyan Jiang and
                  Yexin Liu and
                  Junjie Zhou and
                  Ze Liu and
                  Ziyi Xia and
                  Chaofan Li and
                  Haoge Deng and
                  Jiahao Wang and
                  Kun Luo and
                  Bo Zhang and
                  Defu Lian and
                  Xinlong Wang and
                  Zhongyuan Wang and
                  Tiejun Huang and
                  Zheng Liu},
  title        = {OmniGen2: Towards Instruction-Aligned Multimodal Generation},
  journal      = {CoRR},
  volume       = {abs/2506.18871},
  year         = {2025},
  url          = {https://doi.org/10.48550/arXiv.2506.18871},
  doi          = {10.48550/ARXIV.2506.18871},
  eprinttype   = {arXiv},
  eprint       = {2506.18871},
  bibsource    = {dblp computer science bibliography, https://dblp.org}
}

@article{DBLP:journals/corr/abs-2504-17761,
  author       = {Shiyu Liu and
                  Yucheng Han and
                  Peng Xing and
                  Fukun Yin and
                  Rui Wang and
                  Wei Cheng and
                  Jiaqi Liao and
                  Yingming Wang and
                  Honghao Fu and
                  Chunrui Han and
                  Guopeng Li and
                  Yuang Peng and
                  Quan Sun and
                  Jingwei Wu and
                  Yan Cai and
                  Zheng Ge and
                  Ranchen Ming and
                  Lei Xia and
                  Xianfang Zeng and
                  Yibo Zhu and
                  Binxing Jiao and
                  Xiangyu Zhang and
                  Gang Yu and
                  Daxin Jiang},
  title        = {Step1X-Edit: {A} Practical Framework for General Image Editing},
  journal      = {CoRR},
  volume       = {abs/2504.17761},
  year         = {2025},
  url          = {https://doi.org/10.48550/arXiv.2504.17761},
  doi          = {10.48550/ARXIV.2504.17761},
  eprinttype   = {arXiv},
  eprint       = {2504.17761},
  bibsource    = {dblp computer science bibliography, https://dblp.org}
}

@article{DBLP:journals/corr/abs-2505-14683,
  author       = {Chaorui Deng and
                  Deyao Zhu and
                  Kunchang Li and
                  Chenhui Gou and
                  Feng Li and
                  Zeyu Wang and
                  Shu Zhong and
                  Weihao Yu and
                  Xiaonan Nie and
                  Ziang Song and
                  Shi Guang and
                  Haoqi Fan},
  title        = {Emerging Properties in Unified Multimodal Pretraining},
  journal      = {CoRR},
  volume       = {abs/2505.14683},
  year         = {2025},
  url          = {https://doi.org/10.48550/arXiv.2505.14683},
  doi          = {10.48550/ARXIV.2505.14683},
  eprinttype   = {arXiv},
  eprint       = {2505.14683},
  bibsource    = {dblp computer science bibliography, https://dblp.org}
}

@article{DBLP:journals/corr/abs-2508-02324,
  author       = {Chenfei Wu and
                  Jiahao Li and
                  Jingren Zhou and
                  Junyang Lin and
                  Kaiyuan Gao and
                  Kun Yan and
                  Shengming Yin and
                  Shuai Bai and
                  Xiao Xu and
                  Yilei Chen and
                  Yuxiang Chen and
                  Zecheng Tang and
                  Zekai Zhang and
                  Zhengyi Wang and
                  An Yang and
                  Bowen Yu and
                  Chen Cheng and
                  Dayiheng Liu and
                  Deqing Li and
                  Hang Zhang and
                  Hao Meng and
                  Hu Wei and
                  Jingyuan Ni and
                  Kai Chen and
                  Kuan Cao and
                  Liang Peng and
                  Lin Qu and
                  Minggang Wu and
                  Peng Wang and
                  Shuting Yu and
                  Tingkun Wen and
                  Wensen Feng and
                  Xiaoxiao Xu and
                  Yi Wang and
                  Yichang Zhang and
                  Yongqiang Zhu and
                  Yujia Wu and
                  Yuxuan Cai and
                  Zenan Liu},
  title        = {Qwen-Image Technical Report},
  journal      = {CoRR},
  volume       = {abs/2508.02324},
  year         = {2025},
  url          = {https://doi.org/10.48550/arXiv.2508.02324},
  doi          = {10.48550/ARXIV.2508.02324},
  eprinttype   = {arXiv},
  eprint       = {2508.02324},
  bibsource    = {dblp computer science bibliography, https://dblp.org}
}

@misc{blackforestlabs2025flux2,
  author       = {{Black Forest Labs}},
  title        = {FLUX.2: Frontier Visual Intelligence},
  year         = {2025},
  howpublished = {\url{https://bfl.ai/blog/flux-2}}
}

@article{DBLP:journals/corr/abs-2506-15742,
  author       = {{Black Forest Labs} and
                  Stephen Batifol and
                  Andreas Blattmann and
                  Frederic Boesel and
                  Saksham Consul and
                  Cyril Diagne and
                  Tim Dockhorn and
                  Jack English and
                  Zion English and
                  Patrick Esser and
                  Sumith Kulal and
                  Kyle Lacey and
                  Yam Levi and
                  Cheng Li and
                  Dominik Lorenz and
                  Jonas M{\"{u}}ller and
                  Dustin Podell and
                  Robin Rombach and
                  Harry Saini and
                  Axel Sauer and
                  Luke Smith},
  title        = {{FLUX.1} Kontext: Flow Matching for In-Context Image Generation and
                  Editing in Latent Space},
  journal      = {CoRR},
  volume       = {abs/2506.15742},
  year         = {2025},
  url          = {https://doi.org/10.48550/arXiv.2506.15742},
  doi          = {10.48550/ARXIV.2506.15742},
  eprinttype   = {arXiv},
  eprint       = {2506.15742},
  bibsource    = {dblp computer science bibliography, https://dblp.org}
}

@article{DBLP:journals/corr/abs-2602-13344,
  author       = {Changhao Qiao and
                  Chao Hui and
                  Chen Li and
                  Cunzheng Wang and
                  Dejia Song and
                  Jiale Zhang and
                  Jing Li and
                  Qiang Xiang and
                  Runqi Wang and
                  Shuang Sun and
                  Wei Zhu and
                  Xu Tang and
                  Yao Hu and
                  Yibo Chen and
                  Yuhao Huang and
                  Yuxuan Duan and
                  Zhiyi Chen and
                  Ziyuan Guo},
  title        = {FireRed-Image-Edit-1.0 Technical Report},
  journal      = {CoRR},
  volume       = {abs/2602.13344},
  year         = {2026},
  url          = {https://doi.org/10.48550/arXiv.2602.13344},
  doi          = {10.48550/ARXIV.2602.13344},
  eprinttype   = {arXiv},
  eprint       = {2602.13344},
  bibsource    = {dblp computer science bibliography, https://dblp.org}
}

@inproceedings{DBLP:conf/nips/LuYRHXWSZZL25,
  author       = {Zimu Lu and
                  Yunqiao Yang and
                  Houxing Ren and
                  Haotian Hou and
                  Han Xiao and
                  Ke Wang and
                  Weikang Shi and
                  Aojun Zhou and
                  Mingjie Zhan and
                  Hongsheng Li},
  editor       = {Danielle Belgrave and
                  Cheng Zhang and
                  Laura N. Montoya and
                  Hsuan{-}Tien Lin and
                  Razvan Pascanu and
                  Piotr Koniusz and
                  Marzyeh Ghassemi and
                  Nancy Chen and
                  Iv{\'{a}}n Vladimir Meza Ru{\'{\i}}z and
                  Arturo Loaiza{-}Bonilla},
  title        = {WebGen-Bench: Evaluating LLMs on Generating Interactive and Functional
                  Websites from Scratch},
  booktitle    = {Advances in Neural Information Processing Systems 38: Annual Conference
                  on Neural Information Processing Systems 2025, NeurIPS 2025, San Diego,
                  CA, USA, December 2-7, 2025 / Mexico City, Mexico, November 30 - December
                  5, 2025},
  year         = {2025},
  url          = {http://papers.nips.cc/paper\_files/paper/2025/hash/6841eed8bb6a2ec49e49235c8115efee-Abstract-Datasets\_and\_Benchmarks\_Track.html},
  bibsource    = {dblp computer science bibliography, https://dblp.org}
}

@inproceedings{DBLP:conf/iclr/KahouMAKTB18,
  author       = {Samira Ebrahimi Kahou and
                  Vincent Michalski and
                  Adam Atkinson and
                  {\'{A}}kos K{\'{a}}d{\'{a}}r and
                  Adam Trischler and
                  Yoshua Bengio},
  title        = {FigureQA: An Annotated Figure Dataset for Visual Reasoning},
  booktitle    = {6th International Conference on Learning Representations, {ICLR} 2018,
                  Vancouver, BC, Canada, April 30 - May 3, 2018, Workshop Track Proceedings},
  publisher    = {OpenReview.net},
  year         = {2018},
  url          = {https://openreview.net/forum?id=H1mz0OyDz},
  bibsource    = {dblp computer science bibliography, https://dblp.org}
}

@article{DBLP:journals/corr/abs-2603-09312,
  author       = {Feiyu Wang and
                  Jiayuan Yang and
                  Zhiyuan Zhao and
                  Da Zhang and
                  Bingyu Li and
                  Peng Liu and
                  Junyu Gao},
  title        = {IntroSVG: Learning from Rendering Feedback for Text-to-SVG Generation
                  via an Introspective Generator-Critic Framework},
  journal      = {CoRR},
  volume       = {abs/2603.09312},
  year         = {2026},
  url          = {https://doi.org/10.48550/arXiv.2603.09312},
  doi          = {10.48550/ARXIV.2603.09312},
  eprinttype   = {arXiv},
  eprint       = {2603.09312},
  bibsource    = {dblp computer science bibliography, https://dblp.org}
}

@article{DBLP:journals/corr/abs-2310-02426,
  author       = {Samyadeep Basu and
                  Mehrdad Saberi and
                  Shweta Bhardwaj and
                  Atoosa Malemir Chegini and
                  Daniela Massiceti and
                  Maziar Sanjabi and
                  Shell Xu Hu and
                  Soheil Feizi},
  title        = {EditVal: Benchmarking Diffusion Based Text-Guided Image Editing Methods},
  journal      = {CoRR},
  volume       = {abs/2310.02426},
  year         = {2023},
  url          = {https://doi.org/10.48550/arXiv.2310.02426},
  doi          = {10.48550/ARXIV.2310.02426},
  eprinttype   = {arXiv},
  eprint       = {2310.02426},
  bibsource    = {dblp computer science bibliography, https://dblp.org}
}

@inproceedings{DBLP:conf/nips/ZhangMCSS23,
  author       = {Kai Zhang and
                  Lingbo Mo and
                  Wenhu Chen and
                  Huan Sun and
                  Yu Su},
  editor       = {Alice Oh and
                  Tristan Naumann and
                  Amir Globerson and
                  Kate Saenko and
                  Moritz Hardt and
                  Sergey Levine},
  title        = {MagicBrush: {A} Manually Annotated Dataset for Instruction-Guided
                  Image Editing},
  booktitle    = {Advances in Neural Information Processing Systems 36: Annual Conference
                  on Neural Information Processing Systems 2023, NeurIPS 2023, New Orleans,
                  LA, USA, December 10 - 16, 2023},
  year         = {2023},
  url          = {http://papers.nips.cc/paper\_files/paper/2023/hash/64008fa30cba9b4d1ab1bd3bd3d57d61-Abstract-Datasets\_and\_Benchmarks.html},
  bibsource    = {dblp computer science bibliography, https://dblp.org}
}

@article{DBLP:journals/corr/abs-2504-13143,
  author       = {Siwei Yang and
                  Mude Hui and
                  Bingchen Zhao and
                  Yuyin Zhou and
                  Nataniel Ruiz and
                  Cihang Xie},
  title        = {Complex-Edit: CoT-Like Instruction Generation for Complexity-Controllable
                  Image Editing Benchmark},
  journal      = {CoRR},
  volume       = {abs/2504.13143},
  year         = {2025},
  url          = {https://doi.org/10.48550/arXiv.2504.13143},
  doi          = {10.48550/ARXIV.2504.13143},
  eprinttype   = {arXiv},
  eprint       = {2504.13143},
  bibsource    = {dblp computer science bibliography, https://dblp.org}
}

@article{DBLP:journals/corr/abs-2602-13028,
  author       = {Runzhou Liu and
                  Hailey Weingord and
                  Sejal Mittal and
                  Prakhar Dungarwal and
                  Anusha Nandula and
                  Bo Ni and
                  Samyadeep Basu and
                  Hongjie Chen and
                  Nesreen K. Ahmed and
                  Li Li and
                  Jiayi Zhang and
                  Koustava Goswami and
                  Subhojyoti Mukherjee and
                  Branislav Kveton and
                  Puneet Mathur and
                  Franck Dernoncourt and
                  Yue Zhao and
                  Yu Wang and
                  Ryan A. Rossi and
                  Zhengzhong Tu and
                  Hongru Du},
  title        = {Human-Aligned {MLLM} Judges for Fine-Grained Image Editing Evaluation:
                  {A} Benchmark, Framework, and Analysis},
  journal      = {CoRR},
  volume       = {abs/2602.13028},
  year         = {2026},
  url          = {https://doi.org/10.48550/arXiv.2602.13028},
  doi          = {10.48550/ARXIV.2602.13028},
  eprinttype   = {arXiv},
  eprint       = {2602.13028},
  bibsource    = {dblp computer science bibliography, https://dblp.org}
}

@article{DBLP:journals/corr/abs-2510-05091,
  author       = {Le Zhuo and
                  Songhao Han and
                  Yuandong Pu and
                  Boxiang Qiu and
                  Sayak Paul and
                  Yue Liao and
                  Yihao Liu and
                  Jie Shao and
                  Xi Chen and
                  Si Liu and
                  Hongsheng Li},
  title        = {Factuality Matters: When Image Generation and Editing Meet Structured
                  Visuals},
  journal      = {CoRR},
  volume       = {abs/2510.05091},
  year         = {2025},
  url          = {https://doi.org/10.48550/arXiv.2510.05091},
  doi          = {10.48550/ARXIV.2510.05091},
  eprinttype   = {arXiv},
  eprint       = {2510.05091},
  bibsource    = {dblp computer science bibliography, https://dblp.org}
}

@article{DBLP:journals/corr/abs-2602-00122,
  author       = {Hongzhu Yi and
                  Yujia Yang and
                  Yuanxiang Wang and
                  Zhenyu Guan and
                  Jiahuan Chen and
                  Chenxi Bao and
                  Tiankun Yang and
                  Yixuan Yuan and
                  Tianyu Zong and
                  Xinming Wang and
                  Tao Yu and
                  Ruiwen Tao and
                  Haijin Liang and
                  Jin Ma and
                  Jinwen Luo and
                  Xinyu Zuo and
                  Jungang Xu},
  title        = {{VDE} Bench: Evaluating The Capability of Image Editing Models to
                  Modify Visual Documents},
  journal      = {CoRR},
  volume       = {abs/2602.00122},
  year         = {2026},
  url          = {https://doi.org/10.48550/arXiv.2602.00122},
  doi          = {10.48550/ARXIV.2602.00122},
  eprinttype   = {arXiv},
  eprint       = {2602.00122},
  bibsource    = {dblp computer science bibliography, https://dblp.org}
}

@inproceedings{DBLP:conf/nips/MaJYLWZZSJ24,
  author       = {Yiwei Ma and
                  Jiayi Ji and
                  Ke Ye and
                  Weihuang Lin and
                  Zhibin Wang and
                  Yonghan Zheng and
                  Qiang Zhou and
                  Xiaoshuai Sun and
                  Rongrong Ji},
  editor       = {Amir Globersons and
                  Lester Mackey and
                  Danielle Belgrave and
                  Angela Fan and
                  Ulrich Paquet and
                  Jakub M. Tomczak and
                  Cheng Zhang},
  title        = {I2EBench: {A} Comprehensive Benchmark for Instruction-based Image
                  Editing},
  booktitle    = {Advances in Neural Information Processing Systems 37: Annual Conference
                  on Neural Information Processing Systems 2024, NeurIPS 2024, Vancouver,
                  BC, Canada, December 10 - 15, 2024},
  year         = {2024},
  url          = {http://papers.nips.cc/paper\_files/paper/2024/hash/48fecef47b19fe501d27d338b6d52582-Abstract-Conference.html},
  bibsource    = {dblp computer science bibliography, https://dblp.org}
}

@inproceedings{DBLP:conf/nips/YeHLLYYHY25,
  author       = {Yang Ye and
                  Xianyi He and
                  Zongjian Li and
                  Bin Lin and
                  Shenghai Yuan and
                  Zhiyuan Yan and
                  Bohan Hou and
                  Li Yuan},
  editor       = {Danielle Belgrave and
                  Cheng Zhang and
                  Laura N. Montoya and
                  Hsuan{-}Tien Lin and
                  Razvan Pascanu and
                  Piotr Koniusz and
                  Marzyeh Ghassemi and
                  Nancy Chen and
                  Iv{\'{a}}n Vladimir Meza Ru{\'{\i}}z and
                  Arturo Loaiza{-}Bonilla},
  title        = {ImgEdit: {A} Unified Image Editing Dataset and Benchmark},
  booktitle    = {Advances in Neural Information Processing Systems 38: Annual Conference
                  on Neural Information Processing Systems 2025, NeurIPS 2025, San Diego,
                  CA, USA, December 2-7, 2025 / Mexico City, Mexico, November 30 - December
                  5, 2025},
  year         = {2025},
  url          = {http://papers.nips.cc/paper\_files/paper/2025/hash/bec783f119b969772c35243b1c801090-Abstract-Datasets\_and\_Benchmarks\_Track.html},
  bibsource    = {dblp computer science bibliography, https://dblp.org}
}

@article{DBLP:journals/corr/abs-2505-11493,
  author       = {Yusu Qian and
                  Jiasen Lu and
                  Tsu{-}Jui Fu and
                  Xinze Wang and
                  Chen Chen and
                  Yinfei Yang and
                  Wenze Hu and
                  Zhe Gan},
  title        = {GIE-Bench: Towards Grounded Evaluation for Text-Guided Image Editing},
  journal      = {CoRR},
  volume       = {abs/2505.11493},
  year         = {2025},
  url          = {https://doi.org/10.48550/arXiv.2505.11493},
  doi          = {10.48550/ARXIV.2505.11493},
  eprinttype   = {arXiv},
  eprint       = {2505.11493},
  bibsource    = {dblp computer science bibliography, https://dblp.org}
}

@inproceedings{DBLP:conf/mm/XuDLMWYGW0MZL25,
  author       = {Zitong Xu and
                  Huiyu Duan and
                  Bingnan Liu and
                  Guangji Ma and
                  Jiarui Wang and
                  Liu Yang and
                  Shiqi Gao and
                  Xiaoyu Wang and
                  Jia Wang and
                  Xiongkuo Min and
                  Guangtao Zhai and
                  Weisi Lin},
  editor       = {Cathal Gurrin and
                  Klaus Schoeffmann and
                  Min Zhang and
                  Luca Rossetto and
                  Stevan Rudinac and
                  Duc{-}Tien Dang{-}Nguyen and
                  Wen{-}Huang Cheng and
                  Phoebe Chen and
                  Jenny Benois{-}Pineau},
  title        = {LMM4Edit: Benchmarking and Evaluating Multimodal Image Editing with
                  LMMs},
  booktitle    = {Proceedings of the 33rd {ACM} International Conference on Multimedia,
                  {MM} 2025, Dublin, Ireland, October 27-31, 2025},
  pages        = {6908--6917},
  publisher    = {{ACM}},
  year         = {2025},
  url          = {https://doi.org/10.1145/3746027.3755294},
  doi          = {10.1145/3746027.3755294},
  bibsource    = {dblp computer science bibliography, https://dblp.org}
}

@article{DBLP:journals/corr/abs-2606-00188,
  author       = {Kai Xu and
                  Ellis Brown and
                  Shrikar Madhu and
                  Rob Fergus and
                  He He and
                  Saining Xie},
  title        = {PaintBench: Deterministic Evaluation of Precise Visual Editing},
  journal      = {CoRR},
  volume       = {abs/2606.00188},
  year         = {2026},
  url          = {https://doi.org/10.48550/arXiv.2606.00188},
  doi          = {10.48550/ARXIV.2606.00188},
  eprinttype   = {arXiv},
  eprint       = {2606.00188},
  bibsource    = {dblp computer science bibliography, https://dblp.org}
}

@article{DBLP:journals/corr/abs-2607-02290,
  author       = {Zhaokai Wang and
                  Mingxin Liu and
                  Zirun Zhu and
                  Ziqian Fan and
                  Yiguo He and
                  Mohan Zhang and
                  Leyao Gu and
                  Xiangyu Zhao and
                  Ning Liao and
                  Shaofeng Zhang and
                  Xuanhe Zhou and
                  Zhihang Zhong and
                  Junchi Yan and
                  Xue Yang},
  title        = {DisciplineGen-1M: {A} Large-Scale Dataset for Multidisciplinary Visual
                  Generation and Editing},
  journal      = {CoRR},
  volume       = {abs/2607.02290},
  year         = {2026},
  url          = {https://doi.org/10.48550/arXiv.2607.02290},
  doi          = {10.48550/ARXIV.2607.02290},
  eprinttype   = {arXiv},
  eprint       = {2607.02290},
  bibsource    = {dblp computer science bibliography, https://dblp.org}
}

@article{DBLP:journals/corr/abs-2602-06663,
  author       = {Junxian Li and
                  Kai Liu and
                  Leyang Chen and
                  Weida Wang and
                  Zhixin Wang and
                  Jiaqi Xu and
                  Fan Li and
                  Renjing Pei and
                  Linghe Kong and
                  Yulun Zhang},
  title        = {PlanViz: Evaluating Planning-Oriented Image Generation and Editing
                  for Computer-Use Tasks},
  journal      = {CoRR},
  volume       = {abs/2602.06663},
  year         = {2026},
  url          = {https://doi.org/10.48550/arXiv.2602.06663},
  doi          = {10.48550/ARXIV.2602.06663},
  eprinttype   = {arXiv},
  eprint       = {2602.06663},
  bibsource    = {dblp computer science bibliography, https://dblp.org}
}

@article{DBLP:journals/corr/abs-2607-10301,
  author       = {Jiakang Yu and
                  Yixuan Chai and
                  Tianci Wang and
                  Rihui Jin and
                  Guangkai Xu and
                  Hongtao Deng and
                  Xun Zhu and
                  Wang Gao and
                  Xinrun Guo and
                  Haipang Wu},
  title        = {ChartSync: {A} Benchmark for Visuo-Logical Cascading Chart Editing},
  journal      = {CoRR},
  volume       = {abs/2607.10301},
  year         = {2026},
  url          = {https://doi.org/10.48550/arXiv.2607.10301},
  doi          = {10.48550/ARXIV.2607.10301},
  eprinttype   = {arXiv},
  eprint       = {2607.10301},
  bibsource    = {dblp computer science bibliography, https://dblp.org}
}

@article{DBLP:journals/corr/abs-2502-19453,
  author       = {Kunato Nishina and
                  Yusuke Matsui},
  title        = {SVGEditBench {V2:} {A} Benchmark for Instruction-based {SVG} Editing},
  journal      = {CoRR},
  volume       = {abs/2502.19453},
  year         = {2025},
  url          = {https://doi.org/10.48550/arXiv.2502.19453},
  doi          = {10.48550/ARXIV.2502.19453},
  eprinttype   = {arXiv},
  eprint       = {2502.19453},
  bibsource    = {dblp computer science bibliography, https://dblp.org}
}

@inproceedings{DBLP:conf/lrec/KucharKSS26,
  author       = {Josef Kuchar and
                  Marek Kadlc{\'{\i}}k and
                  Michal Spiegel and
                  Michal Stef{\'{a}}nik},
  editor       = {Stelios Piperidis and
                  N{\'{u}}ria Bel and
                  Henk van den Heuvel and
                  Nancy Ide and
                  Simon Krek and
                  Antonio Toral},
  title        = {VectorEdits: {A} Dataset and Benchmark for Instruction-Based Editing
                  of Vector Graphics},
  booktitle    = {Proceedings of the Fifteenth Language Resources and Evaluation Conference,
                  {LREC} 2026, Palma de Mallorca, Spain, May 13-15, 2026},
  pages        = {11119--11124},
  publisher    = {{ELRA} Language Resource Association / {ACL}},
  year         = {2026},
  url          = {https://doi.org/10.63317/5gc5ibtb5k8i},
  doi          = {10.63317/5GC5IBTB5K8I},
  bibsource    = {dblp computer science bibliography, https://dblp.org}
}

@article{DBLP:journals/corr/abs-2607-19056,
  author       = {Yug Aditi Gupta and
                  Prannay Hebbar},
  title        = {Vector-Bench: Can Models Surgically Edit {SVG} Code?},
  journal      = {CoRR},
  volume       = {abs/2607.19056},
  year         = {2026},
  url          = {https://doi.org/10.48550/arXiv.2607.19056},
  doi          = {10.48550/ARXIV.2607.19056},
  eprinttype   = {arXiv},
  eprint       = {2607.19056},
  bibsource    = {dblp computer science bibliography, https://dblp.org}
}

@article{DBLP:journals/corr/abs-2605-15677,
  author       = {Xiaoyan Su and
                  Peijie Dong and
                  Zhenheng Tang and
                  Song Tang and
                  Yuyao Zhai and
                  Kaitao Lin and
                  Liang Chen and
                  Yuhang Gai and
                  Yuyu Luo and
                  Qian Wang and
                  Xiaowen Chu},
  title        = {VCG-Bench: Towards {A} Unified Visual-Centric Benchmark for Structured
                  Generation and Editing},
  journal      = {CoRR},
  volume       = {abs/2605.15677},
  year         = {2026},
  url          = {https://doi.org/10.48550/arXiv.2605.15677},
  doi          = {10.48550/ARXIV.2605.15677},
  eprinttype   = {arXiv},
  eprint       = {2605.15677},
  bibsource    = {dblp computer science bibliography, https://dblp.org}
}

@inproceedings{DBLP:conf/acl/JungCYYJC26,
  author       = {Kyudan Jung and
                  Hojun Cho and
                  Jooyeol Yun and
                  Soyoung Yang and
                  Jaehyeok Jang and
                  Jaegul Choo},
  editor       = {Maria Liakata and
                  Viviane P. Moreira and
                  Jiajun Zhang and
                  David Jurgens},
  title        = {Talk to Your Slides: High-Efficiency Slide Editing via Language-Driven
                  Structured Data Manipulation},
  booktitle    = {Findings of the Association for Computational Linguistics, {ACL} 2026,
                  San Diego, California, United States, July 2-7, 2026},
  pages        = {3370--3399},
  publisher    = {Association for Computational Linguistics},
  year         = {2026},
  url          = {https://doi.org/10.18653/v1/2026.findings-acl.166},
  doi          = {10.18653/V1/2026.FINDINGS-ACL.166},
  bibsource    = {dblp computer science bibliography, https://dblp.org}
}

@inproceedings{DBLP:conf/cvpr/GeWZPSTSSFND25,
  author       = {Jiaxin Ge and
                  Zora Zhiruo Wang and
                  Xuhui Zhou and
                  Yi{-}Hao Peng and
                  Sanjay Subramanian and
                  Qinyue Tan and
                  Maarten Sap and
                  Alane Suhr and
                  Daniel Fried and
                  Graham Neubig and
                  Trevor Darrell},
  title        = {AutoPresent: Designing Structured Visuals from Scratch},
  booktitle    = {{IEEE/CVF} Conference on Computer Vision and Pattern Recognition,
                  {CVPR} 2025, Nashville, TN, USA, June 11-15, 2025},
  pages        = {2902--2911},
  publisher    = {Computer Vision Foundation / {IEEE}},
  year         = {2025},
  url          = {https://openaccess.thecvf.com/content/CVPR2025/html/Ge\_AutoPresent\_Designing\_Structured\_Visuals\_from\_Scratch\_CVPR\_2025\_paper.html},
  doi          = {10.1109/CVPR52734.2025.00276},
  bibsource    = {dblp computer science bibliography, https://dblp.org}
}

@inproceedings{DBLP:conf/cvpr/Wei0CW0GSYG25,
  author       = {Jingxuan Wei and
                  Cheng Tan and
                  Qi Chen and
                  Gaowei Wu and
                  Siyuan Li and
                  Zhangyang Gao and
                  Linzhuang Sun and
                  Bihui Yu and
                  Ruifeng Guo},
  title        = {From Words to Structured Visuals: {A} Benchmark and Framework for
                  Text-to-Diagram Generation and Editing},
  booktitle    = {{IEEE/CVF} Conference on Computer Vision and Pattern Recognition,
                  {CVPR} 2025, Nashville, TN, USA, June 11-15, 2025},
  pages        = {13315--13325},
  publisher    = {Computer Vision Foundation / {IEEE}},
  year         = {2025},
  url          = {https://openaccess.thecvf.com/content/CVPR2025/html/Wei\_From\_Words\_to\_Structured\_Visuals\_A\_Benchmark\_and\_Framework\_for\_CVPR\_2025\_paper.html},
  doi          = {10.1109/CVPR52734.2025.01243},
  bibsource    = {dblp computer science bibliography, https://dblp.org}
}
\bibliographystyle{iclr2027_conference}

\appendix

\clearpage
\raggedbottom
\let\appendixtable\table
\renewcommand{\table}[1][]{\appendixtable[H]}
\renewcommand{\topfraction}{0.95}
\renewcommand{\bottomfraction}{0.9}
\renewcommand{\textfraction}{0.05}
\renewcommand{\floatpagefraction}{0.75}
\setcounter{topnumber}{4}
\setcounter{bottomnumber}{3}
\setcounter{totalnumber}{6}
\section{Additional Experiments}
\label{sec:app-experiments}
This appendix provides detailed operation profiles, visual evidence, and
supplementary comparisons for Section~\ref{sec:experiments}.

\begin{figure}[H]
  \centering
  \includegraphics[width=\linewidth]{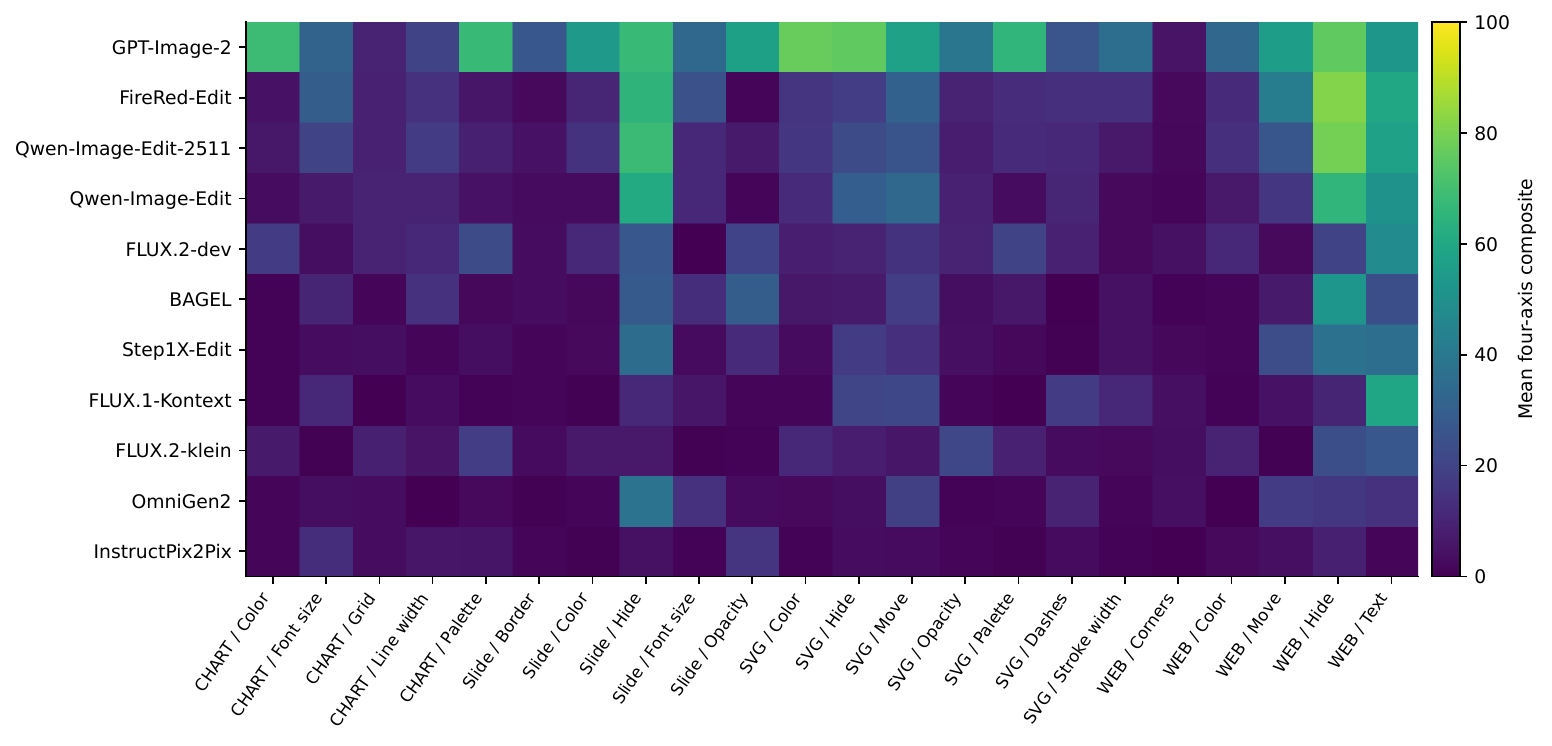}
  \caption{\datasetareview{Four-axis composite scores for all eleven models across the 22
  asset--operation combinations. Each cell averages the corresponding
  cases. }}
  \label{fig:exp-operations}
\end{figure}

\subsection{Effect of Case Weighting}
\label{sec:app-case-weighting}

\textbf{Setup.}
We change only the weights assigned to per-case $H$. With $n_c$ cases and
mean $\overline H_c$ in cell $c$, cell weighting gives
$\frac{1}{22}\sum_c\overline H_c$, whereas case weighting gives
$\frac{1}{1740}\sum_c n_c\overline H_c$.

\begin{table}[!htbp]
\centering
\caption{Effect of case weighting on the same 1,740 cases. Both rows average per-case harmonic composites. Cell-mean $H$ weights the 22 asset--operation cells equally; case-mean $H$ weights individual cases equally. Model order follows Table~\ref{tab:exp-main}; \textbf{bold} marks the best model and \underline{underlining} the best open-weight model.}
\label{tab:app-case-weighting}
\begingroup\scriptsize
\setlength{\tabcolsep}{2.0pt}\renewcommand{\arraystretch}{1.25}
\begin{tabular}{@{}l*{11}{r}@{}}
\noalign{\hrule height 0.9pt}
Metric & \shortstack{GPT-\\Image-2} & \shortstack{FireRed-\\Edit} & \draft{\shortstack{Qwen-Image-\\Edit-2511}} & \draft{\shortstack{Qwen-Image-\\Edit}} & \shortstack{FLUX.2-\\dev} & BAGEL & \draft{\shortstack{Step1X-\\Edit}} & \datasetareview{\shortstack{FLUX.1-\\Kontext}} & \datasetareview{\shortstack{FLUX.2-\\klein}} & OmniGen2 & \shortstack{Instruct\\Pix2Pix} \\
\noalign{\vskip 2pt\hrule height 0.45pt\vskip 2pt}
Cell-mean $H$ & \datasetareview{\textbf{47.13}} & \datasetareview{\underline{21.85}} & \datasetareview{20.47} & \datasetareview{16.26} & \datasetareview{13.11} & \datasetareview{10.81} & \datasetareview{9.78} & \datasetareview{8.78} & \datasetareview{8.42} & \datasetareview{7.32} & \datasetareview{3.83} \\
Case-mean $H$ & \datasetareview{\textbf{44.90}} & \datasetareview{\underline{17.76}} & \datasetareview{16.29} & \datasetareview{11.49} & \datasetareview{11.34} & \datasetareview{9.09} & \datasetareview{7.31} & \datasetareview{5.73} & \datasetareview{7.02} & \datasetareview{6.12} & 3.66 \\
\noalign{\vskip 2pt\hrule height 0.9pt}
\end{tabular}\endgroup
\end{table}

\textbf{Results and analysis.}
Table~\ref{tab:app-case-weighting} retains \draft{GPT-Image-2}, \draft{FireRed-Edit}, and Qwen-Image-Edit-2511
as the top three. \datasetareview{Case weighting places FLUX.2-klein and OmniGen2 above FLUX.1-Kontext;
the leading seven models retain their order.}

\datasetareview{The changes involve lower-ranked models whose adjacent default gaps are
statistically unresolved}
(Table~\ref{tab:app-paired-gaps}). Cell weighting prevents combinations with
more variants from dominating the comparison, whereas case weighting
reflects the benchmark's actual case distribution.

\subsection{Per-operation Axis Profiles}
\label{sec:app-axis}
\begin{figure}[H]
\centering
\includegraphics[width=\linewidth]{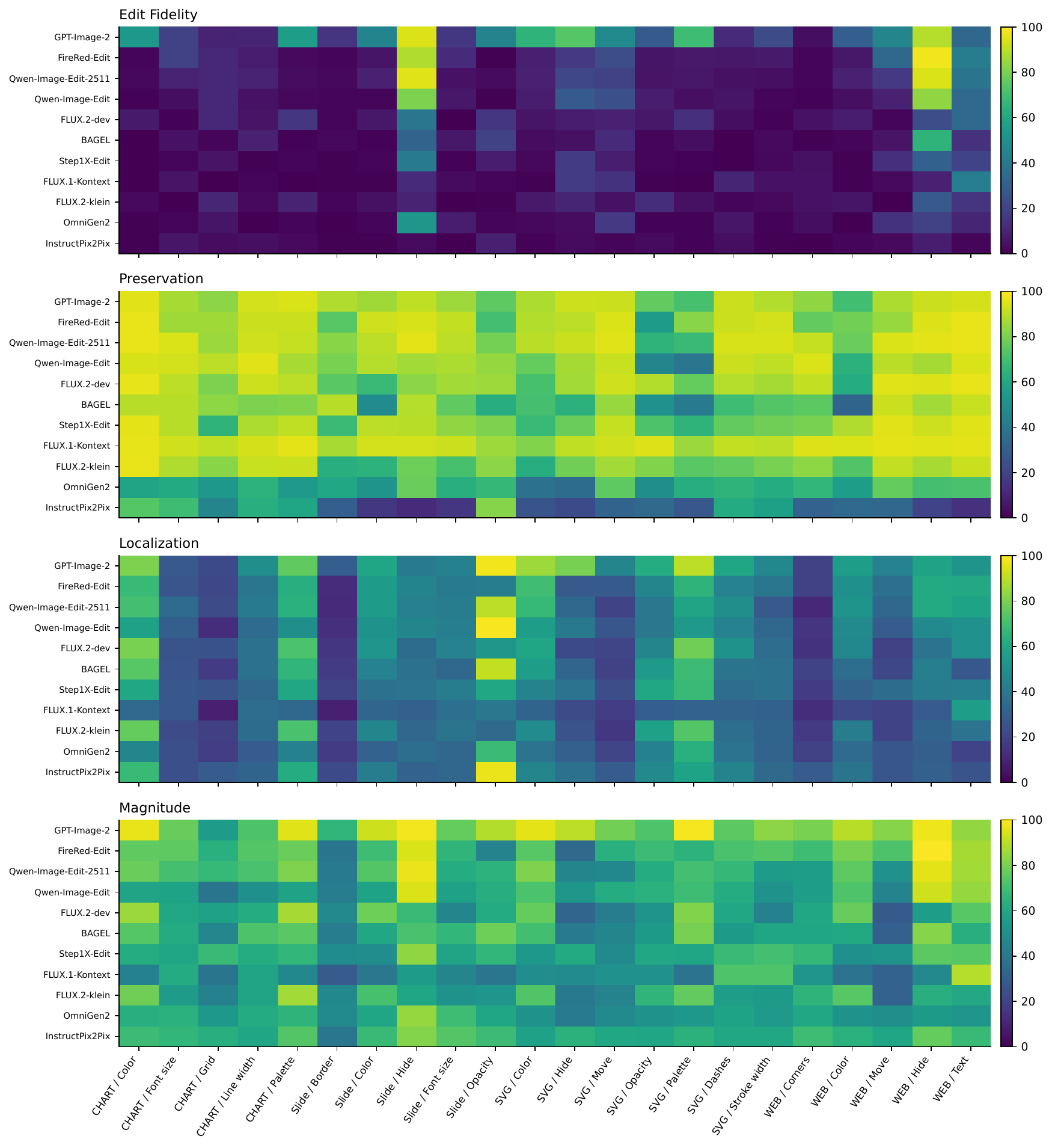}
\caption{\datasetareview{Full eleven-model axis profiles across all 22 asset--operation cells.
Each panel averages one axis within each cell, with a shared 0--100 scale.}}
\label{fig:app-axis}
\end{figure}

\textbf{Setup.}
We average each axis within every asset--operation cell for all eleven models
(Figure~\ref{fig:app-axis}).

\textbf{Results and analysis.}
Separating the four axes identifies whether a low composite reflects weak
target execution, preservation, localization, or change magnitude.

\subsection{Complete Images for the Qualitative Analysis}
\label{sec:app-visuals}
\begin{figure}[H]
\centering
\includegraphics[width=\linewidth]{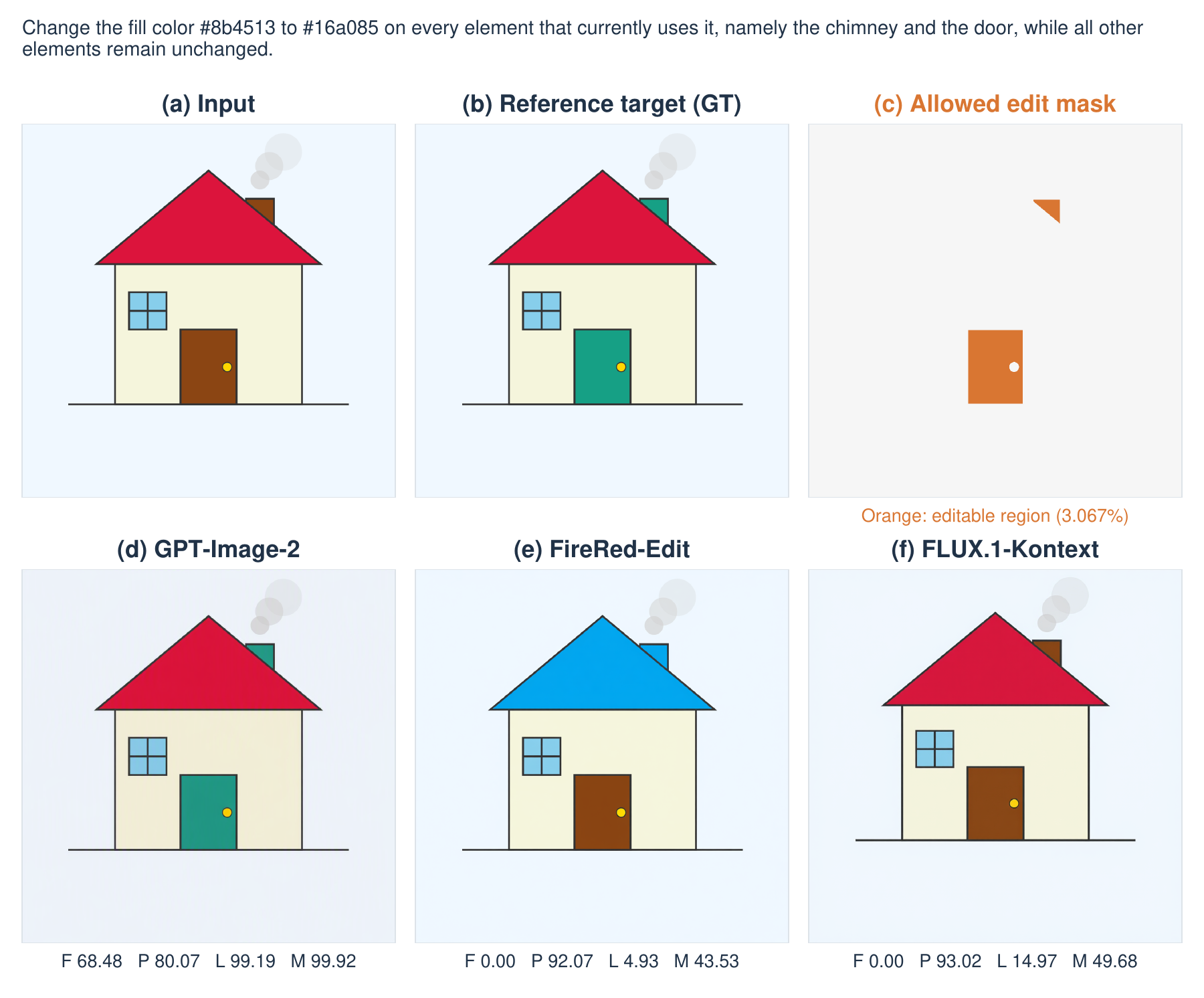}
\caption{SVG recoloring. Only the chimney and door should change color; the roof and surrounding content remain protected.}
\label{fig:app-example-1}
\end{figure}

\begin{figure}[H]
\centering
\includegraphics[width=\linewidth]{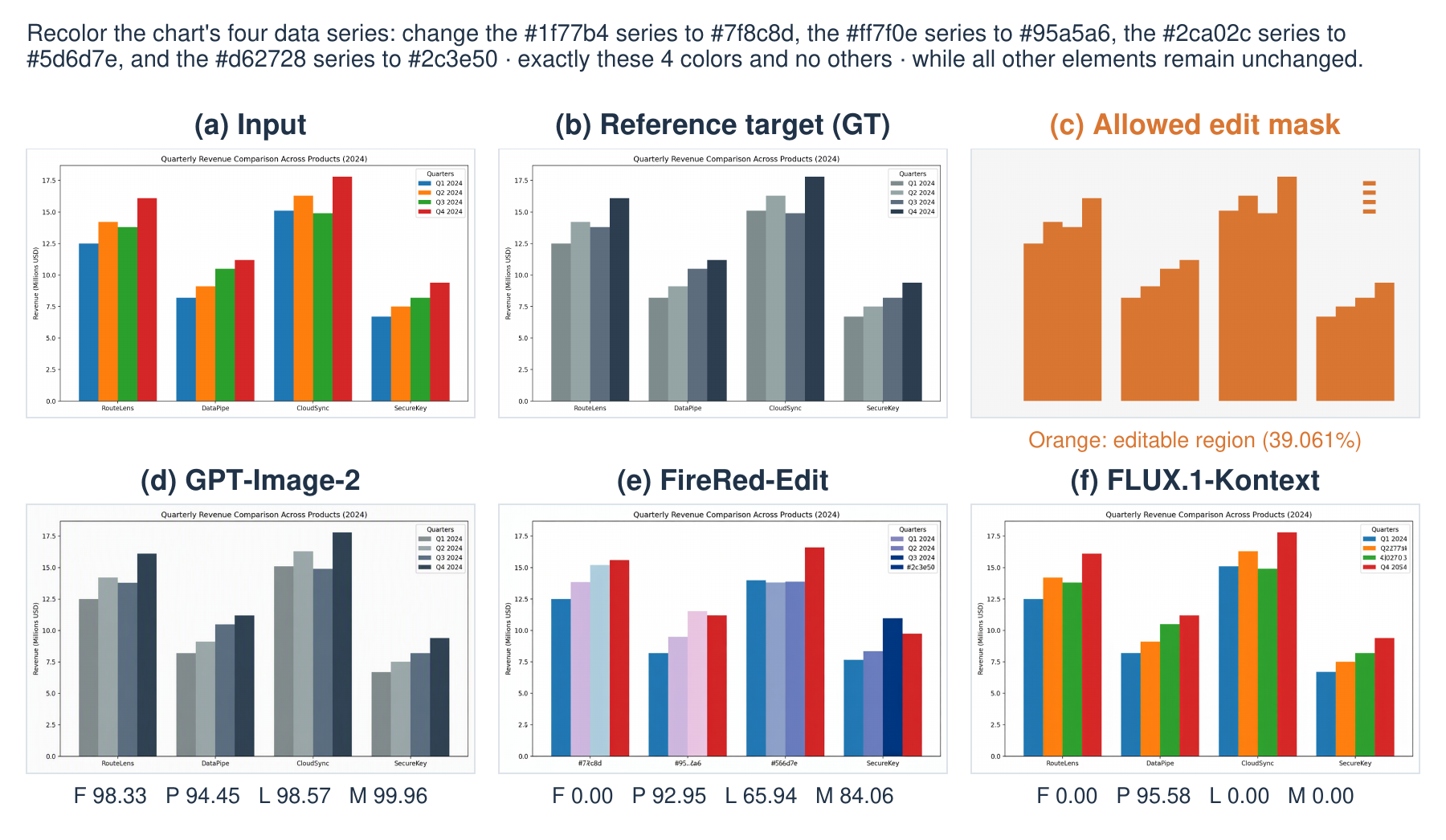}
\caption{Chart palette change. The four bar series should take the new muted palette while axes, labels, and surrounding content remain protected. GPT-Image-2 applies the palette; FireRed-Edit produces a partial recolor with hex-code text artifacts; FLUX.1-Kontext leaves the chart unchanged.}
\label{fig:app-example-2}
\end{figure}

\begin{figure}[H]
\centering
\includegraphics[width=0.96\linewidth]{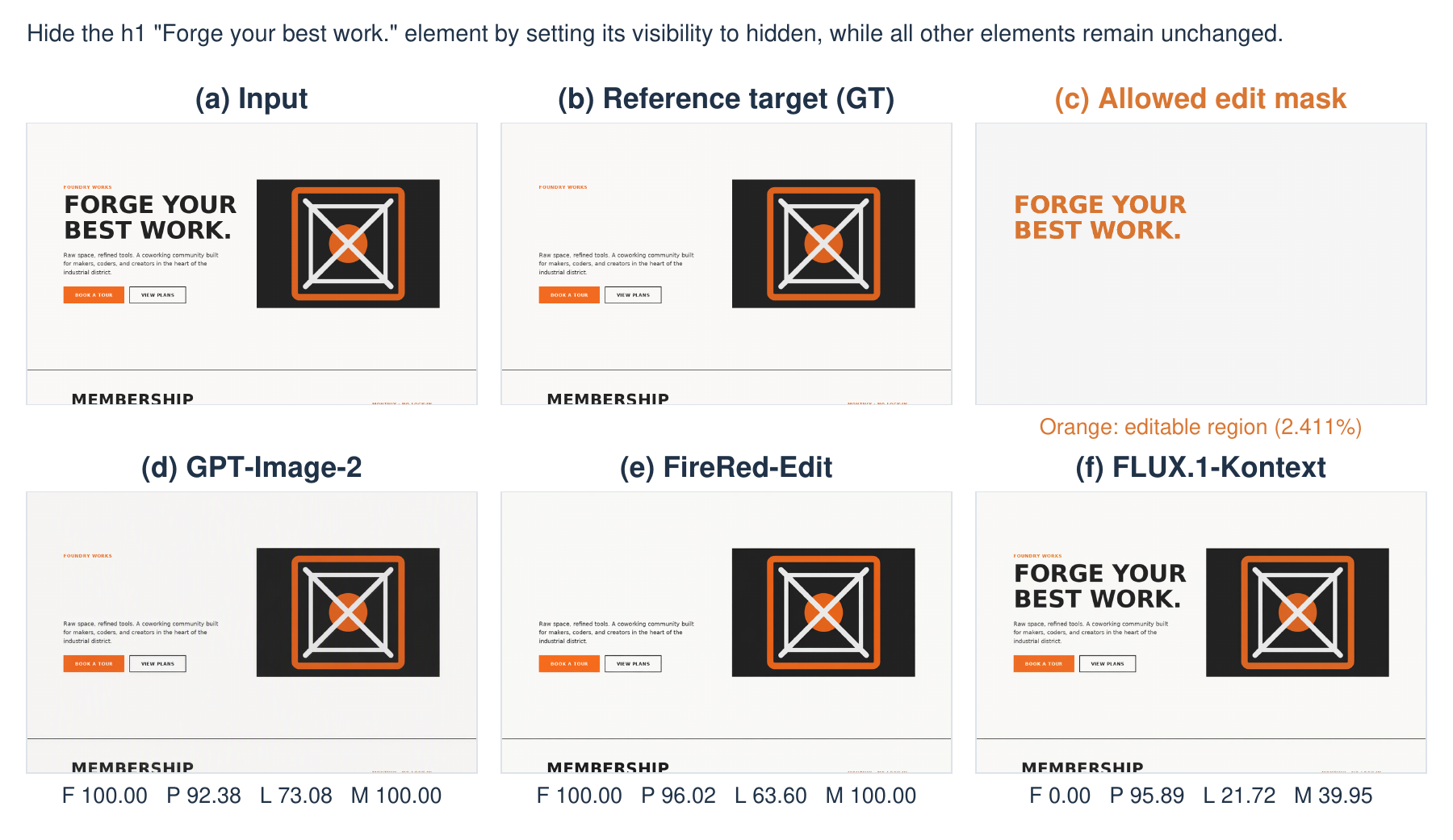}
\caption{Web heading hiding. The specified heading should disappear while its space and all surrounding content remain unchanged. GPT-Image-2 and FireRed-Edit remove the heading; FLUX.1-Kontext retains it.}
\label{fig:app-example-3}
\end{figure}

\begin{figure}[H]
\centering
\includegraphics[width=0.96\linewidth]{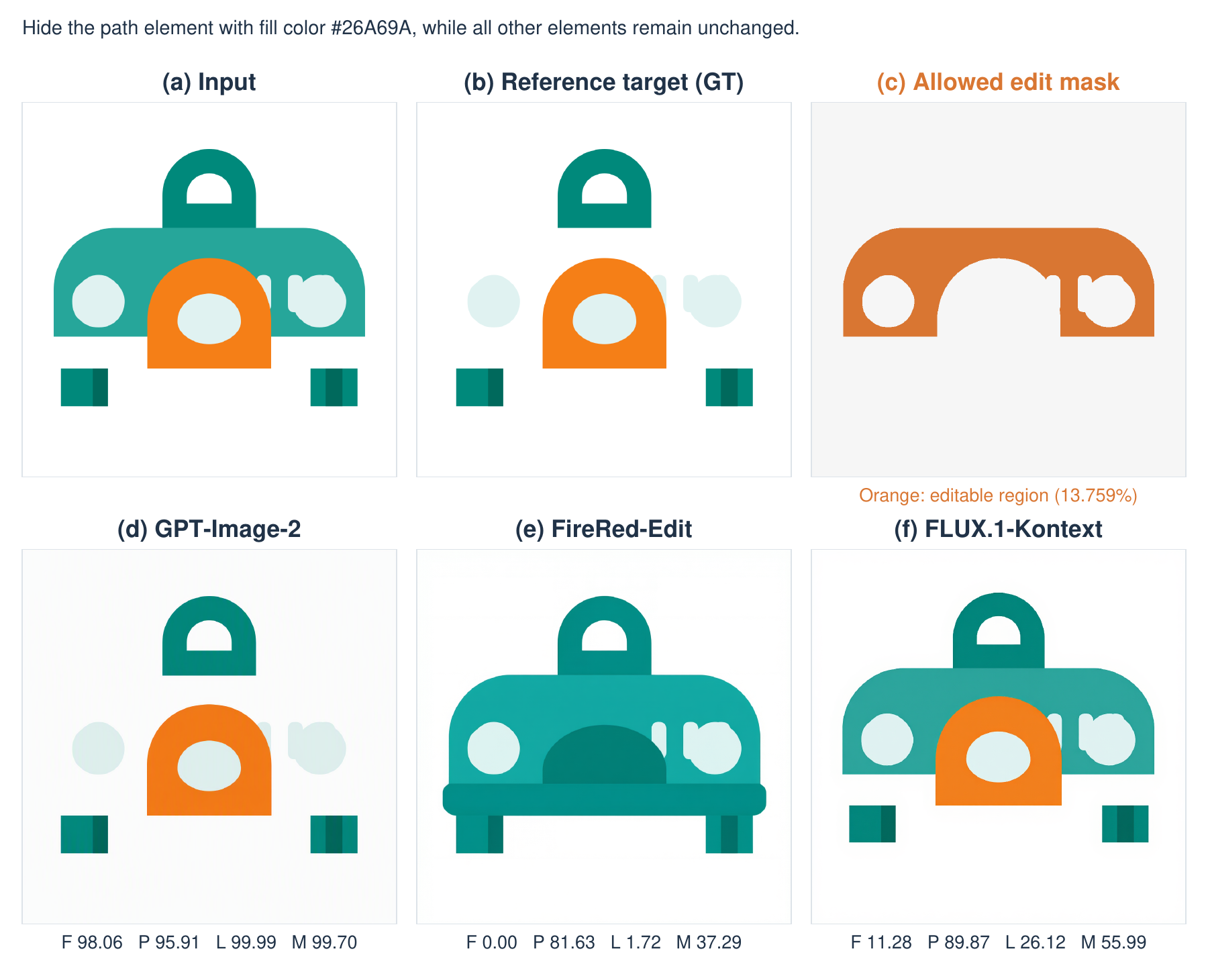}
\caption{SVG element hiding. The large teal body should disappear while every other element remains unchanged. GPT-Image-2 removes it; FireRed-Edit instead recolors it; FLUX.1-Kontext leaves it in place.}
\label{fig:app-example-4}
\end{figure}

\textbf{Setup.}
Figures~\ref{fig:app-example-1}--\ref{fig:app-example-4} share a common
layout: input, target, and edit mask above three model outputs. Orange shows
the evaluated mask without dilation. All scores use the complete images.

\textbf{Results and analysis.}
The complete views place target execution and changes to protected content
in the same visual context. For example, the web case distinguishes removal
of the requested heading from retaining it (Figure~\ref{fig:app-example-3}).

\subsection{Spatial Precision by Edit-region Size}
\label{sec:app-area}
\textbf{Setup.}
We divide detected changed-pixel area by allowed edit-mask area, retaining
zero-score outputs. Ratios above one imply changes outside the allowed region.

\begin{figure}[!htbp]
\centering
\includegraphics[width=\linewidth]{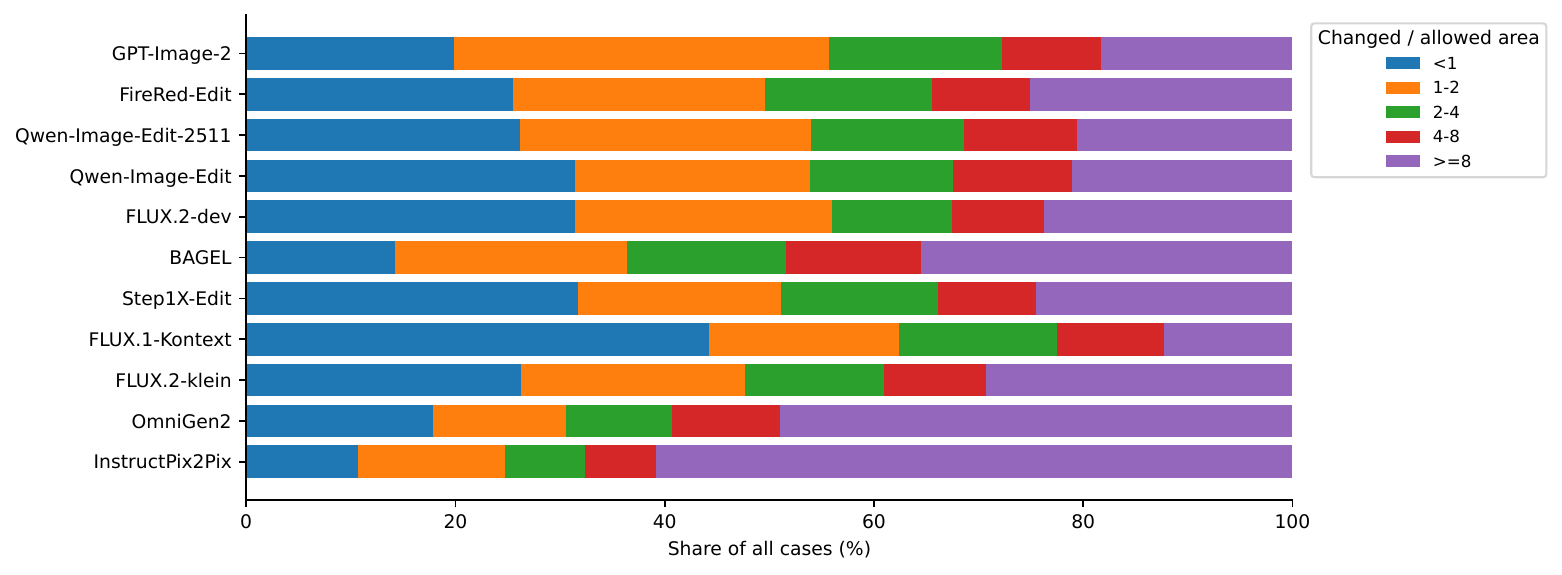}
\includegraphics[width=\linewidth]{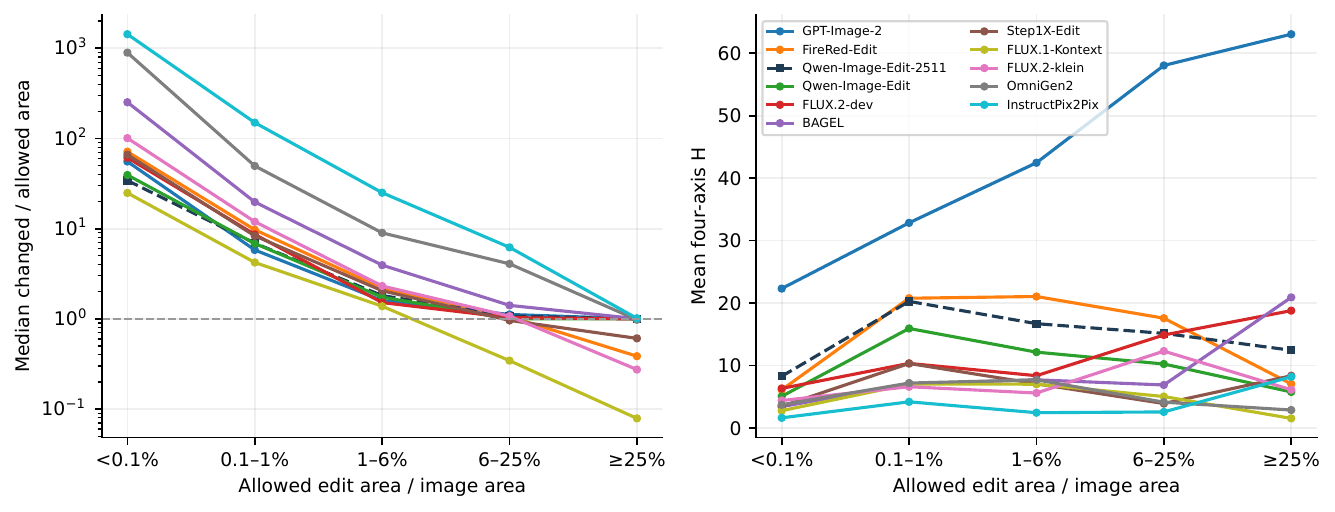}
\caption{\datasetareview{Detected-change area relative to the allowed edit region. Top: each model's distribution across ratio bins. Bottom: median area ratio (left) and mean composite $H$ (right) for each model, grouped by allowed-region size. All 19,140 outputs enter the analysis, including zero-score outputs.}}
\label{fig:app-area}
\end{figure}

\textbf{Results and analysis.}
The pooled medians are 84.28 for masks below 0.1\%, \datasetareview{10.57} for
0.1--1\%, and 1.00 for masks covering at least 25\% of the image.
Figure~\ref{fig:app-area} breaks down the distributions and region-size
effects by model, showing how much the pooled pattern varies across editors.
Operation and content also vary with region size, so this comparison
describes an association rather than an isolated effect of edit size.

\subsection{Complete Aggregation and Scoring Ablations}
\label{sec:app-ablation}
\textbf{Setup.}
We compare three within-case means with cell weighting fixed
(Table~\ref{tab:app-aggregation}) and vary color distance, scoring resolution,
and tolerances on identical outputs (Table~\ref{tab:app-scoring-variants}).
CIEDE2000 replaces the shared CIE76 distance function without retuning
thresholds; Lab-vector fits and magnitude calculations stay fixed.
Scorer-resolution controls downsample the already generated predictions
with Lanczos, then restore them with the indicated interpolation; model
inference settings remain fixed. Tighter and looser settings scale F/P
tolerance sets, the effective-change threshold (6), and the global-fit
reference-separation cap (8) by $2/3$ and $1.5$; magnitude's relative-error
band becomes $0.10$ and $0.20$ (default $0.15$).

The parameter sweep scales each of 11 parameters by 0.5, 0.75, 1.5, or 2
while fixing all others. Table~\ref{tab:app-parameter-settings} identifies
the defaults and settings. For localization's background cap, these
multipliers correspond to two, three, six, and eight times the allowed area,
around the default four.

\begin{figure}[!htbp]
\centering
\includegraphics[width=\linewidth]{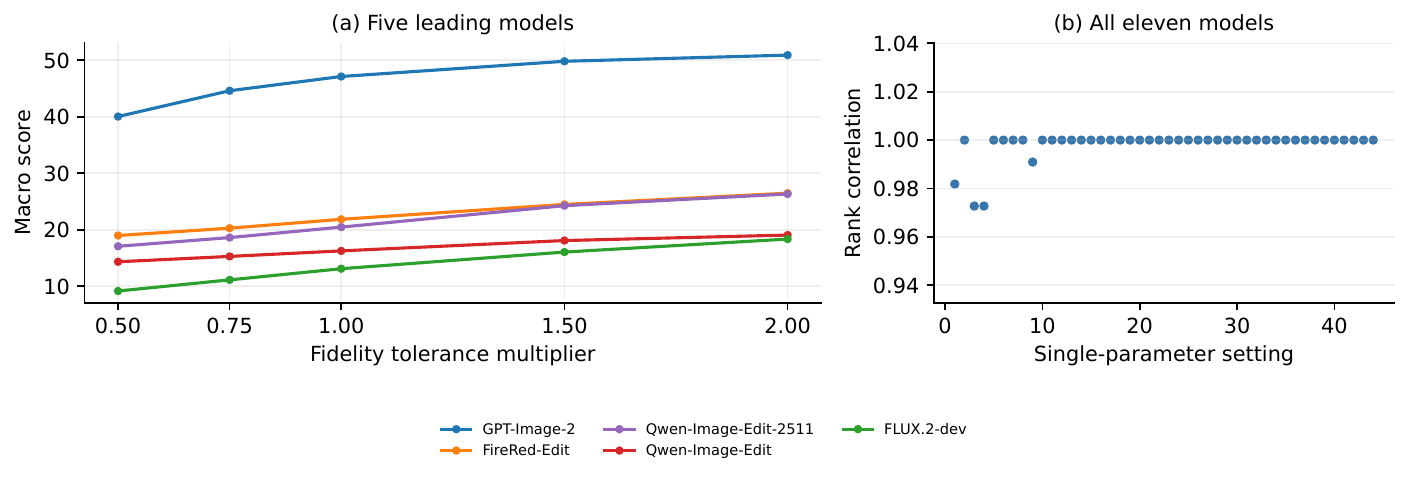}
\caption{\datasetareview{Scoring sensitivity on identical outputs. Left: composite scores of the five models leading under the default configuration as fidelity tolerances are scaled; a multiplier of one is the default. Right: eleven-model rank correlation for each of the 44 single-parameter settings. All other parameters and predictions stay fixed.}}
\label{fig:app-tolerance}
\end{figure}

\textbf{Results and analysis.}
Table~\ref{tab:app-aggregation} quantifies how much each model benefits when
strong axes can compensate for weak ones. Similar rankings can accompany
different absolute scores (Table~\ref{tab:app-scoring-variants}), so scores
from different configurations are not interchangeable.
Figure~\ref{fig:app-tolerance} shows the effects of fidelity tolerances on
the five leading models and of all parameters on the eleven-model ranking.
\datasetareview{The background-cap settings preserve the full model ranking.}
The cap limits how strongly widespread redraw dominates localization,
while preservation evaluates the entire protected region; the comparison
supports ranking stability over the tested range.

\begin{table}[t]
\centering
\caption{Aggregation of identical four-axis measurements on all 19,140 outputs. Each column applies the same 22-cell macro average; bold marks the best model within that column. The columns represent different scoring choices, not model improvements.}
\label{tab:app-aggregation}
\begingroup\small
\setlength{\tabcolsep}{4pt}\renewcommand{\arraystretch}{1.15}
\begin{tabular}{@{}lrrr@{}}
\noalign{\hrule height 0.8pt}
Model & Harmonic & Geometric & Arithmetic \\
\noalign{\vskip 2pt\hrule height 0.4pt\vskip 2pt}
GPT-Image-2 & \datasetareview{\textbf{47.13}} & \datasetareview{\textbf{52.26}} & \datasetareview{\textbf{66.55}} \\
FireRed-Edit & \datasetareview{21.85} & \datasetareview{26.65} & \datasetareview{54.60} \\
\draft{Qwen-Image-Edit-2511} & \datasetareview{20.47} & \datasetareview{25.85} & \datasetareview{54.71} \\
\draft{Qwen-Image-Edit} & \datasetareview{16.26} & \datasetareview{20.11} & \datasetareview{50.53} \\
FLUX.2-dev & \datasetareview{13.11} & \datasetareview{17.11} & \datasetareview{49.47} \\
BAGEL & \datasetareview{10.81} & \datasetareview{13.73} & \datasetareview{46.08} \\
Step1X-Edit & \datasetareview{9.78} & \datasetareview{13.10} & \datasetareview{48.01} \\
FLUX.2-klein & \datasetareview{8.42} & \datasetareview{11.60} & \datasetareview{46.07} \\
FLUX.1-Kontext & \datasetareview{8.78} & \datasetareview{11.52} & \datasetareview{44.07} \\
OmniGen2 & \datasetareview{7.32} & \datasetareview{9.37} & \datasetareview{39.57} \\
InstructPix2Pix & \datasetareview{3.83} & \datasetareview{5.57} & \datasetareview{36.62} \\
\noalign{\vskip 2pt\hrule height 0.8pt}
\end{tabular}\endgroup
\end{table}

\begin{table}[t]
\centering
\caption{Macro composites under six scoring configurations on identical outputs. Bold marks the best model in each column. Color distance, prediction resolution, and tolerance choices can change absolute scores and close model orderings.}
\label{tab:app-scoring-variants}
\begingroup\small
\setlength{\tabcolsep}{4pt}\renewcommand{\arraystretch}{1.15}
\begin{tabular}{@{}lrrrrrr@{}}
\noalign{\hrule height 0.8pt}
Model & Default & CIEDE2000 & \shortstack{Half:\\bicubic} & \shortstack{Half:\\nearest} & Tighter & Looser \\
\noalign{\vskip 2pt\hrule height 0.4pt\vskip 2pt}
GPT-Image-2 & \datasetareview{\textbf{47.13}} & \datasetareview{\textbf{50.92}} & \datasetareview{\textbf{45.03}} & \datasetareview{\textbf{45.64}} & \datasetareview{\textbf{40.81}} & \datasetareview{\textbf{51.31}} \\
FireRed-Edit & \datasetareview{21.85} & \datasetareview{26.15} & \datasetareview{20.52} & \datasetareview{20.82} & \datasetareview{19.45} & \datasetareview{25.20} \\
\draft{Qwen-Image-Edit-2511} & \datasetareview{20.47} & \datasetareview{25.68} & \datasetareview{18.97} & \datasetareview{19.44} & \datasetareview{17.56} & \datasetareview{24.92} \\
\draft{Qwen-Image-Edit} & \datasetareview{16.26} & \datasetareview{19.72} & \datasetareview{15.14} & \datasetareview{15.43} & \datasetareview{14.34} & \datasetareview{18.63} \\
FLUX.2-dev & \datasetareview{13.11} & \datasetareview{18.36} & \datasetareview{12.19} & \datasetareview{12.47} & \datasetareview{10.22} & \datasetareview{16.44} \\
BAGEL & \datasetareview{10.81} & \datasetareview{13.79} & \datasetareview{10.29} & \datasetareview{10.40} & \datasetareview{9.07} & \datasetareview{13.26} \\
Step1X-Edit & \datasetareview{9.78} & \datasetareview{11.04} & \datasetareview{9.15} & \datasetareview{9.34} & \datasetareview{8.77} & \datasetareview{11.39} \\
FLUX.2-klein & \datasetareview{8.42} & \datasetareview{12.60} & \datasetareview{7.98} & \datasetareview{8.15} & \datasetareview{6.75} & \datasetareview{11.59} \\
FLUX.1-Kontext & \datasetareview{8.78} & \datasetareview{9.49} & \datasetareview{8.13} & \datasetareview{8.32} & \datasetareview{8.28} & \datasetareview{9.35} \\
OmniGen2 & \datasetareview{7.32} & \datasetareview{8.58} & \datasetareview{7.13} & \datasetareview{7.18} & \datasetareview{6.59} & \datasetareview{8.19} \\
InstructPix2Pix & \datasetareview{3.83} & \datasetareview{4.60} & \datasetareview{3.29} & \datasetareview{3.41} & \datasetareview{2.84} & \datasetareview{5.37} \\
\noalign{\vskip 2pt\hrule height 0.8pt}
\end{tabular}\endgroup
\end{table}

\begin{table}[!htbp]
\centering
\caption{\draft{Single-parameter sweep. Each default is multiplied by 0.5, 0.75, 1.5, and 2, in that order. Setting numbers identify the points in Figure~\ref{fig:app-tolerance}. All other constants remain fixed.}}
\label{tab:app-parameter-settings}
\small
\begin{tabular}{lrr}
\hline
\draft{Parameter} & \draft{Default} & \draft{Settings} \\
\hline
\draft{Fidelity tolerances} & \draft{$\{5,10,20\}$} & \draft{1--4} \\
\draft{Preservation tolerances} & \draft{$\{2,5,10,20\}$} & \draft{5--8} \\
\draft{Effective-change threshold} & \draft{6} & \draft{9--12} \\
\draft{Global-fit reference-separation cap} & \draft{8} & \draft{13--16} \\
\draft{Minimum relative change / fit residual} & \draft{0.05} & \draft{17--20} \\
\draft{Minimum local coherence} & \draft{0.25} & \draft{21--24} \\
\draft{Amplitude range for coherence check} & \draft{0.9} & \draft{25--28} \\
\draft{Background-component area fraction} & \draft{0.0005} & \draft{29--32} \\
\draft{Localization background cap} & \draft{4} & \draft{33--36} \\
\draft{Blur-match residual cap} & \draft{2} & \draft{37--40} \\
\draft{Magnitude amplitude agreement band} & \draft{0.15} & \draft{41--44} \\
\hline
\end{tabular}
\end{table}

\subsection{Controlled Change Amplitudes}
\label{sec:app-controlled}
\textbf{Setup.}
We vary pixel-change amplitude on the fixed 101-case panel described in
Appendix~\ref{sec:app-paintbench-matched}. Each case contributes six
constructed variants and a reference output, with color measured on the
same 50 eligible cases throughout. The multiplier changes pixel amplitude,
not semantic task completion.

\begin{figure}[!htbp]
  \centering
  \includegraphics[width=\linewidth]{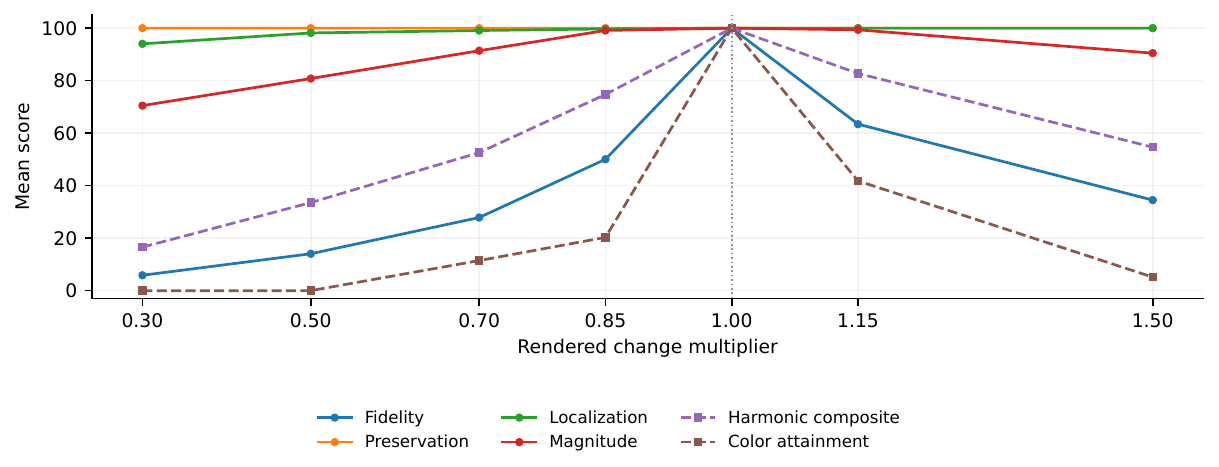}\par\nointerlineskip
  \caption{\datasetareview{Response to controlled rendered changes on 101 fixed cases.
  Outputs are $\mathrm{clip}(I+\alpha(T-I),0,255)$, using six change
  multipliers and the reference target at $\alpha=1$.
  Curves average cases equally; color attainment covers the same 50 eligible
  cases at every setting. $\alpha$ controls pixel-change amplitude, not the
  fraction of a semantic task completed.}}
  \label{fig:exp-controlled}
\end{figure}

\textbf{Results and analysis.}
Figure~\ref{fig:exp-controlled} and Table~\ref{tab:app-controlled} show that
fidelity and magnitude rise as the output approaches the reference;
overshooting lowers them while localization remains perfect. Preservation
remains nearly perfect because changes outside the allowed region are
negligible. The axes retain credit for correct spatial support while
exposing appearance and amplitude errors.

\begin{table}[t]
\centering
\caption{Complete means for controlled change amplitudes. F, P, L, M and H each cover 101 cases; color attainment C covers the same 50 cases for every multiplier. No best-model marking applies to constructed inputs.}
\label{tab:app-controlled}
\small
\setlength{\tabcolsep}{4pt}
\begin{tabular}{lrrrrrr}
\hline
$\alpha$ & F & P & L & M & H & C \\
\hline
0.3 & \datasetareview{5.85} & 100.00 & \datasetareview{94.02} & \datasetareview{70.49} & \datasetareview{16.57} & \revisionblue{0.00} \\
0.5 & \datasetareview{14.05} & 100.00 & \datasetareview{98.16} & \datasetareview{80.80} & \datasetareview{33.55} & \revisionblue{0.00} \\
0.7 & \datasetareview{27.86} & 100.00 & \draft{99.09} & \datasetareview{91.40} & \datasetareview{52.59} & \datasetareview{11.46} \\
0.85 & \datasetareview{50.05} & 100.00 & 99.69 & \datasetareview{99.11} & \datasetareview{74.74} & \datasetareview{20.30} \\
1 & 100.00 & 100.00 & 100.00 & 100.00 & 100.00 & \revisionblue{100.00} \\
1.15 & \datasetareview{63.42} & 100.00 & 100.00 & \draft{99.36} & \datasetareview{82.73} & \datasetareview{41.80} \\
1.5 & \datasetareview{34.50} & 100.00 & 100.00 & \datasetareview{90.47} & \datasetareview{54.65} & \revisionblue{5.17} \\
\hline
\end{tabular}
\end{table}

\textbf{Local errors and small perturbations.}

\textbf{Setup.}
A fixed panel takes five cases per asset--operation cell, prioritizing
source diversity. Each condition in Table~\ref{tab:app-local-controls}
uses all 110 cases. We vary the fraction of target-region pixels edited,
the pixel-change amplitude, and background color, and apply blur, JPEG
compression, shifts, 1\% centered rescaling, and deterministic reconstruction
through the Stable Diffusion v1.4 variational autoencoder (VAE).
Shifting the entire target moves protected content; shifting only the
target-minus-input difference isolates displacement of the requested change.
Rescaling uses bilinear interpolation and center cropping or edge padding;
the VAE uses the latent mode in float32.

\textbf{Results and analysis.}
Replacing half the allowed pixels with target values preserves the background
and earns partial fidelity and localization. A background color offset lowers
preservation and localization even when the target region is correct;
fidelity can also fall near the boundary because its comparison is smoothed.
Editing half the region and applying half the change amplitude produce
distinct profiles, separating spatial coverage from change strength.

Mild blur leaves the scores near the reference, whereas compression,
displacement, and rescaling lower appearance and spatial agreement
(Table~\ref{tab:app-local-controls}). VAE reconstruction preserves magnitude
more closely than target appearance: fidelity has a median of \datasetareview{91.57}, with
8 of 110 cases below 40. These responses distinguish agreement with the
reference appearance from the strength of the visual change, and quantify
their sensitivity to image transformations.
\aSixFix{The scoring stage itself applies no resampling: editor outputs are
restored to the reference coordinates by the shared wrapper before scoring
(Appendix~\ref{sec:app-inference}), and the constructed variants inherit the
canvas dimensions. These rows therefore quantify transformations that arise
editor-side in deployed editors, not artifacts of the evaluation pipeline.}

\begin{table}[!htbp]
\centering
\caption{\draft{Controlled changes on \revisionblue{110 fixed development cases} (five per asset--operation cell). Means retain all 110 cases per row. Shifted differences modify the source by a translated target-minus-source field; whole-image shifts also move protected content.}}
\label{tab:app-local-controls}
\small
\setlength{\tabcolsep}{5pt}
\begin{tabular}{lrrrr}
\hline
\draft{Constructed output} & \draft{F} & \draft{P} & \draft{L} & \draft{M} \\
\hline
\draft{Half of target-region pixels} & \datasetareview{50.17} & \draft{100.00} & \datasetareview{66.65} & \datasetareview{75.03} \\
\draft{Correct target; background RGB +20} & \datasetareview{93.51} & \datasetareview{75.89} & \datasetareview{54.86} & \draft{100.00} \\
\draft{Half change amplitude} & \datasetareview{19.89} & \draft{100.00} & \datasetareview{93.72} & \datasetareview{82.24} \\
\draft{1.5 times change amplitude} & \datasetareview{49.07} & \draft{99.84} & \draft{100.00} & \datasetareview{93.51} \\
\draft{Target, JPEG quality 95} & \datasetareview{90.10} & \datasetareview{98.89} & \datasetareview{87.21} & \draft{99.73} \\
\draft{Target, Gaussian $\sigma=0.4$} & \draft{99.99} & \datasetareview{99.61} & \datasetareview{98.89} & \draft{99.99} \\
\draft{Whole target shifted 0.25 px} & \datasetareview{94.98} & \datasetareview{99.38} & \datasetareview{96.99} & \draft{99.92} \\
\draft{Edit difference shifted 0.25 px} & \datasetareview{92.25} & \draft{99.92} & \datasetareview{99.64} & \draft{99.91} \\
\draft{Whole target shifted 0.5 px} & \datasetareview{88.44} & \datasetareview{99.12} & \datasetareview{95.65} & \draft{99.68} \\
\draft{Edit difference shifted 0.5 px} & \datasetareview{83.61} & \datasetareview{99.90} & \datasetareview{99.06} & \datasetareview{99.54} \\
\draft{Whole target shifted 1 px} & \datasetareview{80.54} & \datasetareview{98.91} & \datasetareview{91.18} & \datasetareview{98.75} \\
\draft{Edit difference shifted 1 px} & \datasetareview{73.23} & \datasetareview{99.87} & \datasetareview{94.72} & \datasetareview{97.70} \\
\revisionred{Target, JPEG quality 85} & \datasetareview{87.93} & \datasetareview{98.12} & \datasetareview{84.45} & \datasetareview{99.64} \\
\revisionred{Whole target shifted 2 px} & \datasetareview{73.42} & \datasetareview{98.31} & \datasetareview{84.52} & \datasetareview{97.28} \\
\revisionred{Edit difference shifted 2 px} & \datasetareview{63.12} & \datasetareview{99.74} & \datasetareview{89.27} & \datasetareview{92.97} \\
\revisionred{Target rescaled by $1.01\times$} & \datasetareview{63.06} & \datasetareview{94.91} & \datasetareview{64.10} & \datasetareview{92.92} \\
\revisionred{Target rescaled by $0.99\times$} & \datasetareview{62.45} & \datasetareview{96.78} & \datasetareview{64.70} & \datasetareview{92.91} \\
\revisionred{Target, SD v1.4 VAE encode--decode} & \datasetareview{82.71} & \datasetareview{93.28} & \datasetareview{81.12} & \datasetareview{99.51} \\
\hline
\end{tabular}
\end{table}

\textbf{Known source-level errors.}

\textbf{Setup.}
To test explicit editing mistakes beyond pixel perturbations, we render
seven constructions on eight fixed source assets (two per type): the exact
target, unchanged input, an omitted color group, a different group, an extra
group, swapped target colors, and background damage. Assets are fixed by
source ID before scoring; candidate color declarations must change visible
pixels. Each target is rendered twice to verify repeatability.

\textbf{Results and analysis.}
Table~\ref{tab:app-source-controls} shows complementary responses:
omission reduces target agreement and coverage, whereas extra edits damage
preservation. Swapping colors largely preserves localization but reduces
fidelity, so magnitude alone does not establish correct assignment.

\begin{table}[!ht]
\centering
\caption{\revisionblue{Source-rendered controls on eight fixed assets (two per type). Values average each per-asset score equally. Groups are defined by source color declarations; they need not be semantic objects.}}
\label{tab:app-source-controls}
\vspace{3pt}
{\small\setlength{\tabcolsep}{4pt}
\begin{tabular}{lrrrrr}
\hline
Construction & $F$ & $P$ & $L$ & $M$ & $H$ \\
\hline
Exact target & 100.00 & 100.00 & 100.00 & 100.00 & 100.00 \\
Unchanged input & 0.00 & 100.00 & 0.00 & 0.00 & 0.00 \\
Omit one group & 49.48 & 100.00 & 69.79 & 75.97 & 62.36 \\
Edit another group & 0.00 & 71.66 & 0.00 & 0.00 & 0.00 \\
Edit an extra group & 96.40 & 71.67 & 73.45 & 99.92 & 71.44 \\
Swap target colors & 2.89 & 100.00 & 99.90 & 68.80 & 6.78 \\
Damage background & 62.54 & 70.44 & 74.43 & 98.79 & 70.05 \\
\hline
\end{tabular}}
\end{table}

\subsection[Effects of Shared Checks and the Composite]{\draft{Effects of Shared Checks and the Composite}}
 \label{sec:app-check-effects}
\begin{figure}[!htbp]
\centering
\includegraphics[width=\linewidth]{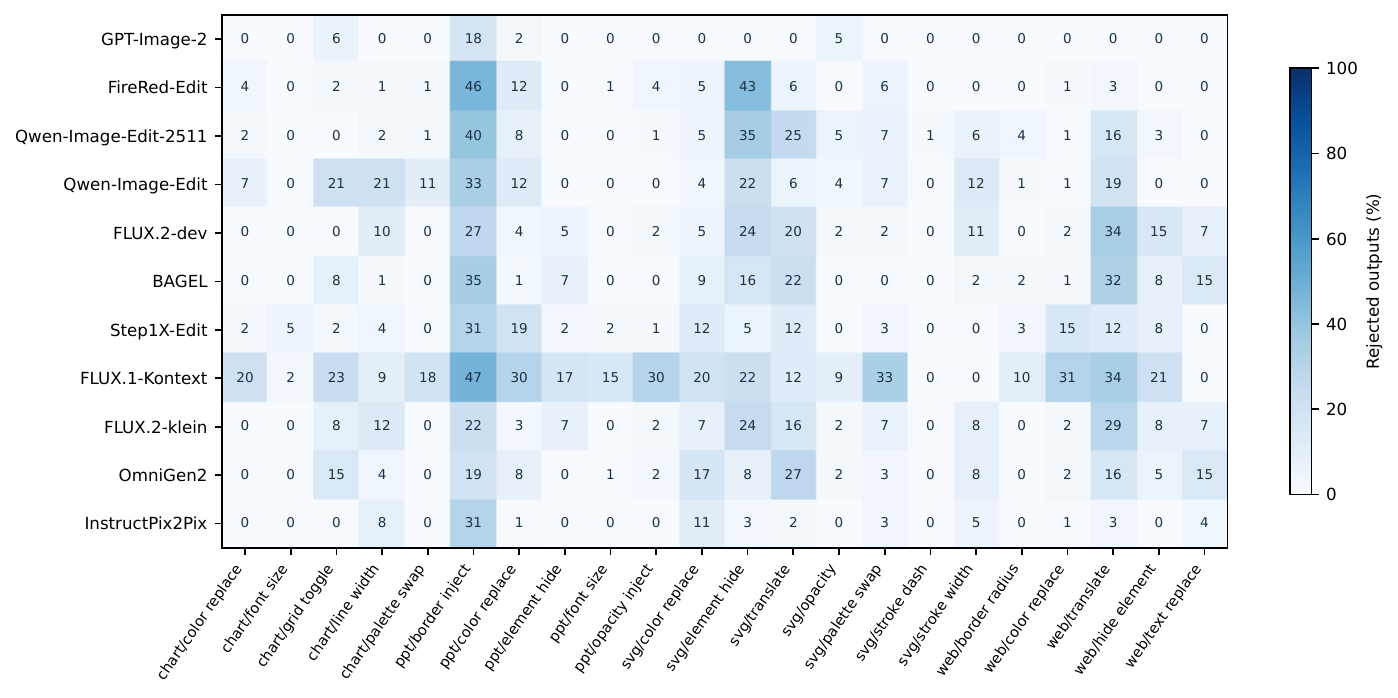}
\caption{\datasetareview{Shared-check rejection rates for every model--operation
combination. Each cell uses all outputs in that combination as its
denominator. Values are percentages, rounded for display. Rejection leaves
preservation independently measured.}}
\label{fig:app-followup-gates}
\end{figure}

\textbf{Setup.}
We examine all 19,140 outputs and remove the shared rejection check and the
localization background cap separately, keeping all other rules fixed.
Table~\ref{tab:app-gate-conditions} separates the first triggered condition,
which determines the reported score, from independent triggers, which may
overlap. Tables~\ref{tab:app-followup-gate} and~\ref{tab:app-followup-cap}
report all axes and the primary score. The model--operation breakdown in
Figure~\ref{fig:app-followup-gates} shows where rejection occurs.
 
 \begin{table}[!htbp]
\centering
\caption{\draft{Shared-check triggers on all 19,140 outputs. First-trigger counts follow the execution order and are mutually exclusive. Independent counts evaluate each condition separately and can overlap. Passing is not proof of correct target appearance.}}
\label{tab:app-gate-conditions}
\vspace{3pt}
\scriptsize
\setlength{\tabcolsep}{4pt}
\begin{tabular}{lrr}
\hline
\draft{Condition} & \draft{First trigger} & \draft{Independent} \\
\hline
\draft{Empty region} & \draft{0} & \draft{0} \\
\draft{Negligible change} & \datasetareview{1122} & \datasetareview{1122} \\
\draft{Blur match} & \datasetareview{95} & \datasetareview{997} \\
\draft{Local cancellation} & \draft{114} & \draft{117} \\
\draft{Global color transformation} & \datasetareview{61} & \datasetareview{384} \\
\draft{Passes all checks} & \datasetareview{17748} & \draft{---} \\
\hline
\end{tabular}
\end{table}

\textbf{Results and analysis.}
Table~\ref{tab:app-H-axes} reports each model's rejection rate,
zero-composite rate, and the change in the primary score when the check is
removed. The change never exceeds 0.05 points, so the check does not drive
the ranking; rejection rates stay below 10\% for all models except
FLUX.1-Kontext (\datasetareview{21\%}). 
 
 \begin{table}[!htbp]
\centering
\caption{Effect of the shared check. Rejected and $H=0$ are percentages of 1,740 outputs per model; $\Delta H$ is the change in the 22-cell mean when the check is removed.}
\label{tab:app-H-axes}
\vspace{3pt}
\scriptsize
\setlength{\tabcolsep}{4pt}
\begin{tabular}{lrrr}
\hline
\draft{Model} & \draft{Rejected} & \draft{H=0} & \draft{$\Delta H$} \\
\hline
\draft{GPT-Image-2} & \datasetareview{1.3} & \datasetareview{21.0} & \draft{0.00} \\
\draft{FireRed-Edit} & \datasetareview{5.5} & \datasetareview{52.5} & \draft{0.00} \\
\draft{Qwen-Image-Edit-2511} & \datasetareview{6.7} & \datasetareview{54.7} & \draft{0.02} \\
\draft{Qwen-Image-Edit} & \datasetareview{8.0} & \datasetareview{65.9} & \draft{0.01} \\
\draft{FLUX.2-dev} & \datasetareview{7.1} & \datasetareview{66.3} & \draft{0.01} \\
\draft{BAGEL} & \datasetareview{6.4} & \datasetareview{73.6} & \draft{0.00} \\
\draft{Step1X-Edit} & \datasetareview{7.6} & \datasetareview{74.6} & \draft{0.01} \\
\draft{FLUX.2-klein} & \revisionblue{6.9} & \datasetareview{75.7} & \draft{0.05} \\
\draft{FLUX.1-Kontext} & \datasetareview{21.0} & \datasetareview{77.5} & \datasetareview{0.01} \\
\draft{OmniGen2} & \datasetareview{6.3} & \datasetareview{78.1} & \datasetareview{0.01} \\
\draft{InstructPix2Pix} & \datasetareview{3.2} & \datasetareview{84.9} & \draft{0.02} \\
\hline
\end{tabular}
\end{table}

\textbf{Sources of zero composites.}
Table~\ref{tab:app-zero-sources} decomposes the $H=0$ outputs by the axis
that vanishes. Outputs rejected by the shared check ($F=L=M=0$) number
\datasetareview{1,392}, matching the first-trigger rejections in
Table~\ref{tab:app-gate-conditions}. Of the \datasetareview{12,611} zero composites,
\datasetareview{10,998} (87\%) pass the check and produce a measurable change
($L,M>0$) yet do not raise agreement with the target above the
unchanged-input baseline; zero preservation accounts for 190 and the remaining combinations
for 31. Zero composites therefore arise predominantly from fidelity, not
from the check discarding the output.

\begin{table}[!ht]
\centering
\caption{Composition of zero composites among each model's 1,740 outputs. Categories are mutually exclusive and assigned in order: rejected by the shared check ($F=L=M=0$); zero preservation ($P=0$); zero fidelity only ($F=0$ with $L,M>0$), meaning agreement with the target does not exceed the unchanged-input baseline; and other combinations. The rejected total equals the first-trigger rejections in Table~\ref{tab:app-gate-conditions}.}
\label{tab:app-zero-sources}
\vspace{3pt}
\small
\setlength{\tabcolsep}{5pt}
\begin{tabular}{lrrrrr}
\hline
Model & $H=0$ & Check rejected & Zero $P$ & Zero $F$ only & Other \\
\hline
GPT-Image-2 & \datasetareview{365} & \datasetareview{23} & 20 & \datasetareview{322} & 0 \\
FireRed-Edit & \datasetareview{914} & \datasetareview{96} & 13 & \datasetareview{800} & 5 \\
Qwen-Image-Edit-2511 & \datasetareview{951} & \datasetareview{116} & 15 & \datasetareview{816} & 4 \\
Qwen-Image-Edit & \datasetareview{1146} & \datasetareview{139} & 21 & \datasetareview{982} & 4 \\
FLUX.2-dev & \datasetareview{1154} & \datasetareview{124} & 9 & \datasetareview{1016} & 5 \\
BAGEL & \datasetareview{1281} & \datasetareview{111} & 29 & \datasetareview{1141} & 0 \\
Step1X-Edit & \datasetareview{1298} & \datasetareview{132} & 4 & \datasetareview{1156} & 6 \\
FLUX.2-klein & \datasetareview{1317} & 120 & 7 & \datasetareview{1189} & 1 \\
FLUX.1-Kontext & \datasetareview{1348} & \datasetareview{366} & 0 & \datasetareview{977} & 5 \\
OmniGen2 & \datasetareview{1359} & \datasetareview{109} & 8 & \datasetareview{1241} & 1 \\
InstructPix2Pix & \datasetareview{1478} & \datasetareview{56} & 64 & \datasetareview{1358} & 0 \\
\hline
All models & \datasetareview{12,611} & \datasetareview{1,392} & 190 & \datasetareview{10,998} & 31 \\
\hline
\end{tabular}
\end{table}

\datasetareview{Removing the cap lowers scores while preserving the full model ranking.}
Paired source-group intervals in Table~\ref{tab:app-followup-comparisons}
separate uncertainty in model gaps from changes induced by this scoring
choice.

\begin{table}[!ht]
\centering
\caption{\revisionblue{Without the shared check. Identical 19,140 outputs; axes are case means and $H$ is the 22-cell macro mean. $\Delta H$ is relative to the fixed default, with all other rules unchanged.}}
\label{tab:app-followup-gate}
\vspace{3pt}
{\small\setlength{\tabcolsep}{4pt}
\begin{tabular}{lrrrrrr}
\hline
Model & $F$ & $P$ & $L$ & $M$ & $H$ & $\Delta H$ \\
\hline
GPT-Image-2 & \datasetareview{37.10} & \datasetareview{84.52} & \datasetareview{56.97} & \datasetareview{84.60} & \datasetareview{47.14} & +0.002 \\
FireRed-Edit & \datasetareview{13.42} & \datasetareview{85.09} & \datasetareview{44.57} & \datasetareview{70.78} & \datasetareview{21.86} & +0.005 \\
Qwen-Image-Edit-2511 & \datasetareview{12.23} & \datasetareview{87.16} & \datasetareview{47.21} & \datasetareview{69.29} & \datasetareview{20.49} & +0.015 \\
Qwen-Image-Edit & \datasetareview{9.39} & \datasetareview{82.06} & \datasetareview{44.78} & \datasetareview{63.23} & \datasetareview{16.27} & \datasetareview{+0.006} \\
FLUX.2-dev & \datasetareview{8.17} & \datasetareview{82.61} & \datasetareview{44.26} & \datasetareview{64.36} & \datasetareview{13.12} & +0.012 \\
BAGEL & \datasetareview{6.66} & \datasetareview{68.75} & \datasetareview{43.21} & \datasetareview{63.84} & \datasetareview{10.81} & +0.003 \\
Step1X-Edit & \datasetareview{5.24} & \datasetareview{82.78} & \datasetareview{41.07} & \datasetareview{62.61} & \datasetareview{9.79} & +0.007 \\
FLUX.2-klein & \datasetareview{4.86} & \datasetareview{78.81} & \datasetareview{39.76} & \datasetareview{63.03} & \datasetareview{8.47} & +0.046 \\
FLUX.1-Kontext & \datasetareview{3.78} & \datasetareview{91.64} & \datasetareview{31.66} & \datasetareview{55.23} & \datasetareview{8.79} & \datasetareview{+0.013} \\
OmniGen2 & \datasetareview{5.17} & \datasetareview{60.52} & \datasetareview{36.07} & \datasetareview{59.18} & \datasetareview{7.33} & \datasetareview{+0.013} \\
InstructPix2Pix & \datasetareview{2.37} & \datasetareview{39.19} & \datasetareview{42.74} & \datasetareview{65.50} & \datasetareview{3.84} & +0.017 \\
\hline
\end{tabular}}
\end{table}

\begin{table}[!ht]
\centering
\caption{\revisionblue{Without the localization background cap. Identical 19,140 outputs; axes are case means and $H$ is the 22-cell macro mean. $\Delta H$ is relative to the fixed default, with all other rules unchanged.}}
\label{tab:app-followup-cap}
\vspace{3pt}
{\small\setlength{\tabcolsep}{4pt}
\begin{tabular}{lrrrrrr}
\hline
Model & $F$ & $P$ & $L$ & $M$ & $H$ & $\Delta H$ \\
\hline
GPT-Image-2 & \datasetareview{37.10} & \datasetareview{84.52} & \datasetareview{53.17} & \datasetareview{84.09} & \datasetareview{44.75} & \datasetareview{-2.389} \\
FireRed-Edit & \datasetareview{13.42} & \datasetareview{85.09} & \datasetareview{39.80} & \datasetareview{68.76} & \datasetareview{20.03} & \datasetareview{-1.824} \\
Qwen-Image-Edit-2511 & \datasetareview{12.22} & \datasetareview{87.16} & \datasetareview{43.23} & \datasetareview{66.92} & \datasetareview{19.04} & \datasetareview{-1.434} \\
Qwen-Image-Edit & \datasetareview{9.38} & \datasetareview{82.06} & \datasetareview{39.58} & \datasetareview{60.53} & \datasetareview{14.40} & \datasetareview{-1.864} \\
FLUX.2-dev & \datasetareview{8.16} & \datasetareview{82.61} & \datasetareview{38.38} & \datasetareview{61.65} & \datasetareview{12.08} & \datasetareview{-1.029} \\
BAGEL & \datasetareview{6.66} & \datasetareview{68.75} & \datasetareview{34.34} & \datasetareview{61.38} & \datasetareview{9.48} & \datasetareview{-1.327} \\
Step1X-Edit & \datasetareview{5.24} & \datasetareview{82.78} & \datasetareview{35.07} & \datasetareview{59.47} & \datasetareview{8.95} & \datasetareview{-0.825} \\
FLUX.2-klein & \datasetareview{4.85} & \datasetareview{78.81} & \datasetareview{32.91} & \datasetareview{60.34} & \datasetareview{7.60} & \datasetareview{-0.821} \\
FLUX.1-Kontext & \datasetareview{3.78} & \datasetareview{91.64} & \datasetareview{26.17} & \datasetareview{46.69} & \datasetareview{8.11} & \datasetareview{-0.670} \\
OmniGen2 & \datasetareview{5.16} & \datasetareview{60.52} & \datasetareview{24.99} & \datasetareview{56.80} & \datasetareview{4.95} & \datasetareview{-2.363} \\
InstructPix2Pix & \datasetareview{2.36} & \datasetareview{39.19} & \datasetareview{26.77} & \datasetareview{64.16} & \datasetareview{2.93} & \datasetareview{-0.900} \\
\hline
\end{tabular}}
\end{table}

\begin{table}[!ht]
\centering
\caption{\revisionblue{Pairwise gaps without the localization cap. Pairs follow the default ordering; intervals use 10,000 paired source-group resamples. The final column measures the change in each gap relative to the default.}}
\label{tab:app-followup-comparisons}
\vspace{3pt}
{\small\setlength{\tabcolsep}{4pt}
\begin{tabular}{lrrr}
\hline
\datasetareview{Model pair} & \datasetareview{$\Delta H$} & \datasetareview{95\% interval} & \datasetareview{Gap change interval} \\
\hline
\datasetareview{GPT-Image-2 / FireRed-Edit} & \datasetareview{+24.72} & \datasetareview{[22.65, 26.66]} & \datasetareview{[-0.91, -0.21]} \\
\datasetareview{FireRed-Edit / Qwen-Image-Edit-2511} & \datasetareview{+0.99} & \datasetareview{[-0.21, 2.20]} & \datasetareview{[-0.69, -0.06]} \\
\datasetareview{Qwen-Image-Edit-2511 / Qwen-Image-Edit} & \datasetareview{+4.64} & \datasetareview{[3.41, 5.98]} & \datasetareview{[-0.03, 0.92]} \\
\datasetareview{Qwen-Image-Edit / FLUX.2-dev} & \datasetareview{+2.32} & \datasetareview{[0.85, 3.83]} & \datasetareview{[-1.32, -0.32]} \\
\datasetareview{FLUX.2-dev / BAGEL} & \datasetareview{+2.60} & \datasetareview{[1.03, 4.23]} & \datasetareview{[-0.26, 0.81]} \\
\datasetareview{BAGEL / Step1X-Edit} & \datasetareview{+0.53} & \datasetareview{[-0.97, 1.86]} & \datasetareview{[-1.03, 0.02]} \\
\datasetareview{Step1X-Edit / FLUX.1-Kontext} & \datasetareview{+0.85} & \datasetareview{[-0.35, 2.14]} & \datasetareview{[-0.52, 0.23]} \\
\datasetareview{FLUX.1-Kontext / FLUX.2-klein} & \datasetareview{+0.50} & \datasetareview{[-0.88, 1.84]} & \datasetareview{[-0.23, 0.55]} \\
\datasetareview{FLUX.2-klein / OmniGen2} & \datasetareview{+2.65} & \datasetareview{[1.15, 4.21]} & \datasetareview{[0.95, 2.17]} \\
\datasetareview{OmniGen2 / InstructPix2Pix} & \datasetareview{+2.03} & \datasetareview{[0.90, 3.20]} & \datasetareview{[-2.21, -0.79]} \\
\hline
\end{tabular}}
\end{table}

\subsection{Generation and Instruction Controls}
\label{sec:app-controls}
\textbf{Setup.}
We evaluate FireRed-Edit and FLUX.2-dev on the same 94 cases with four seeds,
holding checkpoints and sampling settings fixed. For instruction wording,
we append displacement as a percentage of canvas size on 169 translation
cases for FLUX.2-klein and FLUX.1-Kontext.

\textbf{Results and analysis.}
The seed score ranges remain separated (Table~\ref{tab:app-seeds}), showing
that the observed lead exceeds seed variation on this panel. Adding the
percentage description changes case-mean $H$ by less than half a point
for both editors (Table~\ref{tab:app-wording}).

\begin{figure}[!htbp]
\centering
\includegraphics[width=\linewidth]{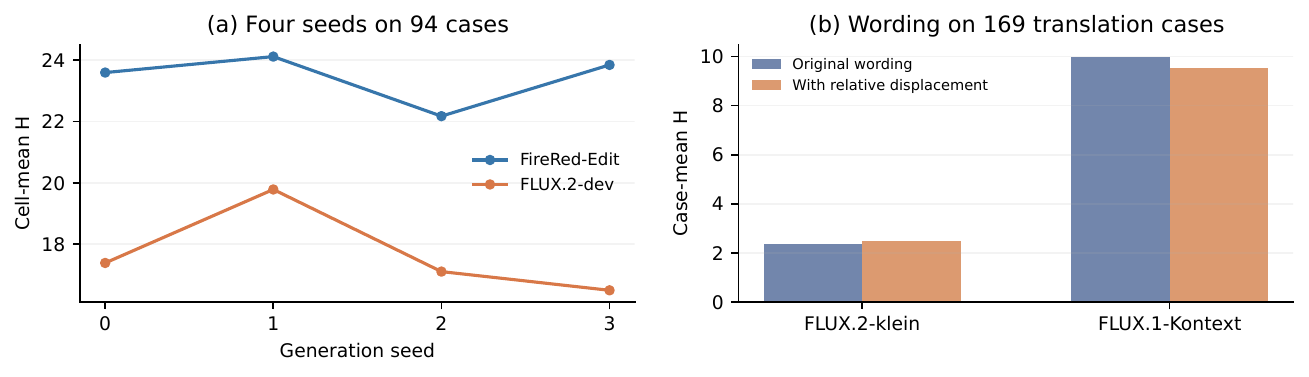}
\caption{\datasetareview{Generation controls on their actual subsets: four seeds on 94
cases for FireRed-Edit and FLUX.2-dev, and two instruction forms on 169 translation
cases for FLUX.2-klein and FLUX.1-Kontext. Seed scores use cell-macro aggregation;
wording scores use case-weighted means.}}
\label{fig:app-generation}
\end{figure}
\begin{table}[t]
\centering
\caption{\draft{All four seed runs on the same 94 cases. Scores average the covered operation cells equally; bold marks the larger score within each seed.}}
\label{tab:app-seeds}
\small
\begin{tabular}{lrrrr}
\hline
Model & Seed 0 & Seed 1 & Seed 2 & Seed 3 \\
\hline
\draft{FireRed-Edit} & \datasetareview{\textbf{23.60}} & \datasetareview{\textbf{24.12}} & \datasetareview{\textbf{22.18}} & \datasetareview{\textbf{23.85}} \\
\draft{FLUX.2-dev} & \datasetareview{17.39} & \datasetareview{19.79} & \datasetareview{17.11} & \datasetareview{16.50} \\
\hline
\end{tabular}
\end{table}

\begin{table}[t]
\centering
\caption{\draft{Matched wording control on 169 cases per model. The same source image and requested edit are retained; the alternative adds displacement as a percentage of the canvas size. Scores are case-mean H; bold marks the larger model score for each instruction form.}}
\label{tab:app-wording}
\small
\begin{tabular}{lrrr}
\hline
Model & Pixel instruction & With percentage & \draft{Change} \\
\hline
\draft{FLUX.2-klein} & \draft{2.37} & \draft{2.48} & \draft{+0.11} \\
\draft{FLUX.1-Kontext} & \draft{\textbf{9.97}} & \draft{\textbf{9.55}} & \draft{-0.42} \\
\hline
\end{tabular}
\end{table}

\subsection{Same-operation Comparison Across Assets}
\label{sec:app-shared-operation}
\textbf{Setup.}
The SVG/chart/web/slide recoloring subsets contain
82/54/183/136 cases from 35/14/40/29 sources. Their median allowed areas are
3.92/\datasetareview{7.74}/3.00/1.78\% of the image.

Within each asset type, we draw 10,000 bootstrap resamples of source assets,
keeping cases from the same source together and sharing the draws across
models. We report pointwise 95\% percentile intervals for case-mean $H$
and paired model differences. The coarse area strata receive equal weight; resamples missing a required
area stratum are excluded from the area-standardized intervals. 

Table~\ref{tab:app-area-strata} describes the four fixed finer bins
($\le1\%$, 1--5\%, 5--10\%, and $>25\%$), retaining \datasetareview{402} cases.
Their sparse-bin means are descriptive and receive no confidence intervals.

\textbf{Results and analysis.}
Table~\ref{tab:app-common-operation} supplies the four-axis breakdown.
The FLUX.2-dev--FireRed-Edit ranking reversal between charts and SVG persists
after area standardization (Table~\ref{tab:app-area-standardization}).
Area weighting adjusts region-size mixture without equating source content,
colors, or layout.

\begingroup
\makeatletter\def\fnum@table{\normalsize\tablename\space\thetable}\makeatother
\scriptsize
\setlength{\tabcolsep}{4pt}
\setlength{\LTcapwidth}{\linewidth}
\setlength{\LTpre}{6pt}
\setlength{\LTpost}{6pt}
\begin{longtable}{llrrrrr}
\caption{\normalsize Recoloring on the same operation across asset types. Entries give mean $H$ [95\% source-cluster bootstrap interval] and the four mean axes. All 455 cases and eleven models are retained. Bold marks the highest mean per asset and measurement.}\label{tab:app-common-operation}\\
\hline
Asset & Model & $H$ [95\% CI] & $F$ & $P$ & $L$ & $M$ \\

\hline
\endfirsthead
\caption[]{\normalsize Recoloring across asset types (continued).}\\
\hline
Asset & Model & $H$ [95\% CI] & $F$ & $P$ & $L$ & $M$ \\

\hline
\endhead
\hline
\endfoot
\endlastfoot
SVG (82) & GPT-Image-2 & \draft{\textbf{76.78} [70.18, 82.43]} & \draft{\textbf{64.99}} & 87.73 & \draft{\textbf{85.47}} & \draft{\textbf{95.78}} \\*
 & \draft{FireRed-Edit} & \draft{15.53 [10.55, 20.93]} & \draft{8.78} & 88.67 & \draft{69.30} & \draft{74.07} \\*
 & \draft{Qwen-Image-Edit-2511} & \draft{15.98 [10.21, 21.82]} & 9.23 & \textbf{89.81} & \draft{66.88} & \draft{80.64} \\*
 & FLUX.2-dev & 8.43 [4.53, 12.61] & 5.48 & 70.88 & \draft{59.28} & \draft{76.37} \\*
 & \draft{Qwen-Image-Edit} & \draft{11.99 [7.27, 17.27]} & \draft{7.74} & 76.24 & \draft{55.87} & \draft{71.59} \\*
 & BAGEL & \draft{6.35 [3.06, 9.84]} & \draft{3.39} & 70.65 & \draft{56.15} & \draft{69.89} \\*
 & FLUX.2-klein & \draft{11.44 [7.13, 16.25]} & \draft{6.81} & 62.27 & \draft{48.42} & \draft{72.96} \\*
 & \draft{Step1X-Edit} & 3.09 [0.82, 5.81] & 2.05 & 67.89 & 44.58 & 53.21 \\*
 & \draft{FLUX.1-Kontext} & \draft{1.67 [0.26, 3.76]} & \draft{0.74} & 81.16 & \draft{32.22} & \draft{49.09} \\*
 & OmniGen2 & 2.63 [0.15, 5.48] & 2.15 & 37.30 & \draft{38.16} & \draft{51.09} \\*
 & InstructPix2Pix & \draft{0.82 [0.05, 2.34]} & \draft{0.84} & 25.86 & \draft{45.40} & \draft{57.39} \\
\hline
Chart (54) & GPT-Image-2 & \datasetareview{\textbf{68.68} [53.97, 81.52]} & \datasetareview{\textbf{53.29}} & \datasetareview{95.68} & \datasetareview{\textbf{80.40}} & \datasetareview{\textbf{96.69}} \\*
 & \draft{FireRed-Edit} & \datasetareview{5.05 [2.16, 8.42]} & \datasetareview{1.69} & \datasetareview{\textbf{96.75}} & \datasetareview{67.74} & \datasetareview{75.72} \\*
 & \draft{Qwen-Image-Edit-2511} & \datasetareview{6.48 [3.17, 9.86]} & \datasetareview{2.29} & \datasetareview{96.74} & \datasetareview{70.10} & \datasetareview{77.06} \\*
 & FLUX.2-dev & \datasetareview{17.39 [11.68, 23.41]} & \datasetareview{7.05} & \datasetareview{96.47} & \datasetareview{79.71} & \datasetareview{84.98} \\*
 & \draft{Qwen-Image-Edit} & \datasetareview{3.22 [0.97, 5.98]} & \datasetareview{1.16} & \datasetareview{93.75} & \datasetareview{57.04} & \datasetareview{58.30} \\*
 & BAGEL & \datasetareview{0.99 [0.10, 2.30]} & \datasetareview{0.31} & \datasetareview{89.38} & \datasetareview{74.13} & \datasetareview{73.63} \\*
 & FLUX.2-klein & \datasetareview{7.27 [3.63, 11.31]} & \datasetareview{2.49} & \datasetareview{96.60} & \datasetareview{76.25} & \datasetareview{78.40} \\*
 & \draft{Step1X-Edit} & \datasetareview{0.87 [0.00, 2.34]} & \datasetareview{0.31} & \datasetareview{95.81} & \datasetareview{60.15} & \datasetareview{62.04} \\*
 & \draft{FLUX.1-Kontext} & \datasetareview{1.09 [0.00, 2.77]} & \datasetareview{0.34} & \datasetareview{96.10} & \datasetareview{34.36} & \datasetareview{43.77} \\*
 & OmniGen2 & \datasetareview{1.39 [0.45, 2.45]} & \datasetareview{0.44} & \datasetareview{58.50} & \datasetareview{45.96} & \datasetareview{63.18} \\*
 & InstructPix2Pix & \datasetareview{1.69 [0.27, 3.44]} & \datasetareview{0.48} & \datasetareview{73.47} & \datasetareview{67.89} & \datasetareview{68.46} \\
\hline
Web (183) & GPT-Image-2 & \draft{\textbf{32.83} [25.13, 41.16]} & \draft{\textbf{30.03}} & 69.62 & \draft{\textbf{55.47}} & \draft{\textbf{89.08}} \\*
 & \draft{FireRed-Edit} & \draft{11.80 [8.14, 15.72]} & \draft{6.56} & 78.81 & \draft{50.12} & \draft{79.46} \\*
 & \draft{Qwen-Image-Edit-2511} & \draft{13.58 [8.54, 19.38]} & \draft{9.28} & 76.98 & \draft{51.80} & \draft{75.06} \\*
 & FLUX.2-dev & \draft{11.14 [7.78, 14.79]} & \draft{7.69} & 61.97 & \draft{46.87} & \draft{76.57} \\*
 & \draft{Qwen-Image-Edit} & \draft{6.90 [4.43, 9.41]} & \draft{3.98} & 64.32 & \draft{47.88} & \draft{72.59} \\*
 & BAGEL & \draft{1.39 [0.36, 2.73]} & 1.19 & 32.33 & \draft{36.13} & \draft{60.58} \\*
 & FLUX.2-klein & \draft{9.98 [6.49, 13.96]} & \draft{5.58} & 72.67 & \draft{42.23} & \draft{74.10} \\*
 & \draft{Step1X-Edit} & 1.68 [0.47, 3.22] & 0.77 & 87.94 & 31.70 & 50.37 \\*
 & \draft{FLUX.1-Kontext} & \draft{0.96 [0.41, 1.59]} & \draft{0.57} & \textbf{94.15} & \draft{21.92} & \draft{38.00} \\*
 & OmniGen2 & \draft{0.36 [0.09, 0.69]} & \draft{0.17} & 55.57 & \draft{34.86} & \draft{50.91} \\*
 & InstructPix2Pix & \draft{2.40 [0.63, 5.14]} & 1.31 & 34.23 & \draft{38.44} & \draft{63.93} \\
\hline
Slide (136) & GPT-Image-2 & \draft{\textbf{54.03} [45.43, 62.30]} & \textbf{45.22} & 85.91 & \draft{\textbf{59.28}} & \draft{\textbf{92.20}} \\*
 & \draft{FireRed-Edit} & \draft{10.68 [7.16, 14.39]} & 5.72 & 92.59 & \draft{54.98} & \draft{69.13} \\*
 & \draft{Qwen-Image-Edit-2511} & \draft{14.60 [9.20, 20.50]} & 9.19 & 90.04 & \draft{54.75} & \draft{73.17} \\*
 & FLUX.2-dev & \draft{11.17 [8.08, 14.44]} & 6.43 & 67.38 & \draft{52.27} & \draft{78.08} \\*
 & \draft{Qwen-Image-Edit} & \draft{3.09 [1.45, 4.96]} & \draft{1.53} & 88.97 & \draft{50.93} & \draft{57.94} \\*
 & BAGEL & 1.99 [0.30, 4.15] & 1.08 & 48.22 & \draft{44.20} & \draft{60.09} \\*
 & FLUX.2-klein & \draft{6.69 [3.74, 10.00]} & \draft{4.67} & 64.75 & \draft{45.97} & \draft{70.76} \\*
 & \draft{Step1X-Edit} & 2.58 [1.02, 4.41] & 1.41 & 90.22 & 37.30 & 49.18 \\*
 & \draft{FLUX.1-Kontext} & \draft{0.77 [0.12, 1.72]} & \draft{0.54} & \textbf{93.06} & \draft{32.62} & \draft{40.03} \\*
 & OmniGen2 & \draft{1.80 [0.88, 2.86]} & \draft{1.21} & 51.58 & \draft{31.38} & \draft{58.82} \\*
 & InstructPix2Pix & 0.48 [0.01, 1.24] & 0.54 & 16.39 & 41.92 & \draft{67.27} \\
\hline
\end{longtable}
\endgroup

\begin{table}[t]
\centering
\caption{Cases / source assets in the recoloring area strata. Coarse strata retain all 455 cases; the four shared fine strata retain \datasetareview{402}. An asset can contribute cases to more than one stratum.}
\label{tab:app-area-strata}
\small
\setlength{\tabcolsep}{5pt}
\begin{tabular}{lrrrrrr}
\noalign{\hrule height 0.8pt}
& \multicolumn{2}{c}{Coarse area strata} & \multicolumn{4}{c}{Fine shared area strata} \\
\cline{2-3}\cline{4-7}
Asset & $\le10\%$ & \draft{$>10\%$} & \draft{$\le1\%$} & \draft{1--5\%} & \draft{5--10\%} & \draft{$>25\%$} \\
\noalign{\vskip 2pt\hrule height 0.4pt\vskip 2pt}
SVG & 51 / 28 & \draft{31 / 19} & \draft{19 / 14} & \draft{23 / 16} & \draft{9 / 8} & \draft{21 / 14} \\
Chart & \datasetareview{30 / 8} & \datasetareview{24 / 6} & \datasetareview{8 / 2} & \datasetareview{14 / 4} & \datasetareview{8 / 2} & \datasetareview{8 / 2} \\
Web & 151 / 36 & \draft{32 / 6} & \draft{57 / 17} & \draft{60 / 14} & \draft{34 / 7} & \draft{18 / 3} \\
Slide & 105 / 29 & \draft{31 / 13} & \draft{53 / 15} & \draft{27 / 10} & \draft{25 / 7} & \draft{18 / 9} \\
\noalign{\hrule height 0.8pt}
\end{tabular}
\end{table}

\begin{table}[t]
\centering
\caption{Mean $H$ after equal weighting of the area strata defined in Table~\ref{tab:app-area-strata}. Best model per asset and weighting is bold. These comparisons describe area-standardized performance, not a causal effect of asset type.}
\label{tab:app-area-standardization}
\small
\setlength{\tabcolsep}{4pt}
\begin{tabular}{lrrrrrrrr}
\noalign{\hrule height 0.8pt}
 & \multicolumn{4}{c}{Two coarse strata (all 455 cases)} & \multicolumn{4}{c}{\datasetareview{Four fine strata (402 cases)}} \\
\cline{2-5}\cline{6-9}
Model & SVG & \draft{Chart} & \draft{Web} & \draft{Slide} & \draft{SVG} & \draft{Chart} & \draft{Web} & \draft{Slide} \\
\noalign{\vskip 2pt\hrule height 0.4pt\vskip 2pt}
GPT-Image-2 & \draft{\textbf{78.19}} & \datasetareview{\textbf{70.56}} & \draft{\textbf{30.48}} & \draft{\textbf{62.33}} & \draft{\textbf{75.29}} & \datasetareview{\textbf{64.74}} & \draft{\textbf{28.13}} & \draft{\textbf{56.35}} \\
\draft{FireRed-Edit} & \draft{16.63} & \datasetareview{5.12} & \draft{10.30} & \draft{11.99} & \draft{14.97} & \datasetareview{4.52} & \draft{9.07} & \draft{10.69} \\
\draft{Qwen-Image-Edit-2511} & \draft{17.78} & \datasetareview{6.81} & \draft{10.59} & \draft{15.44} & \draft{14.59} & \datasetareview{5.28} & \draft{11.78} & \draft{17.35} \\
FLUX.2-dev & 8.88 & \datasetareview{18.15} & \draft{12.85} & \draft{12.08} & \draft{7.11} & \datasetareview{16.89} & \draft{9.61} & \draft{12.49} \\
\draft{Qwen-Image-Edit} & \draft{12.67} & \datasetareview{3.29} & \draft{8.47} & \draft{3.12} & \draft{10.60} & \datasetareview{2.86} & \draft{6.29} & \draft{3.86} \\
BAGEL & \draft{8.02} & \datasetareview{1.07} & \draft{2.74} & \draft{4.05} & \draft{6.15} & \datasetareview{0.69} & \draft{1.34} & \draft{2.59} \\
FLUX.2-klein & \draft{11.41} & \datasetareview{7.69} & \draft{12.46} & \draft{8.11} & \draft{10.79} & \datasetareview{5.52} & \draft{9.03} & \draft{7.62} \\
\draft{Step1X-Edit} & 3.52 & \datasetareview{0.97} & 2.27 & 3.33 & 2.24 & \datasetareview{1.47} & 2.23 & 3.03 \\
\draft{FLUX.1-Kontext} & \draft{1.92} & \datasetareview{0.98} & \draft{0.92} & \draft{0.58} & \draft{0.52} & \datasetareview{1.32} & \draft{1.19} & \draft{0.65} \\
OmniGen2 & 3.03 & \datasetareview{1.50} & \draft{0.36} & \draft{1.51} & \draft{3.67} & \datasetareview{1.56} & \draft{0.30} & \draft{1.33} \\
InstructPix2Pix & \draft{0.66} & \datasetareview{1.54} & \draft{1.60} & \draft{0.81} & \draft{0.82} & \datasetareview{2.12} & \draft{2.00} & \draft{0.79} \\
\noalign{\hrule height 0.8pt}
\end{tabular}
\end{table}

\subsection{Color Attainment and Component Diagnosis on the Same Outputs}
\label{sec:app-color-diagnosis}
\textbf{Setup.}
We extend the conditional analysis to all models and three color cutoffs
on the same 621 cases (Table~\ref{tab:app-color-conditional}). Counts for
different axes overlap; rates are undefined when no output meets a cutoff.
We resample numerators and denominators together over 10,000 source-group
draws from the 127 covered assets.

\textbf{Results and analysis.}
For GPT-Image-2 at the main color cutoff, the
95\% percentage intervals are \datasetareview{[7.2, 18.4]} for low preservation and
\datasetareview{[11.7, 24.6]} for low localization.

\begin{table}[t]
\centering
\caption{Conditional diagnosis on the same 621 color cases per model. For each color-attainment cutoff $c$, $n$ is the number of outputs with color score $\ge c$; subsequent columns count $F/P/L/M<50$. These are descriptive slices, not new pass criteria.}
\label{tab:app-color-conditional}
\scriptsize
\setlength{\tabcolsep}{4pt}
\begin{tabular}{lrrrrrrrrrrrrrrr}
\noalign{\hrule height 0.8pt}
& \multicolumn{5}{c}{Color $\ge32$} & \multicolumn{5}{c}{Color $\ge50$} & \multicolumn{5}{c}{Color $\ge80$} \\
\cline{2-6}\cline{7-11}\cline{12-16}
Model & $n$ & $F$ & \draft{$P$} & \draft{$L$} & \draft{$M$} & \draft{$n$} & \draft{$F$} & \draft{$P$} & \draft{$L$} & \draft{$M$} & \draft{$n$} & \draft{$F$} & \draft{$P$} & \draft{$L$} & \draft{$M$} \\
\noalign{\vskip 2pt\hrule height 0.4pt\vskip 2pt}
GPT-Image-2 & \datasetareview{449} & \datasetareview{134} & \draft{60} & \datasetareview{95} & \draft{5} & \datasetareview{401} & \datasetareview{97} & \revisionblue{50} & \revisionblue{71} & \draft{5} & \datasetareview{270} & \datasetareview{43} & \draft{27} & \draft{44} & \draft{0} \\
\draft{FireRed-Edit} & \datasetareview{20} & \datasetareview{14} & \draft{5} & \revisionblue{9} & \draft{0} & \draft{13} & \draft{9} & \draft{4} & \draft{5} & \draft{0} & \draft{7} & \draft{4} & \draft{1} & \draft{2} & \draft{0} \\
\draft{Qwen-Image-Edit-2511} & \datasetareview{38} & \datasetareview{27} & \draft{7} & \draft{15} & \draft{2} & \datasetareview{30} & \datasetareview{21} & \draft{6} & \draft{11} & \draft{2} & \draft{18} & \draft{12} & \draft{3} & \draft{5} & \draft{0} \\
FLUX.2-dev & \datasetareview{52} & \datasetareview{32} & \draft{14} & \revisionblue{21} & \draft{2} & \datasetareview{37} & \datasetareview{22} & \draft{12} & \draft{16} & \draft{1} & \datasetareview{15} & \draft{8} & \draft{2} & \draft{6} & \draft{0} \\
\draft{Qwen-Image-Edit} & \draft{16} & \draft{14} & \draft{4} & \draft{6} & \draft{0} & \draft{8} & \draft{7} & \draft{2} & \draft{3} & \draft{0} & \draft{4} & \draft{4} & \draft{0} & \draft{2} & \draft{0} \\
BAGEL & 5 & 4 & \draft{1} & \draft{1} & \draft{0} & \draft{4} & \draft{3} & \draft{1} & \draft{1} & \draft{0} & \draft{3} & \draft{3} & \draft{1} & \draft{1} & \draft{0} \\
FLUX.2-klein & \datasetareview{21} & \datasetareview{11} & \draft{2} & \draft{7} & \draft{0} & \datasetareview{17} & \datasetareview{8} & \draft{2} & \draft{7} & \draft{0} & \datasetareview{9} & \draft{5} & \draft{2} & \draft{5} & \draft{0} \\
\draft{Step1X-Edit} & 4 & 2 & 1 & 1 & 1 & 4 & 2 & 1 & 1 & 1 & 4 & 2 & 1 & 1 & 1 \\
\draft{FLUX.1-Kontext} & \draft{0} & \draft{0} & \draft{0} & \draft{0} & \draft{0} & \draft{0} & \draft{0} & \draft{0} & \draft{0} & \draft{0} & \draft{0} & \draft{0} & \draft{0} & \draft{0} & \draft{0} \\
OmniGen2 & 1 & 0 & \draft{0} & \draft{0} & \draft{0} & \draft{1} & \draft{0} & \draft{0} & \draft{0} & \draft{0} & \draft{1} & \draft{0} & \draft{0} & \draft{0} & \draft{0} \\
InstructPix2Pix & 4 & 4 & \draft{3} & \draft{4} & \draft{1} & \draft{1} & \draft{1} & \draft{1} & \draft{1} & \draft{0} & \draft{0} & \draft{0} & \draft{0} & \draft{0} & \draft{0} \\
\noalign{\hrule height 0.8pt}
\end{tabular}
\end{table}

Table~\ref{tab:app-color-execution} shows that open-weight models alter the
requested regions in most outputs but rarely reach the requested color.

\begin{table}[!ht]
\centering
\caption{Color attainment and edit execution on the same 621 color-editing cases. Columns give the case-mean color attainment (Table~\ref{tab:exp-color} instead averages asset--operation cells), the percentage of outputs with zero attainment and with attainment of at least 50, the case-mean localization and magnitude, and the percentage of outputs with both $L>0$ and $M>0$.}
\label{tab:app-color-execution}
\vspace{3pt}
\small
\setlength{\tabcolsep}{4pt}
\begin{tabular}{lrrrrrr}
\hline
Model & Attainment & Zero (\%) & $\ge50$ (\%) & $L$ & $M$ & $L,M>0$ (\%) \\
\hline
GPT-Image-2 & \datasetareview{61.47} & 8.2 & \datasetareview{64.6} & \datasetareview{69.94} & \datasetareview{93.46} & 99.5 \\
FireRed-Edit & \datasetareview{3.04} & \datasetareview{74.7} & 2.1 & \datasetareview{58.96} & \datasetareview{73.81} & 95.2 \\
Qwen-Image-Edit-2511 & \datasetareview{5.48} & \datasetareview{68.8} & \datasetareview{4.8} & \datasetareview{58.50} & \datasetareview{75.55} & \datasetareview{96.1} \\
Qwen-Image-Edit & \datasetareview{2.32} & \datasetareview{74.7} & 1.3 & \datasetareview{51.54} & \datasetareview{65.67} & \datasetareview{93.9} \\
FLUX.2-dev & \datasetareview{6.44} & \datasetareview{63.4} & \datasetareview{6.0} & \datasetareview{60.13} & \datasetareview{79.47} & 97.7 \\
BAGEL & \datasetareview{0.93} & \datasetareview{82.8} & 0.6 & \datasetareview{52.19} & \datasetareview{67.30} & 98.2 \\
Step1X-Edit & \datasetareview{0.91} & \datasetareview{84.1} & 0.6 & \datasetareview{45.68} & \datasetareview{54.63} & \datasetareview{88.4} \\
FLUX.2-klein & \datasetareview{3.31} & \datasetareview{71.3} & \datasetareview{2.7} & \datasetareview{54.88} & \datasetareview{75.29} & 96.8 \\
FLUX.1-Kontext & \datasetareview{0.22} & \datasetareview{84.5} & 0.0 & \datasetareview{29.48} & \datasetareview{41.48} & \datasetareview{72.6} \\
OmniGen2 & \datasetareview{0.50} & \datasetareview{82.6} & 0.2 & \datasetareview{40.74} & \datasetareview{55.80} & \datasetareview{94.8} \\
InstructPix2Pix & \datasetareview{0.60} & \datasetareview{81.0} & 0.2 & \datasetareview{48.41} & \datasetareview{65.40} & 97.7 \\
\hline
\end{tabular}
\end{table}

\subsection{Official PaintBench Measurements on Identical Outputs}
\label{sec:app-paintbench-matched}
\textbf{Setup.}
The fixed diagnostic panel contains 12 SVG, 12 chart, 41 web, and 36 slide
cases \revisionblue{from 40 sources}. The panel was assembled to cover known scoring boundaries, namely edits
that resemble global image perturbations and edits near the
change-detection threshold, together with 24 stratified random controls.
It is not a random sample of the benchmark, so pair frequencies do not
estimate prevalence.

We score its 1,111 model outputs using PaintBench's
\texttt{compute\_problem\_stats} at commit \texttt{a648b9f}.
For CIE76 tolerance $t\in\{0,\ldots,10\}$, correct edit pixels $CE_t$ and
incorrect preservation pixels $IP_t$ give
\[
  \operatorname{IoU}_t=\frac{CE_t}{|E|+IP_t},\qquad
  E=\{x:I(x)\ne T(x)\},\qquad
  \operatorname{mIoU}=\frac{1}{11}\sum_{t=0}^{10}\operatorname{IoU}_t.
\]
We retain PaintBench's own edit region, both accuracy curves, and
changed-pixel counts. \draft{Its preservation accuracy on this panel is a
separate measure from our $P$, which uses the protected region and tolerance
set defined in Section~\ref{sec:metric-preservation}.} Both scorers use our image alignment, without
PaintBench's separate resize/center-crop preprocessing. Reference targets
attain mIoU 100; unchanged inputs have zero edit accuracy at tolerance zero
and preservation accuracy 100.

We compare 27 matching windows based on mIoU and mean edit/preservation
accuracy (Table~\ref{tab:app-paintbench-windows}). \revisionblue{For the 101-case panel's 5,555 same-case model pairs, the strict comparison requires} an mIoU difference of at most one point and
edit- and preservation-accuracy differences of at most two points at every
tolerance, with all scores on a 0--100 scale.

\textbf{Results and analysis.}
Table~\ref{tab:app-paintbench-means} gives per-model means.
The main figure illustrates a pair satisfying this strict rule.
\datasetareview{The differing matched pairs span 13 cases.} \exampleswap{In the illustrated webpage,
FLUX.2-dev and Step1X-Edit both miss the requested target appearance.}
\datasetareview{Their spatial coverage and change magnitude nevertheless differ
(Table~\ref{tab:app-paintbench-means} gives panel-wide means).}

\begin{figure}[!htbp]
  \centering
  \includegraphics[width=\linewidth]{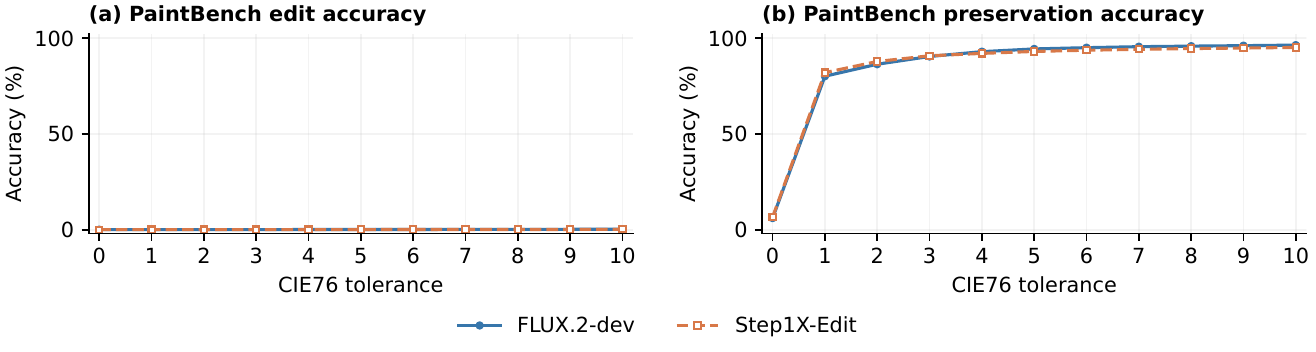}\par\nointerlineskip
  \caption{\datasetareview{Complete PaintBench accuracy curves for the \exampleswap{FLUX.2-dev/Step1X-Edit} pair
  in Figure~\ref{fig:exp-paintbench-matched}. Each curve retains all eleven
  CIE76 tolerances. The maximum between-model accuracy gaps are \exampleswap{0.07} points
  for editing and \exampleswap{1.87} points for preservation, both within the two-point
  matching bound at every tolerance.}}
  \label{fig:app-paintbench-curves}
\end{figure}

Figure~\ref{fig:app-paintbench-curves} retains both complete accuracy curves
for the pair shown in Figure~\ref{fig:exp-paintbench-matched}.

\begin{table}[t]
\centering
\caption{Same aligned outputs, official PaintBench measurements and our four-axis measurements. Means over 101 fixed diagnostic cases per model (0--100), covering 11 models. These are measurements of this panel, not scores on PaintBench\textquotesingle s original benchmark. Best values are bold within each column.}
\label{tab:app-paintbench-means}
\scriptsize
\setlength{\tabcolsep}{4pt}
\begin{tabular}{lrrrrrrr}
\noalign{\hrule height 0.8pt}
Model & mIoU & Edit acc. & Pres. acc. & $F$ & $P$ & $L$ & $M$ \\
\noalign{\vskip 2pt\hrule height 0.4pt\vskip 2pt}
GPT-Image-2 & \datasetareview{\textbf{8.20}} & \datasetareview{\textbf{26.67}} & \datasetareview{77.96} & \datasetareview{\textbf{36.89}} & \datasetareview{90.36} & \datasetareview{\textbf{50.99}} & \datasetareview{\textbf{84.84}} \\
\draft{FireRed-Edit} & \datasetareview{2.80} & \datasetareview{16.11} & \datasetareview{82.19} & \datasetareview{15.93} & \datasetareview{88.66} & \datasetareview{50.97} & \datasetareview{72.32} \\
\draft{Qwen-Image-Edit-2511} & \datasetareview{2.48} & \datasetareview{17.11} & \datasetareview{81.83} & \datasetareview{15.36} & \datasetareview{89.01} & \datasetareview{47.74} & \datasetareview{70.61} \\
\draft{Qwen-Image-Edit} & \datasetareview{2.18} & \datasetareview{15.38} & \datasetareview{78.57} & \datasetareview{13.88} & \datasetareview{87.18} & \datasetareview{45.39} & \datasetareview{64.98} \\
FLUX.2-dev & \datasetareview{1.36} & \datasetareview{9.11} & \datasetareview{76.45} & \datasetareview{9.57} & \datasetareview{86.10} & \datasetareview{45.30} & \datasetareview{69.70} \\
FLUX.2-klein & \datasetareview{1.18} & \datasetareview{7.83} & \datasetareview{70.68} & \datasetareview{5.43} & \datasetareview{81.77} & \datasetareview{40.76} & \draft{67.89} \\
BAGEL & \datasetareview{1.39} & \datasetareview{10.26} & \datasetareview{62.33} & 7.93 & \datasetareview{72.01} & \datasetareview{36.24} & \datasetareview{58.88} \\
\draft{Step1X-Edit} & \datasetareview{1.19} & \datasetareview{10.26} & \datasetareview{78.03} & \datasetareview{6.90} & \datasetareview{86.87} & \datasetareview{37.79} & \datasetareview{63.58} \\
OmniGen2 & \datasetareview{0.40} & \datasetareview{8.61} & \datasetareview{46.65} & \datasetareview{6.85} & \datasetareview{54.83} & \datasetareview{30.44} & \datasetareview{62.04} \\
\draft{FLUX.1-Kontext} & \datasetareview{0.78} & \datasetareview{7.06} & \datasetareview{\textbf{85.84}} & \datasetareview{2.57} & \datasetareview{\textbf{93.22}} & \datasetareview{22.69} & \datasetareview{39.94} \\
InstructPix2Pix & \datasetareview{0.19} & \datasetareview{3.41} & \datasetareview{19.71} & \datasetareview{2.02} & \datasetareview{27.74} & \datasetareview{35.11} & \datasetareview{64.13} \\
\noalign{\hrule height 0.8pt}
\end{tabular}
\end{table}

\textbf{Comparison across all source assets.}

\textbf{Setup.}
To examine these distinctions across sources, we use a stratified random panel: one case
per source--operation group (seed 20260917), giving 607 cases from all
153 sources and 6,677 existing outputs. Both panels use the same scoring
rules and similarity windows.

\textbf{Results and analysis.}
Among 33,385 same-case model pairs, \datasetareview{416} satisfy the strict
matching rule; \datasetareview{223} of these differ by at least 20 points in $L$ or $M$,
spanning \datasetareview{68 cases and 56 sources}. Table~\ref{tab:app-pb-607} reports the
complete difference distributions, including unmatched pairs.
This comparison shows that the diagnostic distinction extends beyond the
boundary-focused panel. Frequencies describe this source--operation
sample, not the prevalence among all 1,740 cases; models sharing a source
are not independent observations.

\begin{table}[!ht]
\centering
\caption{\revisionblue{Complete difference distributions on the fixed 607-case panel: 33,385 same-case pairs, including \datasetareview{416} matched pairs. Maximum curve gap considers both accuracy curves at all eleven tolerances. All differences are on a 0--100 scale.}}
\label{tab:app-pb-607}
\vspace{3pt}
{\small\setlength{\tabcolsep}{4pt}
\begin{tabular}{llrrrrr}
\hline
Pairs & Difference & Min. & Q1 & Median & Q3 & Max. \\
\hline
All pairs & $|\Delta\mathrm{mIoU}|$ & 0.00 & \datasetareview{0.08} & \datasetareview{1.00} & \datasetareview{6.30} & 96.00 \\
All pairs & Maximum curve gap & 0.00 & \datasetareview{38.70} & \datasetareview{69.29} & \datasetareview{92.27} & 100.00 \\
All pairs & $\max(|\Delta L|,|\Delta M|)$ & 0.00 & \datasetareview{12.25} & \datasetareview{27.49} & \datasetareview{47.32} & 100.00 \\
Matched pairs & $|\Delta\mathrm{mIoU}|$ & 0.00 & 0.00 & 0.00 & 0.02 & \datasetareview{0.85} \\
Matched pairs & Maximum curve gap & 0.00 & 0.02 & \datasetareview{0.40} & \datasetareview{1.10} & \datasetareview{1.99} \\
Matched pairs & $\max(|\Delta L|,|\Delta M|)$ & 0.00 & \datasetareview{7.06} & \datasetareview{23.79} & \datasetareview{44.91} & 100.00 \\
\hline
\end{tabular}}
\end{table}

\begin{table}[t]
\centering
\caption{\draft{Sensitivity to the numerical similarity window. $d_I$ bounds the mIoU gap; $d_A$ bounds each mean edit/preservation-accuracy gap; $d_{LM}$ is the minimum localization or magnitude gap. Counts concern 5,555 same-case model pairs on the 101-case panel. The stricter rule used in the main text bounds each accuracy gap at every tolerance by two points, with $d_I=1$: \datasetareview{78 pairs qualify, and 28 differ by at least 20 points in localization or magnitude, spanning 13 cases.}}}
\label{tab:app-paintbench-windows}
\scriptsize
\setlength{\tabcolsep}{6pt}
\begin{tabular}{rrrrrr}
\noalign{\hrule height 0.8pt}
$d_I$ & $d_A$ & $d_{LM}$ & Similar pairs & Differing pairs & Cases \\
\noalign{\vskip 2pt\hrule height 0.4pt\vskip 2pt}
0.5 & 1 & 10 & \datasetareview{213} & \datasetareview{123} & \datasetareview{48} \\
0.5 & 1 & 20 & \datasetareview{213} & \datasetareview{92} & \datasetareview{38} \\
0.5 & 1 & 30 & \datasetareview{213} & \datasetareview{71} & \datasetareview{31} \\
0.5 & 2 & 10 & \datasetareview{379} & \datasetareview{219} & \datasetareview{66} \\
0.5 & 2 & 20 & \datasetareview{379} & \datasetareview{164} & \datasetareview{56} \\
0.5 & 2 & 30 & \datasetareview{379} & \datasetareview{127} & \datasetareview{46} \\
0.5 & 5 & 10 & \datasetareview{874} & \datasetareview{516} & 78 \\
0.5 & 5 & 20 & \datasetareview{874} & \datasetareview{395} & 71 \\
0.5 & 5 & 30 & \datasetareview{874} & \datasetareview{293} & \datasetareview{57} \\
1 & 1 & 10 & \datasetareview{221} & \datasetareview{124} & \datasetareview{49} \\
1 & 1 & 20 & \datasetareview{221} & \datasetareview{93} & \datasetareview{39} \\
1 & 1 & 30 & \datasetareview{221} & \datasetareview{72} & \datasetareview{32} \\
1 & 2 & 10 & \datasetareview{402} & \datasetareview{230} & \datasetareview{66} \\
1 & 2 & 20 & \datasetareview{402} & \datasetareview{170} & \datasetareview{57} \\
1 & 2 & 30 & \datasetareview{402} & \datasetareview{132} & \datasetareview{47} \\
1 & 5 & 10 & \datasetareview{945} & \datasetareview{556} & 79 \\
1 & 5 & 20 & \datasetareview{945} & \datasetareview{419} & 73 \\
1 & 5 & 30 & \datasetareview{945} & \datasetareview{310} & \datasetareview{60} \\
2 & 1 & 10 & \datasetareview{222} & \datasetareview{124} & \datasetareview{49} \\
2 & 1 & 20 & \datasetareview{222} & \datasetareview{93} & \datasetareview{39} \\
2 & 1 & 30 & \datasetareview{222} & \datasetareview{72} & \datasetareview{32} \\
2 & 2 & 10 & \datasetareview{408} & \datasetareview{232} & \datasetareview{66} \\
2 & 2 & 20 & \datasetareview{408} & \datasetareview{172} & \datasetareview{57} \\
2 & 2 & 30 & \datasetareview{408} & \datasetareview{134} & \datasetareview{47} \\
2 & 5 & 10 & \datasetareview{990} & \datasetareview{583} & 82 \\
2 & 5 & 20 & \datasetareview{990} & \datasetareview{440} & 75 \\
2 & 5 & 30 & \datasetareview{990} & \datasetareview{325} & \datasetareview{62} \\
\noalign{\hrule height 0.8pt}
\end{tabular}
\end{table}

\begin{figure}[!htbp]
  \centering
  \includegraphics[width=\linewidth]{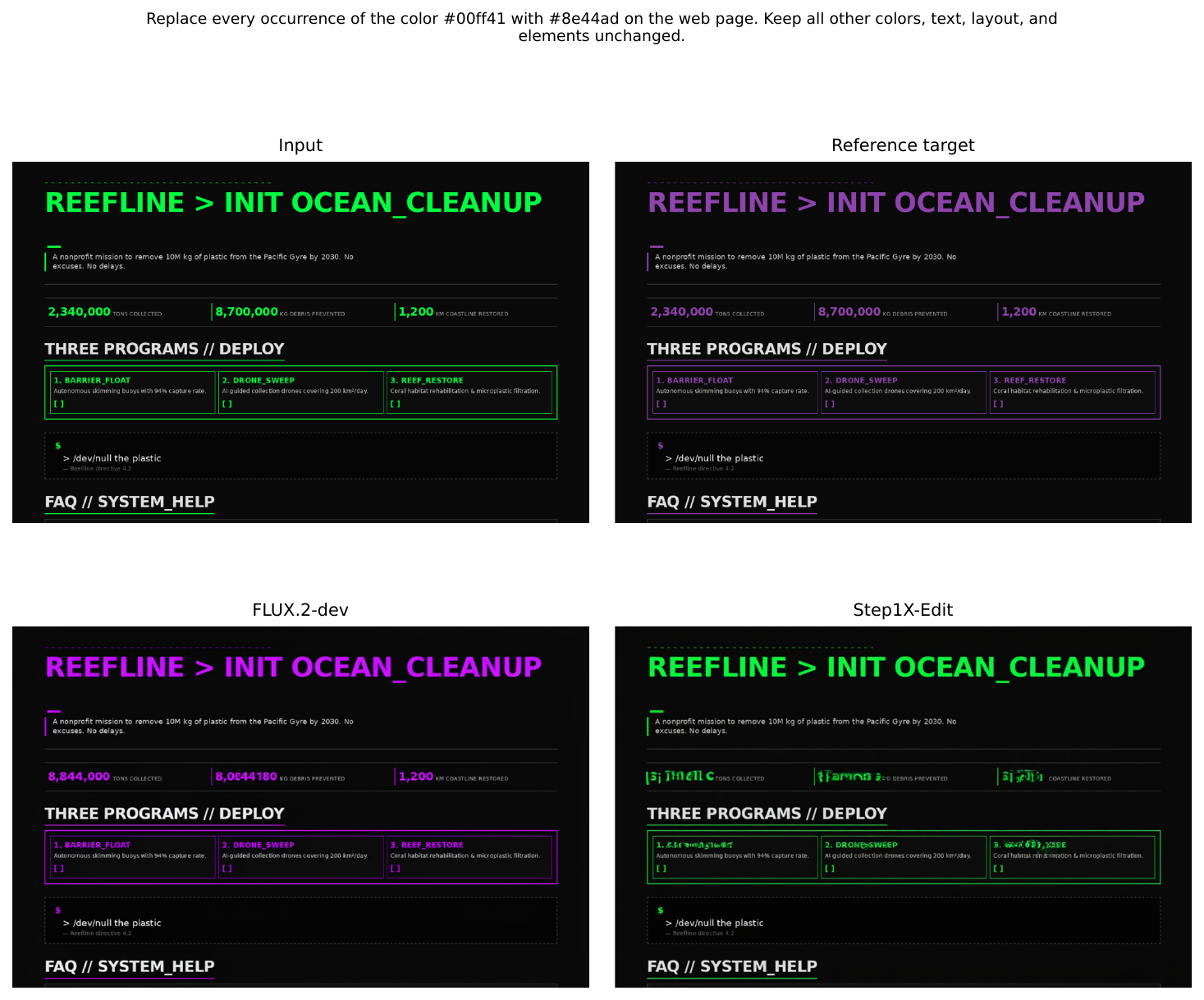}
  \caption{\datasetareview{Complete task and instruction, with the input, reference target, and \exampleswap{FLUX.2-dev/Step1X-Edit} outputs used in the comparison.}}
  \label{fig:app-paintbench-example}
\end{figure}

\subsection{Comparing Qwen-Image-Edit Versions}
\label{sec:app-qwen-versions}
\textbf{Setup.}
We compare Qwen-Image-Edit and Qwen-Image-Edit-2511 on identical source
images and instructions for all 1,740 cases. Both use \abbrreview{bfloat16 (BF16)}, true \abbrreview{classifier-free
guidance (CFG)} scale 4.0, and seed 42, with 50 steps for Qwen-Image-Edit and 40
for Qwen-Image-Edit-2511 (Appendix~\ref{sec:app-inference}).
Table~\ref{tab:app-qwen-versions} reports per-operation means and paired
intervals using the source-group procedure in Section~\ref{sec:exp-setup}.

\begin{table}[!htbp]
\centering
\caption{Qwen-Image-Edit version comparison by operation. The two $H$ columns use all cases in each cell. All 1,740 cases use identical actual source images and instructions; $\Delta$ is Qwen-Image-Edit-2511 minus Qwen-Image-Edit, with pointwise 95\% source-group BCa intervals. Bold marks the higher full-set mean, not a significance test. The final row weights cells equally.}
\label{tab:app-qwen-versions}
\begingroup\footnotesize
\setlength{\tabcolsep}{4pt}\renewcommand{\arraystretch}{1.08}
\begin{tabular}{llrrrl}
\hline
Asset & Operation & Original $H$ & 2511 $H$ & $n$ & $\Delta$ [95\% CI] \\
\hline
Chart & Color & \datasetareview{3.22} & \datasetareview{\textbf{6.48}} & 54 & \datasetareview{+3.26 [-0.34, +7.60]} \\
 & Font size & \datasetareview{7.11} & \datasetareview{\textbf{20.32}} & 42 & \datasetareview{+13.21 [+2.18, +21.57]} \\
 & Grid & \datasetareview{\textbf{9.75}} & \datasetareview{9.27} & 48 & \datasetareview{-0.49 [-1.44, -0.13]} \\
 & Line width & \datasetareview{9.68} & \datasetareview{\textbf{17.50}} & 96 & \datasetareview{+7.81 [+3.85, +13.98]} \\
 & Palette & \datasetareview{4.95} & \datasetareview{\textbf{8.62}} & 76 & \datasetareview{+3.66 [-2.47, +8.56]} \\
\hline
Slide & Border & \draft{2.94} & \draft{\textbf{4.82}} & 78 & \draft{+1.87 [+0.08, +5.15]} \\
 & Color & \draft{3.09} & \draft{\textbf{14.60}} & 136 & \draft{+11.51 [+6.64, +17.39]} \\
 & Hide & 61.26 & \textbf{68.35} & 42 & \draft{+7.10 [+1.37, +14.96]} \\
 & Font size & \draft{\textbf{11.66}} & \draft{11.50} & 131 & \draft{-0.16 [-3.60, +3.78]} \\
 & Opacity & 1.48 & \draft{\textbf{7.08}} & 122 & \draft{+5.59 [+2.76, +10.11]} \\
\hline
SVG & Color & \draft{11.99} & \textbf{15.98} & 82 & \draft{+3.99 [-1.19, +9.16]} \\
 & Hide & \draft{\textbf{29.73}} & \draft{23.00} & 37 & \draft{-6.72 [-18.33, +4.77]} \\
 & Move & \draft{\textbf{33.35}} & \draft{25.78} & 51 & \draft{-7.57 [-22.00, +5.71]} \\
 & Opacity & \revisionblue{\textbf{9.32}} & \revisionblue{7.98} & 56 & \revisionblue{-1.35 [-8.29, +5.24]} \\
 & Palette & \draft{3.42} & \draft{\textbf{12.11}} & 90 & \draft{+8.68 [+4.02, +15.36]} \\
 & Dashes & \draft{10.75} & \draft{\textbf{11.37}} & 75 & \draft{+0.62 [-8.32, +6.50]} \\
 & Stroke width & \draft{2.52} & \draft{\textbf{6.76}} & 64 & \draft{+4.24 [+0.29, +9.26]} \\
\hline
Web & Corners & \draft{1.54} & \draft{\textbf{2.29}} & 93 & \draft{+0.74 [-0.79, +4.01]} \\
 & Color & \draft{6.90} & \draft{\textbf{13.58}} & 183 & \draft{+6.68 [+1.57, +13.32]} \\
 & Move & \draft{15.90} & \draft{\textbf{26.57}} & 118 & \draft{+10.66 [+4.19, +18.18]} \\
 & Hide & \draft{65.72} & \textbf{79.17} & 39 & \draft{+13.45 [+8.94, +19.26]} \\
 & Text & \draft{51.47} & \draft{\textbf{57.32}} & 27 & \draft{+5.85 [+1.83, +13.40]} \\
\hline
\multicolumn{2}{l}{Cell-mean $H$} & \datasetareview{16.26} & \datasetareview{\textbf{20.47}} & 1740 & \datasetareview{+4.21 [+2.95, +5.46]} \\
\hline
\end{tabular}\endgroup
\end{table}

\textbf{Results and analysis.}
Qwen-Image-Edit-2511 improves aggregate performance, with the largest gains
\datasetareview{in web visibility, chart font size, slide recoloring, web translation,
SVG palette replacement, and chart line width.} Slide typography
shows no clear improvement; the gain in web corner-radius control is small,
and both versions still score poorly. Negative point estimates for SVG
visibility, translation, and opacity have intervals containing zero.
\datasetareview{Chart grid editing shows a small decline with an interval below zero.}
The comparison reflects the configured editors and their step
counts.
\emph{A version-level improvement is informative when accompanied by an
account of which capabilities improve.}

\subsection[Uncertainty with Shared Source Assets]{\draft{Uncertainty with Shared Source Assets}}
\label{sec:app-source-statistics}
\textbf{Setup.}
\draft{For each of 10,000 bootstrap draws, we sample sources with replacement
within each asset type and include all operations belonging to each sampled
source. Models share the same draw. We then recompute the 22 cell means and
weight cells equally, as in the primary score. No cell was empty in these
draws. Bias-corrected and accelerated intervals use leave-one-source-out
estimates with the same grouping. This preserves dependencies between
operations that reuse a source, while conditioning on the tested outputs
and inference settings. The same source grouping is used for the paired
Qwen-Image-Edit comparison in Table~\ref{tab:app-qwen-versions}.
Same-operation comparisons retain the asset-specific source resampling
described in Appendix~\ref{sec:app-shared-operation}.}

\textbf{Results and analysis.}
\draft{Table~\ref{tab:exp-main} reports the marginal intervals and
Table~\ref{tab:app-paired-gaps} the paired gaps between adjacent models.
The paired intervals separate the six leading models from one another and
OmniGen2 from InstructPix2Pix, whereas the four adjacent pairs between BAGEL
and OmniGen2 remain statistically unresolved; close point estimates should
not be read as a reliable ordering.}

\begin{table}[!ht]
\centering
\caption{Paired gaps in cell-mean $H$ between adjacent models of Table~\ref{tab:exp-main} under the default scoring configuration. $\Delta H$ is the first model minus the second; intervals are 95\% BCa intervals from 10,000 source-group bootstrap draws shared across models. An interval excluding zero separates the pair.}
\label{tab:app-paired-gaps}
\vspace{3pt}
\small
\setlength{\tabcolsep}{5pt}
\begin{tabular}{lrr}
\hline
\datasetareview{Model pair} & \datasetareview{$\Delta H$} & \datasetareview{95\% interval} \\
\hline
\datasetareview{GPT-Image-2 / FireRed-Edit} & \datasetareview{$+25.28$} & \datasetareview{$[23.29,\,27.38]$} \\
\datasetareview{FireRed-Edit / Qwen-Image-Edit-2511} & \datasetareview{$+1.38$} & \datasetareview{$[0.02,\,2.61]$} \\
\datasetareview{Qwen-Image-Edit-2511 / Qwen-Image-Edit} & \datasetareview{$+4.21$} & \datasetareview{$[2.95,\,5.46]$} \\
\datasetareview{Qwen-Image-Edit / FLUX.2-dev} & \datasetareview{$+3.15$} & \datasetareview{$[1.72,\,4.65]$} \\
\datasetareview{FLUX.2-dev / BAGEL} & \datasetareview{$+2.30$} & \datasetareview{$[0.53,\,3.91]$} \\
\datasetareview{BAGEL / Step1X-Edit} & \datasetareview{$+1.03$} & \datasetareview{$[-0.58,\,2.70]$} \\
\datasetareview{Step1X-Edit / FLUX.1-Kontext} & \datasetareview{$+1.00$} & \datasetareview{$[-0.40,\,2.36]$} \\
\datasetareview{FLUX.1-Kontext / FLUX.2-klein} & \datasetareview{$+0.35$} & \datasetareview{$[-1.02,\,1.88]$} \\
\datasetareview{FLUX.2-klein / OmniGen2} & \datasetareview{$+1.11$} & \datasetareview{$[-0.53,\,2.71]$} \\
\datasetareview{OmniGen2 / InstructPix2Pix} & \datasetareview{$+3.49$} & \datasetareview{$[2.00,\,4.93]$} \\
\hline
\end{tabular}
\end{table}

\section{Dataset Construction and Scoring Details}
\label{sec:app-methods}
This appendix provides the asset sources and implementation choices
supporting the measurements in Section~\ref{sec:benchmark}.

\subsection{Source Asset Construction}
\label{sec:app-asset-sources}

For SVG assets, we adapt the template generation stage of Chat2SVG~\citep{DBLP:conf/cvpr/WuS025},
and generation components of IntroSVG~\citep{DBLP:journals/corr/abs-2603-09312},
with simplified variants and custom outline templates.
Chart generation uses reference code from ChartMimic~\citep{DBLP:conf/iclr/0002SLS0JXZLZLN25},
alongside LLM-based and programmatic synthesis inspired by ChartLlama~\citep{DBLP:journals/corr/abs-2311-16483}
and FigureQA~\citep{DBLP:conf/iclr/KahouMAKTB18}, respectively.
For web assets, we adapt
OpenUI\footnote{\url{https://github.com/wandb/openui}} system prompts,
draw on WebGen-Bench~\citep{DBLP:conf/nips/LuYRHXWSZZL25}
for instruction design, and use a custom two-stage generation workflow
with layout categories informed by
Open Design\footnote{\url{https://github.com/nexu-io/open-design}}.
For slides, we adapt AutoPresent~\citep{DBLP:conf/cvpr/GeWZPSTSSFND25}
code generation prompts and convert the resulting slides to HTML.
\revisionblue{The source counts refer to constructed assets, not deployed
websites or collected presentation files. The SVG pool has 12 sources from
each of four construction variants (Chat2SVG, a simplified variant, IntroSVG,
and custom outlines). Charts comprise 10 ChartMimic-inspired, 10
ChartLlama-inspired, and five FigureQA-inspired sources. Web assets comprise
25 custom two-stage, 13 OpenUI-prompt, and 13 WebGen-inspired sources.}

\textbf{Task verification.}


We require paired renders to have matching canvas dimensions, a visible
requested change, preserved surrounding layout, and agreement between
the instruction and the edited objects and parameters. Repeated rendered
answers for the same source and operation are deduplicated. 
AI-assisted screening uses Kimi K3~\citep{kimi2026k3}, GLM-5.3-Flash~\citep{zai2026glm53flash}, 
and GPT-6~\citep{openai2026gpt6astra}.
All retained cases have an AI inspection record.An author with a computer vision research background
reviews all retained tasks and resolves reported ambiguities against the
editable sources.
Object references are stated in
visible terms, and rendered targets preserve
background cutouts and other protected content.

Inputs, targets, masks, and edit metadata are retained together.

\begin{figure}[!htbp]
\centering
\includegraphics[width=\linewidth]{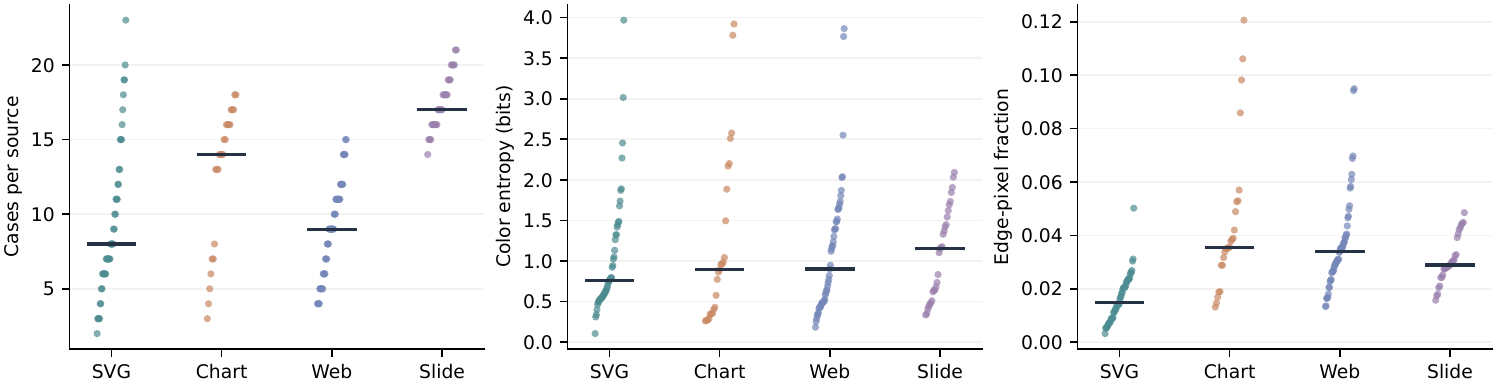}
\caption{\datasetareview{Source-level descriptors. Each point is one source; horizontal
bars mark medians. Multiple edits of a source contribute only one point.
Color entropy describes palette dispersion, and edge density describes
visible detail under fixed preprocessing.}}
\label{fig:app-source-diversity}
\end{figure}

\textbf{\draft{Source-level variation.}}
\draft{We describe the 153 sources before counting their edit variants
(Figure~\ref{fig:app-source-diversity}). For these descriptors only, source images are resized to at most 512
pixels on the longest side; color entropy uses eight bins per RGB channel,
and edge density uses fixed Canny thresholds of 80 and 160. These pixel
descriptors capture variation in palette and visual structure; sources
differ within each asset type.}

\submissionreview{Model-specific image preparation is specified in
Appendix~\ref{sec:app-inference}.}

\subsubsection[Human Verification of Task Validity]{{Human Verification of Task Validity}}
\label{sec:app-human-verification}

\paragraph{\humanauditreview{Setup.}}
\humanauditreview{One author with a computer vision research background
inspected the input image, instruction, rendered target, and edit mask
for all 1,740 tasks retained through screening with Kimi K3, GLM-5.3-Flash,
and GPT-6, covering 153 source assets and all 22 asset--operation combinations.
The criteria were identifiable target objects, agreement between the
requested change and target, and preservation of unrelated content.
\humanauditfix{The human confirmation rate of the screening outcome was measured as the fraction of
retained tasks accepted without concerns during human review.}}

\paragraph{\humanauditreview{Results and analysis.}}
\humanauditreview{\datasetareview{During curation, human review accepted 1,721 of 1,740 tasks without concerns}
(98.91\% were accepted without concerns on initial human inspection) and raised questions about
19 tasks. Source and render inspection identified 11 cases requiring clearer
object references and two requiring target correction; the other six were
consistent with their instructions. The clarifications and corrections
were checked against the rendered outcomes. \humanauditfix{These results
show how human review resolves residual object-reference and rendering
issues that survive multi-model screening.} All confirmed
issues were resolved, and the corrected tasks passed rechecking. All benchmark
scores reported in this paper use the final, corrected evaluation references.}

\subsection{Image Preparation and Scoring Details}
\label{sec:app-scoring-details}

\textbf{Image preparation.}
After the inference wrapper restores the reference canvas
(Appendix~\ref{sec:app-inference}), the scorer compares RGB images at the
target dimensions. Its fallback for mismatched dimensions uses bicubic interpolation; when the
absolute difference in aspect ratio exceeds 0.02, centered black padding
precedes resizing. No content-based registration is applied. The allowed
region in Equation~\ref{eq:edit-region} follows rendered change, including
vacated and newly occupied pixels, while unchanged pixels inside an object's
bounding box remain protected.

\textbf{Fidelity and preservation.}
For fidelity, each image is smoothed by a $3\!\times\!3$ Gaussian filter
with $\sigma=0.6$, and the mask is dilated by a $3\!\times\!3$ rectangular
kernel, giving $\widetilde I,\widetilde T,\widetilde O$ and $E^+$.
Both the prediction--target agreement and unchanged-input baseline use
the same channel and tolerances $\{5,10,20\}$. A baseline $b>0.995$ is
flagged as uninformative and receives zero fidelity; an empty mask also
receives zero fidelity. Preservation instead uses unblurred images, the
original mask, and tolerances $\{2,5,10,20\}$. When no pixels lie outside
$E$, preservation is 100 because there is no protected region to evaluate.

\textbf{Spatial change maps.}
We use a localization threshold of 6 if at least
$\max(16,0.05|E|)$ input--target pixels exceed that distance, and 2 otherwise.
Thresholding input--target and input--output differences gives $G$ and the
candidate output change map; an empty $G$ falls back to $E$. All candidate
pixels inside $E$ are retained. Outside $E$, only eight-connected components
with at least $\max(16,\lfloor0.0005N\rfloor)$ pixels survive, where $N$ is
the canvas area. This produces $D$ without erasing coherent thin bands.
Empty detected changes or zero recall/precision give $L=0$.
The precision denominator caps background change at $4|E|$ to keep
widespread redraw from overwhelming spatial diagnosis; preservation still
measures agreement across the full protected region.
All four axes are pixel-area weighted, so an object contributes in
proportion to its rendered area; background components below the
threshold above are treated as rendering noise by $L$.

\textbf{Effective-change core and amplitude.}
The core $C$ first retains pixels in $E$ whose input--target distance
exceeds 6. If fewer than $\max(20,0.2|E|)$ pixels remain, the threshold
falls to 2; if the core is still too small, it becomes $E$.
Using Lab pixel vectors, the amplitude ratio and agreement are
\begin{equation}
 r_M=\frac{\sum_{x\in C}\|O_x-I_x\|_1}
              {\sum_{x\in C}\|T_x-I_x\|_1},\qquad
 a(r_M)=\max\{0,1-\max(0,|r_M-1|-0.15)\}.
 \label{eq:change-amplitude}
\end{equation}
If the target amplitude is below $10^{-6}$, the ratio is 1 when the
output amplitude is also below that threshold, and 0 otherwise.
\draft{The direction term compares the complete Lab change fields on $C$:}
\begin{equation}
 \draft{c=\frac{\sum_{x\in C}(O_x-I_x)^\top(T_x-I_x)}
 {\sqrt{\sum_{x\in C}\|O_x-I_x\|_2^2}\sqrt{\sum_{x\in C}\|T_x-I_x\|_2^2}}.}
 \label{eq:change-direction}
\end{equation}
\draft{Keeping pixel locations avoids cancellation between opposing changes,
for example the vacated and newly occupied regions of a translated object.
A change field with norm below $10^{-6}$ is treated as zero: two zero
fields receive $c=1$, while only one zero field gives $c=0$.}
The amplitude term gives full credit within 15\% of the requested visual
change and decreases linearly beyond that band.

\textbf{Eligibility for attributable editing.}
Fidelity, localization, and magnitude share the same deterministic check
$g$. It examines change strength relative to the target, cancellation of
changes within local blocks, similarity to a blurred input, and whether a
background-fitted global color transform also explains the apparent edit.
\draft{Blur and global-transform checks require both agreement with the fitted
transformation and residual change below 5\% of the target amplitude.
Target comparisons protect requested transform-like edits; a fitted
transformation alone cannot erase a measurable local edit.} A failed check sets these
three axes to zero; preservation is measured separately. Passing does not
establish that the requested appearance is correct, which remains the role
of fidelity. The checks and thresholds are fixed across models; their exact
implementation and configuration are included in the supplementary code. 

\draft{\textbf{Exact rejection rules.}
Let $m_T$ be the median input--target CIE76 distance on $C$, and define the
residual ratio for an explanatory image $X$ as
$\rho_X=\sum_{x\in C}\|O_x-X_x\|_1/
\max(\sum_{x\in C}\|T_x-I_x\|_1,10^{-6})$, using Lab vectors.
The first satisfied rule sets $g=0$; otherwise $g=1$.}
\begin{enumerate}
\item \draft{\emph{Negligible change:} $E$ or $C$ is empty, or $r_M<0.05$.}
\item \draft{\emph{Blur match:} for a Gaussian-blurred input $B_\sigma$,
$\sigma\in\{1,2,3\}$, require $\rho_{B_\sigma}<0.05$ and a median
$B_\sigma$--$O$ distance on $C$ below both $\min(2,0.8m_T)$ and the
median $O$--$T$ distance. The median $B_\sigma$--$T$ distance must be at
least $\min(2,0.8m_T)$.}
\item \draft{\emph{Local cancellation:} when $r_M<0.9$, dilate $C$ by
eight pixels and retain Lab changes there. Within $16\!\times\!16$ blocks,
sum the signed changes in each channel. Divide the sum of absolute block
sums by the sum of absolute pixel changes, assigning zero when that
denominator is at most $10^{-9}$; reject if the maximum ratio across
channels is below 0.25.}
\item \draft{\emph{Global color transformation:} fit an affine Lab map
$A I+t$ outside $E$, using at most 65,536 pixels selected by deterministic
stride. Require at least 4,096 background pixels, $|E|\le0.8N$, and a
smallest singular value of centered background Lab samples, normalized by
$\sqrt{n-1}$, of at least 0.3. An eligible fit has
$\max_{ij}|(A-\mathrm{Id})_{ij}|\le1$, $\|t\|_\infty\le40$, and median
background residual below 6. Reject only if $\rho_{AI+t}<0.05$, its median
core residual to $O$ is below $\min(2,0.75m_T)$, and that to $T$ is at
least $\min(8,0.75m_T)$. If the singular-value condition fails, fit only
luminance when its background variance is at least 1: require
$0.8\le a\le1.25$, $|b|\le40$, and the same median residual thresholds
using luminance differences. For this fallback, $\rho$ still uses all three
Lab channels, with the input's chromatic channels unchanged. Insufficient
data or an ineligible fit leaves this rule inactive.}
\end{enumerate}
\revisionblue{The affine fit minimizes
$\sum_{x\in S}\|A I_x+t-O_x\|_2^2$ over sampled background pixels $S$
by ordinary least squares, without robust weighting; the luminance fallback
minimizes its one-dimensional counterpart. These are per-output explanatory
fits. Thresholds are fixed operational choices developed using reference
checks and constructed counterexamples. Sensitivity
and removal tests quantify the consequences of these choices
(Appendices~\ref{sec:app-ablation} and~\ref{sec:app-check-effects}).}

\draft{These rules detect specified incidental changes, rather than certifying
instruction correctness. On all 1,740 cases, the rendered target receives
$(100,100,100,100)$ and the unchanged input $(0,100,0,0)$ at the reported
precision. The largest fidelity baseline is 0.9709, below the 0.995 cutoff.}

\subsection{Separate Color-attainment Measurement}
\label{sec:app-color-measurement}
\draft{CIEDE2000 accounts for perceptual nonuniformity in Lab color space;
the color-agreement tests and change-region thresholds instead use
Euclidean Lab distance (CIE76).}
For the 621 covered color-editing cases, we compare predicted colors with
target references inside stable object cores. These cores follow the stored
object metadata: two erosions with a $3\!\times\!3$ kernel, or distance-transform
ridges for thin objects. A pixel matches its nearest target color within
CIEDE2000 tolerance $\min(5,0.4d_{\mathrm{src,tgt}})$, where the source--target
separation is recorded during construction; the tolerance is 5 when that
separation is unavailable.
\revisionblue{Target references use these same cores. We retain a single-color
reference if it covers the core sufficiently for full credit. Otherwise,
and for multicolor references, we augment the stored colors with the most
frequent uncovered target colors until every core pixel is within tolerance.
Target shares use the same nearest-color assignment as prediction shares;
unused color centers are removed. Reference construction never uses model
outputs.}

For a single-color object, matching fraction $q$ gives
$100\operatorname{clip}_{[0,1]}(q/0.55)$.
For a multicolor object, the score is
$100\sum_k\min(\widehat{s}_k,s_k)$, where $s_k$ is the target share and
$\widehat{s}_k$ the matched prediction share assigned to target color $k$.
An object instead receives zero when at least 90\% of its core matches the
background and at most 10\% of its full region differs from the background,
using CIEDE2000 tolerance 5; this rule is disabled when the target itself
resembles the background. Case scores average object scores. This diagnostic
supports Section~\ref{sec:exp-color} and does not enter $(F,P,L,M)$ or $H$.

\subsection{Inference Configurations}
\label{sec:app-inference}

Each model--case pair contributes one fixed output, with the exact input
image, instruction, and inference settings retained in the run records.

Table~\ref{tab:app-inference} records the inference settings. Open-weight runs
use BF16, except InstructPix2Pix (\abbrreview{half-precision floating point, FP16}), and seed 42, except FLUX.2-dev,
which uses $(42+\operatorname{CRC32}(\text{case ID}))\bmod 2^{31}$.
For a reference canvas of size $W\times H$, the shared input wrapper
centers the image on a white square of side $S=\max(W,H,1024)$; for the
released assets, this adds padding without scaling the image content.
The open-weight editors use a $1024\times1024$ image for the
generative branch, except InstructPix2Pix, which retains the incoming square
size. Step1X-Edit v1.0 uses \texttt{size\_level=1024}.
GPT-Image-2 requests a $1024\times1024$ output through its API.

For every editor, the saved content box is scaled to the returned image's
width and height, cropped to remove padding, and resized to the reference
dimensions with Lanczos interpolation when needed. This coordinate mapping
uses the recorded padding and canvas dimensions to return each
output to the reference coordinate system.

\begin{table}[!htbp]
  \centering
  \caption{Inference settings for the evaluated editors. Checkpoint gives the
  exact public release evaluated; for open-weight models, Revision denotes the
  public repository commit. Guidance values follow each model's own
  parameterization; T/I denotes text/image guidance.}
  \label{tab:app-inference}
  \small
  \setlength{\tabcolsep}{5pt}
  \begin{tabular}{lllrl}
    \hline
    Model & Checkpoint & Revision & Steps & Guidance \\
    \hline
    GPT-Image-2 & --- & API & --- & --- \\
    FireRed-Edit & FireRed-Edit-1.1 & \texttt{3bc3f2a1} & 40 & True CFG 4.0 \\
    Qwen-Image-Edit-2511 & Qwen-Image-Edit-2511 & \texttt{6f3ccc0b} & 40 & True CFG 4.0 \\
    FLUX.2-dev & FLUX.2-dev & \texttt{26afe3a7} & 50 & 4.0 \\
    Qwen-Image-Edit & Qwen-Image-Edit & \texttt{ac7f9318} & 50 & True CFG 4.0 \\
    BAGEL & BAGEL-7B-MoT & \texttt{5019f57d} & 50 & T/I: 4.0/1.5 \\
    Step1X-Edit & Step1X-Edit v1.0 & \texttt{752c0bc2} & 50 & 4.0 \\
    FLUX.2-klein & FLUX.2-klein-base-4B & \texttt{a3b4f484} & 50 & 4.0 \\
    FLUX.1-Kontext & FLUX.1-Kontext-dev & \texttt{24e9dedc} & 50 & 4.0 \\
    OmniGen2 & OmniGen2 & \texttt{df5dca8a} & 50 & T/I: 5.0/2.0 \\
    InstructPix2Pix & InstructPix2Pix & \texttt{31519b5c} & 30 & T/I: 7.5/1.5 \\
    \hline
  \end{tabular}
\end{table}

FireRed-Edit and both Qwen-Image-Edit checkpoints use \texttt{true\_cfg\_scale=4.0}
with a single-space negative prompt.
BAGEL applies guidance over $[0.4,1.0]$ with timestep shift 3.0.
\mbox{GPT-Image-2} uses the Azure image-editing API
(version \texttt{2025-04-01-preview}), medium quality, and one
$1024\!\times\!1024$ output per request; no seed, step count, or guidance
scale is sent to the service.

\section[Coverage of Related Datasets]{{Coverage of Related Datasets}}
\label{sec:app-benchmark-comparison}

Table~\ref{tab:app-benchmark-comparison} compares representative general
image-editing benchmarks with closely related chart and SVG editing
benchmarks. The latter include both image-output and SVG-code editing
interfaces. VeriEdit-Bench combines source-rendered references and separate
pixel-based edit/preservation measurements across four design asset types.

\begin{table}[ht]
  \centering
  \caption{Asset coverage, editing operations, and verification in related datasets.
  $\checkmark$ indicates coverage documented in the paper's task taxonomy
  or examples; --- means not explicitly reported there.
  $\checkmark^{\dagger}$ denotes PaintBench's TinyGrafixBench extension.
  Presence does not imply equal task counts or coverage of every
  domain--operation combination.}
  \label{tab:app-benchmark-comparison}
  \begingroup
  \footnotesize
  \setlength{\tabcolsep}{1.5pt}
  \renewcommand{\arraystretch}{1.08}
  \newcommand{\covcell}[2][c]{{\renewcommand{\arraystretch}{1}%
    \begin{tabular}[c]{@{}#1@{}}#2\end{tabular}}}
  \newcommand{\cy}{\ensuremath{\checkmark}}
  \newcommand{\ce}{\ensuremath{\checkmark^{\dagger}}}
  \newcommand{\cn}{---}
  \textbf{(a) Asset coverage and verification}\par\smallskip
  \begin{tabular*}{\linewidth}{@{\extracolsep{\fill}}llcccccc@{}}
    \hline
    & & \multicolumn{4}{c}{Design assets} & \multicolumn{2}{c}{Verification} \\
    \cline{3-6}\cline{7-8}
    Dataset & \covcell[l]{Other visual\\domains} & \covcell{SVG\\artwork} & Charts & Web & Slides &
    \covcell{Source\\target} & \covcell{Regional\\pixels} \\
    \hline
    \covcell[l]{MagicBrush\\\citep{DBLP:conf/nips/ZhangMCSS23}} & Natural images & \cn & \cn & \cn & \cn & \cn & \cn \\
    \noalign{\vskip 1.5pt\hrule height 0.2pt\vskip 1.5pt}
    \covcell[l]{GIE-Bench\\\citep{DBLP:journals/corr/abs-2505-11493}} & \covcell[l]{Photographs;\\illustrations} & \cn & \cn & \cn & \cn & \cn & \cn \\
    \noalign{\vskip 1.5pt\hrule height 0.2pt\vskip 1.5pt}
    \covcell[l]{PaintBench\\\citep{DBLP:journals/corr/abs-2606-00188}} & \covcell[l]{Shape canvases;\\network graphs$^{\dagger}$} & \cn & \ce & \cn & \cn & \cy & \cy \\
    \noalign{\vskip 1.5pt\hrule height 0.2pt\vskip 1.5pt}
    \covcell[l]{StructBench\\\citep{DBLP:journals/corr/abs-2510-05091}} & \covcell[l]{Math, science,\\puzzles, graphs,\\tables} & \cn & \cy & \cn & \cn & \cy & \cn \\
    \noalign{\vskip 1.5pt\hrule height 0.2pt\vskip 1.5pt}
    \covcell[l]{FigEdit\\\citep{DBLP:journals/corr/abs-2512-00752}} & \cn & \cn & \cy & \cn & \cn & \cy & \cn \\
    \noalign{\vskip 1.5pt\hrule height 0.2pt\vskip 1.5pt}
    \covcell[l]{ChartSync\\\citep{DBLP:journals/corr/abs-2607-10301}} & \cn & \cn & \cy & \cn & \cn & \cy & \cn \\
    \noalign{\vskip 1.5pt\hrule height 0.2pt\vskip 1.5pt}
    \covcell[l]{SVGEditBench V2\\\citep{DBLP:journals/corr/abs-2502-19453}} & \cn & \cy & \cn & \cn & \cn & \cy & \cn \\
    \noalign{\vskip 1.5pt\hrule height 0.2pt\vskip 1.5pt}
    \covcell[l]{VectorEdits\\\citep{DBLP:conf/lrec/KucharKSS26}} & \cn & \cy & \cn & \cn & \cn & \cy & \cn \\
    \noalign{\vskip 1.5pt\hrule height 0.2pt\vskip 1.5pt}
    \covcell[l]{Vector-Bench\\\citep{DBLP:journals/corr/abs-2607-19056}} & \cn & \cy & \cn & \cn & \cn & \cy & \cn \\[2pt]
    \hline
    \textbf{VeriEdit-Bench} & \cn & \cy & \cy & \cy & \cy & \cy & \cy \\
    \hline
  \end{tabular*}

  \medskip
  \textbf{(b) Editing operations}\par\smallskip
  \begin{tabular*}{\linewidth}{@{\extracolsep{\fill}}lccccccc@{}}
    \hline
    Dataset & Color & Geometry & \covcell{Text /\\labels} & \covcell{Font\\size} &
    \covcell{Lines /\\borders} & \covcell{Opacity /\\blending} & \covcell{Hide /\\remove} \\
    \hline
    MagicBrush & \cy & \cn & \cy & \cn & \cn & \cn & \cy \\
    GIE-Bench & \cy & \cy & \cy & \cn & \cn & \cn & \cy \\
    PaintBench & \cy & \cy & \cn & \cn & \cy & \cy & \cy \\
    StructBench & \cy & \cy & \cy & \cn & \cy & \cn & \cy \\
    FigEdit & \cy & \cy & \cy & \cy & \cn & \cn & \cy \\
    ChartSync & \cn & \cy & \cy & \cn & \cn & \cn & \cn \\
    SVGEditBench V2 & \cy & \cy & \cn & \cn & \cy & \cn & \cy \\
    VectorEdits & \cn & \cy & \cn & \cn & \cy & \cn & \cn \\
    Vector-Bench & \cy & \cy & \cn & \cn & \cy & \cy & \cy \\
    \hline
    \textbf{VeriEdit-Bench} & \cy & \cy & \cy & \cy & \cy & \cy & \cy \\
    \hline
  \end{tabular*}

  \smallskip
  \begin{minipage}{\linewidth}
  \scriptsize
  Source target: a renderable reference program, including paired target SVGs.
  Regional pixels: separate target-agreement and preservation checks using fixed
  pixel rules. Vector-Bench verifies requested attributes and protected structure
  directly in SVG code;
  SVGEditBench V2 and VectorEdits combine rendered-image similarity with geometric
  or validity measures. All three evaluate SVG-code inputs and outputs.
  ChartSync combines textual and image measurements with VLM judgments of
  text--geometry consistency. VeriEdit-Bench evaluates image-and-instruction
  inputs and image outputs, with source programs hidden from editors.
  Geometry includes position, size, layout, or shape; lines/borders includes
  outline style and border appearance; opacity/blending includes alpha blending.
  \end{minipage}
  \endgroup
\end{table}

\end{document}